\documentclass[runningheads]{llncs}

\usepackage[mobile]{eccv}

\usepackage{eccvabbrv}

\usepackage{graphicx}
\usepackage{booktabs}
\usepackage{multirow}
\usepackage{tabularx}
\usepackage{amssymb}
\usepackage{pifont}
\usepackage{caption}
\usepackage{subcaption}
\usepackage{mathtools}
\usepackage{algpseudocode}
\usepackage{amsmath}

\usepackage[accsupp]{axessibility}  

\usepackage[pagebackref,breaklinks,colorlinks,citecolor=eccvblue]{hyperref}

\usepackage{orcidlink}

\usepackage{algorithm}
\usepackage{wrapfig}

\usepackage{multirow}
\usepackage{tabularx}
\usepackage{arydshln}
\usepackage{array}

\begin{document}

\title{Self-Rectifying Diffusion Sampling\\ with Perturbed-Attention Guidance}

\titlerunning{Self-Rectifying Diffusion Sampling with Perturbed-Attention Guidance}

\author{Donghoon Ahn$^{\ast}$\inst{1}\orcidlink{0009-0007-2602-6689} \and
Hyoungwon Cho$^{\ast}$ \inst{1}\orcidlink{0009-0007-1669-6013} \and
Jaewon Min \inst{1}\orcidlink{0009-0002-9130-7587} \and
Wooseok Jang \inst{1}\orcidlink{0009-0000-7104-2943} \and
Jungwoo Kim\inst{1}\orcidlink{0009-0001-5915-056X} \and
SeonHwa Kim\inst{1}\orcidlink{0009-0006-5621-2688} \and
Hyun Hee Park\inst{2}\orcidlink{0009-0008-9880-4168} \and \\
Kyong Hwan Jin$^{\dagger}$\inst{1}\orcidlink{0000-0001-7885-4792} \and
Seungryong Kim$^{\dagger}$\inst{1}\orcidlink{0000-0003-2927-6273}
}
\authorrunning{D.~Ahn et al.}

\institute{Korea University
\and 
Samsung Electronics \\
}
\maketitle

\begingroup
\renewcommand{\thefootnote}{}
\footnotetext{\scriptsize $\ast$: Equal contribution}
\footnotetext{\scriptsize $\dagger$: Co-corresponding author}
\endgroup

\vspace{-15pt}
\begin{center}
\small
\url{https://ku-cvlab.github.io/Perturbed-Attention-Guidance}
\end{center}

\begin{figure}
    \centering
    \vspace{-15pt}
    \includegraphics[width=\textwidth]{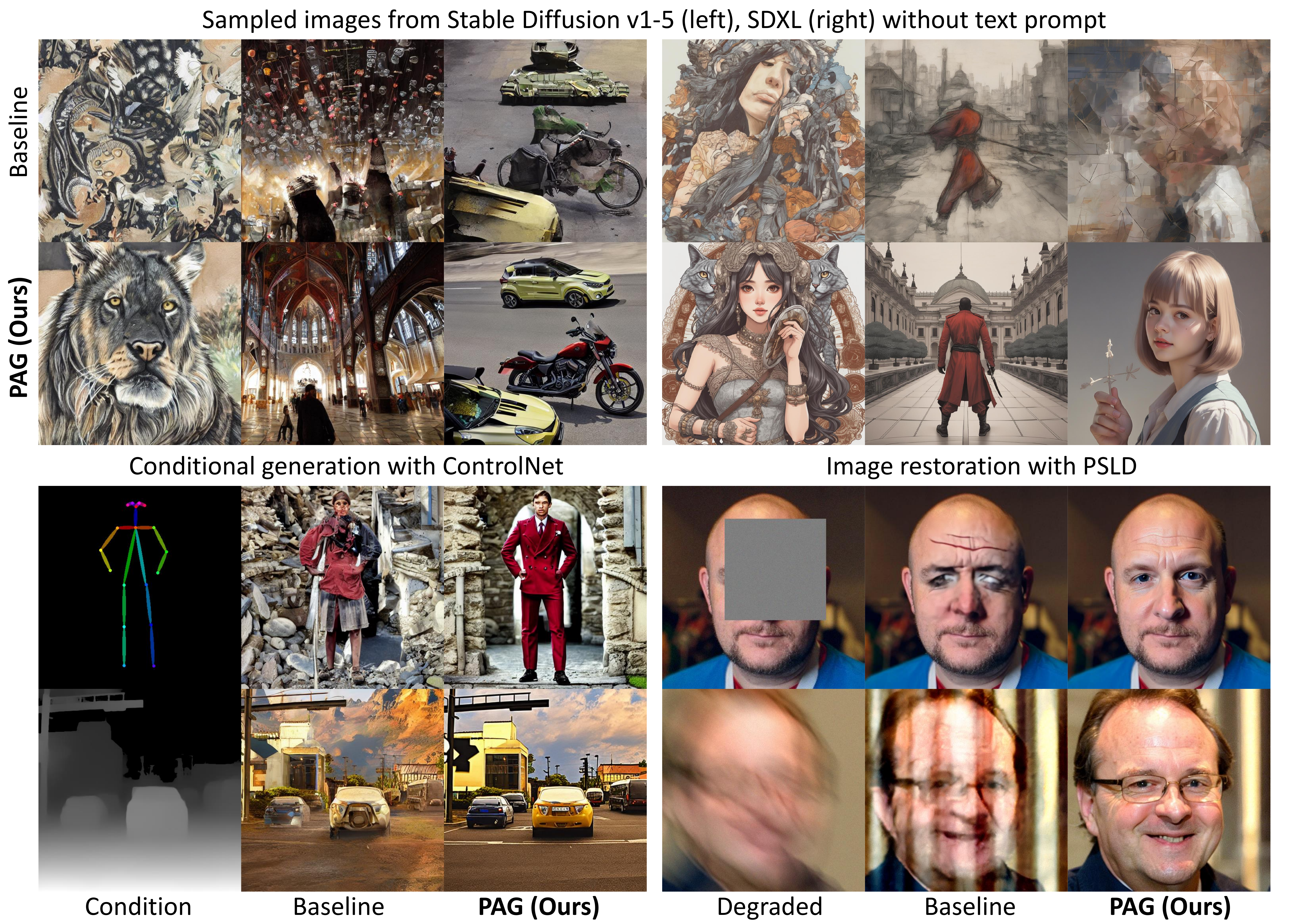}
    \vspace{-15pt}
    \caption{\textbf{Qualitative comparisons between unguided (baseline) and perturbed-attention-guided (PAG) diffusion samples.} Without any \textbf{\textit{external conditions}}, \eg, class labels or text prompts, or \textbf{\textit{additional training}}, our \textbf{PAG} dramatically elevates the quality of diffusion samples even in unconditional generation, where classifier-free guidance (CFG)~\cite{ho2022classifier} is inapplicable. Our guidance can also enhance the baseline performance in various downstream tasks such as ControlNet~\cite{zhang2023adding} with empty prompt and inverse problems such as inpainting and deblurring~\cite{chung2022diffusion,rout2024solving}.}
    \label{fig:teaser}
    \vspace{-30pt}
\end{figure}

\begin{abstract}

Recent studies have demonstrated that diffusion models can generate high-quality samples, but their quality heavily depends on sampling guidance techniques, such as classifier guidance (CG) and classifier-free guidance (CFG). These techniques are often not applicable in unconditional generation or various downstream tasks such as the solving inverse problems. In this paper, we propose novel sampling guidance, called \textbf{Perturbed-Attention Guidance (PAG)}, which improves diffusion sample quality across both unconditional and conditional settings, achieving this without requiring additional training or the integration of external modules. PAG progressively enhances the structure of samples throughout the denoising process by generating intermediate samples with degraded structures and guiding the denoising process away from these degraded samples. These degraded samples are created by substituting selected self-attention maps in the diffusion U-Net, which capture structural information between image patches, with an identity matrix. 
In both ADM and Stable Diffusion, PAG surprisingly improves sample quality in conditional and even unconditional generation. Moreover, PAG significantly enhances baseline performance in various downstream tasks where existing guidance methods such as CG or CFG cannot be fully utilized, including ControlNet with empty prompts and solving inverse problems such as inpainting and deblurring. To the best of our knowledge, this is the first approach to apply guidance in solving inverse problems using diffusion models.

\end{abstract}
\section{Introduction}

 
Diffusion models~\cite{sohl2015deep,song2019generative,ho2020denoising,song2020score,rombach2022high} have gained prominence in image generation, demonstrating their capability to produce high-fidelity and diverse samples. Sampling guidance techniques, such as classifier guidance (CG)~\cite{dhariwal2021diffusion} and classifier-free guidance (CFG)~\cite{ho2022classifier}, are crucial for directing diffusion models to generate higher-quality images. Without these techniques, as shown in Fig.~\ref{fig:teaser} and Fig.~\ref{fig:cfg-ours-visualization}, diffusion models often produce lower-quality images, typically exhibiting collapsed structures. Despite their widespread use, these guidance methods have drawbacks: they require additional training or the integration of external modules, often reduce the diversity of the output samples, and are unavailable in unconditional generation.

Meanwhile, unconditional generation offers significant practical advantages. It aids in understanding the fundamental principles of data creation and its underlying structures~\cite{chen2020simple, RCG2023}. Furthermore, advancements in unconditional techniques often enhance conditional generation. Importantly, it eliminates the need for potentially costly and complex human annotations such as class labels, text, and segmentation maps, which can be a major hurdle in tasks where accurate labeling is difficult, such as modeling molecular structures~\cite{RCG2023}. Finally, unconditional generative models provide powerful general priors, as evidenced by their use in solving inverse problems~\cite{song2020score,kawar2022denoising,wang2022zero,chung2022improving,chung2022diffusion,rout2024solving,rout2023beyond}. However, the unavailability of CG~\cite{dhariwal2021diffusion} or CFG~\cite{ho2022classifier} can lead to sub-optimal performance.

Recognizing the importance of unconditional generation, we propose a novel sampling guidance method called \textbf{Perturbed-Attention Guidance (PAG)}. PAG improves diffusion sample quality in both unconditional and conditional settings without requiring additional training or the integration of external modules. Our approach leverages an implicit discriminator to distinguish between desirable and undesirable samples. By utilizing the capability of self-attention maps in the diffusion U-Net to capture structural information~\cite{nam2024dreammatcher,balaji2022ediffi,tewel2023key,tumanyan2022plugand,hertz2022prompt}, we generate undesirable samples by substituting the diffusion model's self-attention map with an identity matrix and guide the denoising process away from these degraded samples. These undesirable samples help steer the denoising trajectory away from the structural collapse commonly observed in unguided generation.

Extensive experiments validate the effectiveness of our guidance method. Applied to ADM~\cite{dhariwal2021diffusion}, it exceptionally improves sample quality in both conditional and unconditional settings. We also observe remarkable enhancements, both qualitatively and quantitatively, when applied to the widely-used Stable Diffusion~\cite{rombach2022high}. Additionally, combining PAG with conventional guidance methods such as CFG~\cite{ho2022classifier} leads to further improvements. Finally, our guidance profoundly enhances the performance of diffusion models in various downstream tasks, such as inverse problems~\cite{chung2022diffusion,rout2024solving} and ControlNet~\cite{zhang2023adding} with empty prompts, where the lack of conditions renders CFG~\cite{ho2022classifier} unusable. Notably, we have opened new avenues for fully leveraging the generative capabilities of diffusion models in solving inverse problems.
\section{Related Work}
\vspace{-5pt}
\subsubsection{Diffusion models.}

Diffusion models (DMs)~\cite{sohl2015deep, song2020score, song2019generative} have set a high benchmark in image generation, achieving remarkable results in both sample quality and distribution estimation. DDIM~\cite{song2020denoising} improves sampling speed by applying a non-Markovian process. Latent diffusion models (LDMs)~\cite{rombach2022high} operate in a compressed latent space, balancing computational efficiency and synthesis quality.

\vspace{-10pt}
\subsubsection{Sampling guidance for diffusion models.}
The surge in diffusion model research is largely attributed to advancements in sampling guidance techniques~\cite{dhariwal2021diffusion, ho2022classifier}. Classifier guidance (CG)~\cite{dhariwal2021diffusion} increases fidelity at the expense of diversity by adding the gradient of a pre-trained classifier. Classifier-free guidance (CFG)~\cite{ho2022classifier} models an implicit classifier to achieve similar effects as CG. Self-attention guidance (SAG)~\cite{hong2023improving} enhances sample quality in an unconditional framework by using adversarial blurring to obscure crucial information and then guiding the sampling process with noise predicted from both blurred and original samples. Additionally, various guidance methods focus on conditioning~\cite{luo2023readout} or image editing~\cite{epstein2024diffusion, bansal2023universal}.

\section{Preliminaries}

\subsubsection{Diffusion models.}

In diffusion models~\cite{ho2020denoising, ho2022classifier, dhariwal2021diffusion, song2020score}, random noise $\epsilon \sim \mathcal{N}(0,I)$ is added during forward path to an image $x_0$ to produce a noisy image $x_t$ at an arbitrary timestep $t$:
\begin{equation}
x_{t} = \sqrt{\bar{{\alpha}}_{t}} x_{0} + \sqrt{1-\bar{{\alpha}}_{t}}\epsilon,
\end{equation} 
with ${\alpha}_t = 1-{\beta}_t$ and $\bar{{\alpha}}_{t} = \prod_{s=1}^{t} {\alpha}_s$ according to a variance schedule ${\beta}_1,...,{\beta}_t$. A denoising network $\epsilon_\theta$ is learned to predict $\epsilon$ by optimizing an objective
\begin{equation}
\mathcal{L} = \mathbb{E}_{x_{0}, t, \epsilon \sim \mathcal{N}(0,I)} \left[ \left\lVert \epsilon - \epsilon_{\theta}(x_{t}, t) \right\rVert_2^2 \right],
\end{equation}
for uniformly sampled $t\in \{ 1,...,T \}$. 

During sampling, the model produces denoised image $x_{t-1}$ from $x_{t}$ at each timestep $t$ based on the noise estimation $\epsilon_{\theta}(x_{t}, t)$ as follows:
\begin{equation}
\label{eq:ddpm_reverse_onestep}
    x_{t-1} = \frac{1}{\sqrt{\bar{\alpha}_{t}}} \left( x_t - \frac{\beta_t}{\sqrt{1-\bar{\alpha}_t}} \epsilon_{\theta}(x_{t}, t) \right) + \sigma_t z,
\end{equation}
where $z \sim \mathcal{N}(0,I)$ and  $\sigma_t^2 $ is set to $\beta_t$. Starting with randomly sampled noise $x_T \sim \mathcal{N}(0,I)$, the process is applied iteratively to generate a clean image $x_0$. For the sake of simplicity, throughout the remainder of this paper, we adopt the notation \(\epsilon_{\theta}(x_{t})\) to represent \(\epsilon_{\theta}(x_{t}, t)\). Note that noise estimation of the diffusion model can be considered as ${\epsilon}_{\theta}(x_t) \approx -{\sigma}_t {\nabla}_{x_t} \mathrm{log}\, p(x_t)$~\cite{ho2022classifier, dhariwal2021diffusion, song2019generative, song2020score}, where $p(x_t)$ denotes the distribution of $x_t$.

In addition, using the reparameterization trick, it is possible to obtain the intermediate prediction of $x_0$ at a given timestep $t$ as
\begin{equation}
\hat{x}_{0} = {(x_t - \sqrt{1-\bar{\alpha}_{t}}\epsilon_{\theta}(x_{t}, t))}/
    {\sqrt{\bar{\alpha}_{t}}}.
\label{eq:eps-to-x0}
\end{equation}

\vspace{-10pt}
\subsubsection{Classifier-free guidance.}
\label{subsubsec:cfg}
To enhance the generation towards arbitrary class label $c$, CG~\cite{dhariwal2021diffusion} introduces a new sampling distribution $\Tilde{p}_{\theta}(x_t|c)$ composed with both ${p}_{\theta}(x_t|c)$ and the classifier distribution ${{p}_{\theta}(c|x_t)}$, which is expressed as 
\begin{equation}
\Tilde{p}_{\theta}(x_t|c) \propto {p}_{\theta}(x_t|c) {{p}_{\theta}(c|x_t)}^s,
\end{equation}
where $s$ is the scale parameter. It turns out that sampling from this distribution with  $s>0$ leads the model to generate saturated samples with high probabilities for the input class labels, resulting in increased quality but decreased sample diversity~\cite{dhariwal2021diffusion}.

CG, however, has a drawback in that it requires a pretrained classifier for noisy images of each timestep. To address this issue, CFG~\cite{ho2022classifier}  modifies the classifier distribution ${{p}_{\theta}(c|x_t)}$ by combining the conditional distribution ${p}_{\theta}(x_t|c)$ and the unconditional distribution ${p}_{\theta}(x_t)$:
\begin{align}
\Tilde{p}_{\theta}(x_t|c)
& \propto {p}_{\theta}(x_t|c) {{p}_{\theta}(c|x_t)}^s
\nonumber = {p}_{\theta}(x_t|c) \left[ \frac{{p}_{\theta}(x_t|c){p}_{\theta}(c)}{{p}_{\theta}(x_t)} \right]^s
\nonumber\\ 
& = {{p}_{\theta}(x_t|c)}^{1+s} {{p}_{\theta}(x_t)}^{-s}.
\end{align}
Then the score of new conditional distribution $\Tilde{p}_\theta(x_t|c)$ would be
$
\nabla_{x_t}\log\,\Tilde{p}_\theta(x_t|c) \\= (1+s){\epsilon}^*(x_t, c) - s{\epsilon}^*(x_t),
$ where $\epsilon^*$ denotes true score. By approximating this score using conditional and unconditional score estimates, we have
\begin{align}
\Tilde{\epsilon}_{\theta}(x_t, c)
& = (1+s){\epsilon}_{\theta}(x_t, c) - s{\epsilon}_{\theta}(x_t)
\nonumber\\ & = {\epsilon}_{\theta}(x_t, c) + s ( {\epsilon}_{\theta}(x_t, c) - {\epsilon}_{\theta}(x_t) ) = {\epsilon}_{\theta}(x_t, c) + s \Delta_t.
\end{align}
\begin{figure}[t]
    \centering
    \begin{subfigure}{\textwidth}
        \includegraphics[width=\textwidth]{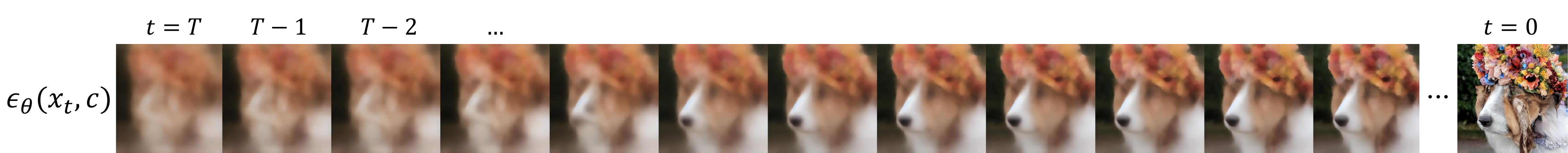}
        \caption{Diffusion sampling without CFG}
    \end{subfigure}
    \begin{subfigure}{\textwidth}
        \centering
        \includegraphics[width=\textwidth]{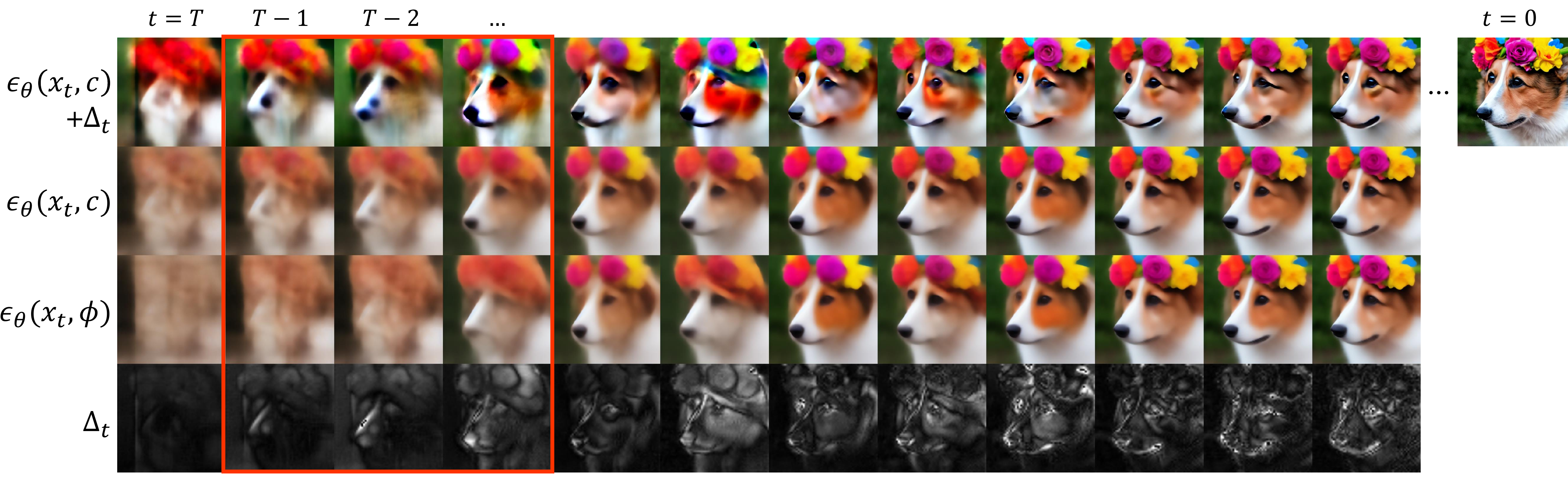}
        \caption{Diffusion sampling with CFG}
    \end{subfigure}
    \vspace{-15pt}
    \caption{\textbf{Visualization of reverse process w/o and w/CFG~\cite{ho2022classifier}}. To visualize the predicted epsilon, we first convert it into \(\hat{x}_0\) following Eq.~\ref{eq:eps-to-x0}. For the guidance signal $\Delta_t={\epsilon}_{\theta}(x_t, c) - {\epsilon}_{\theta}(x_t, \phi)$, we apply an absolute value function and calculate the mean across all channels.  We use the same latent and seed for both cases. \textbf{(a)} Without CFG, diffusion models generate samples with collapsed structures. \textbf{(b)} With CFG, diffusion models generate samples that are well-aligned to the prompt. The \textcolor{red}{red} rectangles highlight the distinction between \textit{conditional} (${\epsilon}_{\theta}(x_t, c)$) and \textit{unconditional} (${\epsilon}_{\theta}(x_t, \phi)$) predictions. Without prompt, diffusion models lack guidance on what to generate in the early stages, often leading to the omission of salient features such as eyes and nose, and thus adding $\Delta_t$ amplifies features relevant to the prompt. Here the prompt \textit{``a corgi with flower crown''} is used.}
    \vspace{-10pt}
    \label{fig:cfg-ours-visualization}
\end{figure}
In practice, \({\epsilon}_{\theta}(x_t, c)\) and \({\epsilon}_{\theta}(x_t)\) are parameterized by a single neural network, which is jointly trained for both conditional and unconditional generation by assigning a null token $\phi$ as the class label for the unconditional model, such that  ${\epsilon}_{\theta}(x_t) \approx {\epsilon}_{\theta}(x_t,\phi)$. The guidance signal $\Delta_t ={\epsilon}_{\theta}(x_t, c) - {\epsilon}_{\theta}(x_t, \phi)$ acts as the gradient of the implicit classifier, producing images that closely adhere to condition \(c\). In Fig.~\ref{fig:cfg-ours-visualization}, we visualize \(\Delta_t\) across timesteps and explain its role in enhancing sample quality. A more detailed exploration of CFG's workings is available in the Appendix~\ref{sec:sup:cfg-further-analysis}.
\section{PAG: Perturbed-Attention Guidance}

\subsection{Self-rectifying sampling with implicit discriminator}
\subsubsection{Perturbation guidance.} Recently, it has been shown that the sampling guidance of diffusion models can be generalized as the gradient of the energy function, for instance, which can be a negative class probability of classifier~\cite{dhariwal2021diffusion}, negative CLIP similarity score~\cite{nichol2021glide}, any type of time-independent energy~\cite{bansal2023universal}, the distance between extracted signal such as pose and edges and reference signal~\cite{luo2023readout} or any energy function which takes the noisy sample~\cite{epstein2024diffusion}.

In this work, we introduce an implicit discriminator denoted $\mathcal{D}$ that differentiates \textit{desirable} samples following real data distribution from \textit{undesirable} ones during the diffusion process. Similar to CFG~\cite{ho2022classifier} where the implicit classifier guides samples to be more closely aligned with the given class label, the implicit discriminator $\mathcal{D}$ guides samples towards the desirable distribution and away from the undesirable distribution.
By applying Bayes' rule, we first define the implicit discriminator as
\begin{align}
\mathcal{D}(x_t) = 
\frac{p(y|x_t)}{p(\hat{y}|x_t)} =
\frac{p(y){p(x_t|y)}}{p(\hat{y}){p(x_t|\hat{y})}},
\label{eq:disc}
\end{align}
where $y$ and $\hat{y}$ denote the imaginary labels for desirable sample and undesirable sample, respectively.

Then similar to WGAN~\cite{arjovsky2017wasserstein,wu2018wasserstein}, we set the generator loss of the implicit discriminator as our energy function, $\mathcal{L}_\mathcal{G}$, and compute its derivative as
\begin{align}
{\nabla}_{x_t} \mathcal{L}_\mathcal{G}
&={\nabla}_{x_t} \left[- \mathrm{log}\: \mathcal{D}(x_t) \right] 
\nonumber\\ & = {\nabla}_{x_t}\left[ - \mathrm{log}\: \frac{p(y){p(x_t|y)}}{p(\hat{y}){p(x_t|\hat{y})}}\right] = {\nabla}_{x_t}\left[ - \mathrm{log}\: \frac{p(x_t|y)}{p(x_t|\hat{y})}\right] 
\nonumber\\ & = -{\nabla}_{x_t}(\mathrm{log}\:p(x_t|y) - \mathrm{log}\:p(x_t|\hat{y})).
\label{eq:generator_grad}
\end{align}

Then, using Eq.~\ref{eq:generator_grad}, we define a new diffusion sampling such that 
\begin{align}
\Tilde{\epsilon}_{\theta}(x_t) &= {\epsilon}_{\theta}(x_t) + s{\sigma}_t {\nabla}_{x_t} \mathcal{L}_\mathcal{G}
\nonumber\\&= {\epsilon}_{\theta}(x_t) - s{\sigma}_t {\nabla}_{x_t} ( \mathrm{log}\: p(x_t|y) - \mathrm{log}\: p(x_t|\hat{y}) )
\nonumber\\& = {\epsilon}_{\theta}(x_t) + s ( \epsilon_{\theta}(x_t) - \hat{\epsilon}_{\theta}(x_t) ) = {\epsilon}_{\theta}(x_t) + s \hat{\Delta}_t.
\label{eq:PAG-derivation}
\end{align}
Since diffusion models have already learned the desired distribution, we use the pretrained score estimation network $\epsilon_\theta(x_t)$ as an approximation of $-\sigma_t{\nabla}_{x_t} \mathrm{log}\, p(x_t|y)$. For the score with undesirable label $\hat{y}$, we approximate it by \textit{perturbing} the forward pass of pretrained network which we denote $\hat{\epsilon}_\theta(x_t)$. Note that $\hat{\epsilon}_\theta(x_t)$ can embody any form of perturbation during the epsilon prediction process, including perturbations applied to the input~\cite{hong2023improving} or internal representations, or both. We call this \textbf{perturbation guidance}, as it guides sampling by simulating undesirable predictions via perturbations.

\vspace{-15pt}
\subsubsection{Connections to CFG.} The formulation in Eq.~\ref{eq:PAG-derivation} resembles CFG~\cite{ho2022classifier}. Indeed, it is noteworthy that CFG can be considered a particular instance within our broader formulation.
First, Eq.~\ref{eq:PAG-derivation} can also be defined in class-conditional diffusion models such that 
\begin{align}
\Tilde{\epsilon}_{\theta}(x_t, c) = {\epsilon}_{\theta}(x_t,c) + s ( \epsilon_{\theta}(x_t,c) - \hat{\epsilon}_{\theta}(x_t,c) ).
\end{align}
In CFG, $\hat{\epsilon}_{\theta}(x_t,c)$ is implemented by dropping the class label, resulting in ${\epsilon}_{\theta}(x_t, \phi)$, which in our terminology can be described as a \textit{perturbed} forward pass. In this paper, we extend the concept of the \textit{perturbed} forward pass to be more applicable even to the unconditional diffusion models.

\vspace{-10pt}
\subsection{Perturbing self-attention of U-Net diffusion model}

\begin{figure}[t]
    \centering
    \begin{subfigure}{\textwidth}
        \centering
        \includegraphics[width=\textwidth]{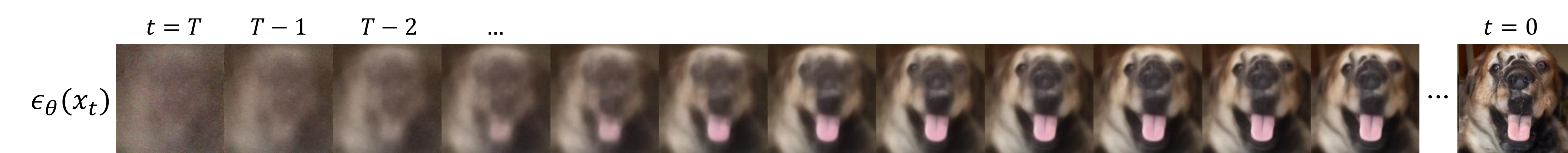}
        \caption{Sampling process without PAG}
    \end{subfigure}
    \begin{subfigure}{\textwidth}
        \centering
        \includegraphics[width=\textwidth]{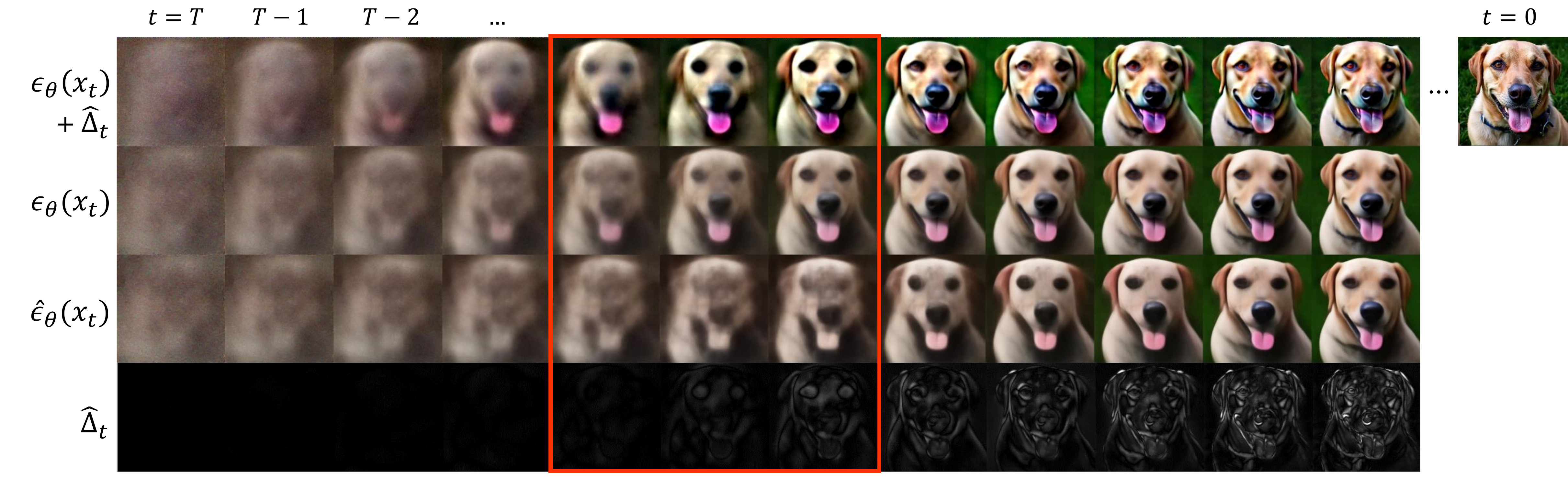}
        \caption{Sampling process with PAG}
    \end{subfigure}
    \vspace{-15pt}
    \caption{\textbf{Visualization of sampling process w/o and w/ PAG.} To visualize predicted epsilon, we first convert it into \(\hat{x}_0\) following Eq.~\ref{eq:eps-to-x0}.  For the guidance signal $\hat{\Delta}_t = {\epsilon}_{\theta}(x_t) - \hat{\epsilon}_{\theta}(x_t)$, we apply an absolute value function and calculate the mean across all channels. We use the same latent and seed for both cases.
    \textbf{(a)} Without guidance, diffusion models generate samples with collapsed structures. \textbf{(b)} With our PAG, diffusion models generate improved samples. The \textcolor{red}{red} rectangles highlight the distinction between the \textit{original} (\({\epsilon}_{\theta}(x_t)\)) and \textit{perturbed} (\({\hat{\epsilon}}_{\theta}(x_t)\)) predictions. With perturbed self-attention, the diffusion model lacks an understanding of the global structure, often leading to the omission of salient features such as eyes, nose, and tongue. Adding \(\hat{\Delta}_t\) thus enhances features that can only be accurately rendered with global structure information.}
    \vspace{-10pt}
    \label{fig:pag_path_visualization}
\end{figure}

In our perturbation guidance framework, the strategy for implementing $\hat{\epsilon}_\theta(x_t)$ can be chosen arbitrarily. However, perturbing the input image or the condition directly can cause the out-of-distribution problem, lead the diffusion model to create incorrect guidance signals, and steer the diffusion sampling toward the erroneous direction. To overcome this, CFG~\cite{ho2022classifier} explicitly trains an unconditional model. In addition, SAG~\cite{hong2023improving} employs partial blurring to minimize deviation, but without careful selection of hyperparameters, it often deviates from the desired trajectory. This behavior is illustrated in Fig.~\ref{fig:sup:sag-ours-scale-comparision} in Appendix~\ref{sec:sup:comparision-with-sag}.

On the other hand, some studies have explored manipulating cross-attention and self-attention maps of the diffusion models for various tasks~\cite{hertz2022prompt, simsar2023lime,cao2023masactrl,qi2023fatezero,khachatryan2023text2video}. They show that modifying the attention maps has minimal impact on the model's ability to generate plausible outputs. We target the self-attention mechanism to design a perturbation strategy applicable to both conditional and unconditional models.

Another criterion for selecting perturbations involves determining which aspects of the samples should be improved during the sampling process. As illustrated in the top row of Fig.~\ref{fig:teaser} and Fig.~\ref{fig:cfg-ours-visualization}, images generated by diffusion models without guidance often exhibit collapsed structures. To address this, the desired guidance should steer the denoising trajectory away from the sample exhibiting a collapsed structure, akin to how the null prompt in CFG is employed to strengthen class conditioning. Recently, several studies~\cite{nam2024dreammatcher,balaji2022ediffi,tewel2023key,tumanyan2022plugand,hertz2022prompt} demonstrate that the attention map contains structural information or semantic correspondence between patches. Thus, perturbing the self-attention map can generate a sample with a collapsed structure. We visualize the perturbed epsilon prediction in Fig.~\ref{fig:pag_path_visualization} in the same manner as in Fig.~\ref{fig:cfg-ours-visualization}. Notably, within the red box in Fig.~\ref{fig:pag_path_visualization} (b), it can be seen that the generated samples have collapsed structures compared to the original prediction, while preserving the overall appearance of the original sample, attributable to the attention map's robustness to manipulation.

\vspace{-15pt}
\subsubsection{Perturbed self-attention.}

\begin{figure}[t]
    \centering
    \makebox[\textwidth]{\includegraphics[width=\textwidth]{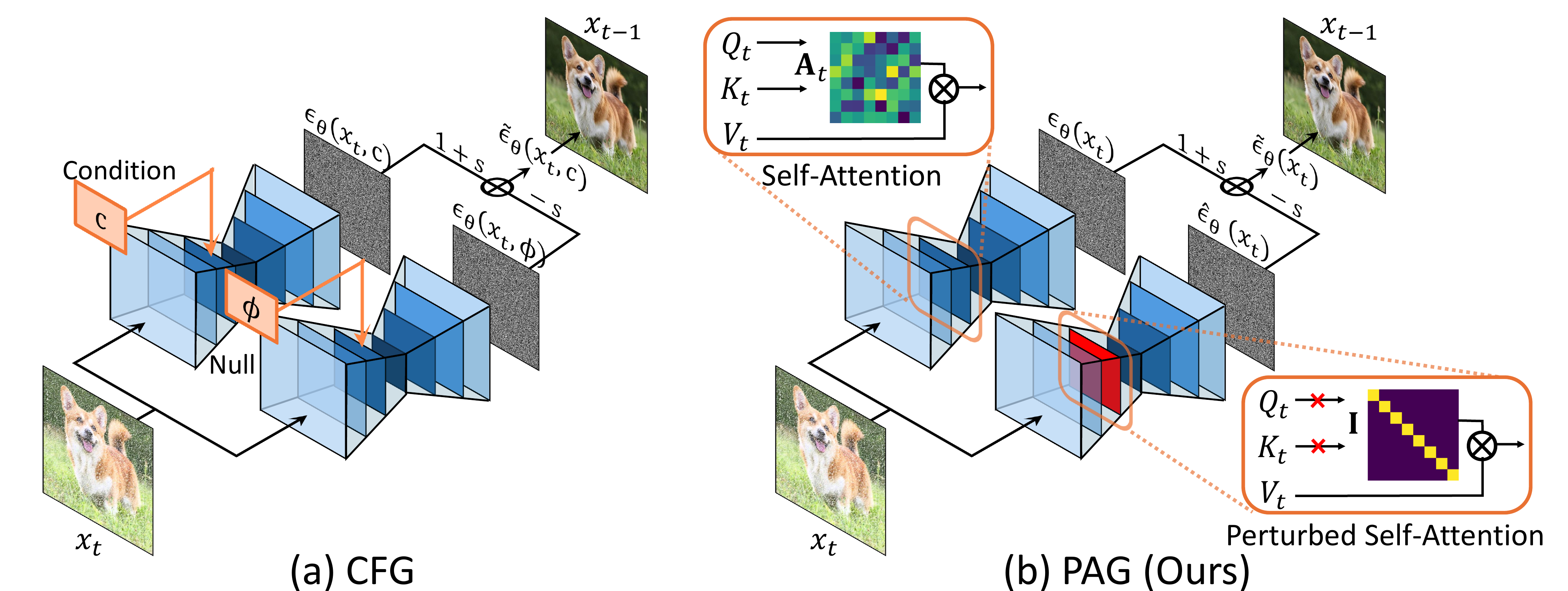}}
    \caption{\textbf{Conceptual comparison between CFG~\cite{ho2022classifier} and PAG.} CFG~\cite{ho2022classifier} employs jointly trained unconditional model as the \textit{undesirable} path, whereas PAG utilizes perturbed self-attention for the same purpose. $\mathbf{A}_t$ corresponds to the self-attention map $\mathrm{Softmax}({Q_t}{K^T_t}/\sqrt{d})$. In PAG, we perturb this by replacing with an identity matrix $\mathbf{I}$.}
    \vspace{-1em}
    \label{fig:overall}
\end{figure}

Recent studies~\cite{tumanyan2022plugand, tewel2023key, balaji2022ediffi, hertz2022prompt} have shown that the self-attention module in diffusion U-Net~\cite{ronneberger2015u} has two paths that have different roles, the query-key similarities for \textit{structure} and values for \textit{appearance}.
Specifically, in the self-attention module, we compute the query $ Q_t \in \mathbb{R}^
{(h \times w) \times d}$, key $K_t \in \mathbb{R}^{(h \times w) \times d}$, value $V_t \in \mathbb{R}^{(h \times w) \times d}$ at timestep $t$, where $h$, $w$, and $d$ refer to the height, width, and channel dimensions, 
respectively. The resulting output from this module is defined by:
\begin{equation}
\label{equ:attention}
\mathrm{SA}(Q_t, K_t, V_t) = \underbrace{\mathrm{Softmax}\left(\frac{{Q_t}{K^T_t}}{\sqrt{d}}\right)}_\textit{structure}\overbrace{V_t}^\textit{appearance}=\mathbf{A}_t V_t,
\end{equation}
where the \textit{structure} part is commonly referred to as the self-attention map.

Motivated by this insight, we focus on perturbing only the self-attention map to minimize excessive deviation from the original sample. This perspective can also be understood from the viewpoint of addressing out-of-distribution (OOD) issues for neural network inputs. Directly perturbing the appearance component $V_t$ may cause the subsequent multilayer perceptron (MLP) to encounter inputs that it has not previously seen. This leads to OOD issues for MLP, resulting in significantly distorted samples. We will discuss this further in the experiments.

However, a linear combination of value features, such as using an identity matrix as a self-attention map that maintains the value of each element, is more likely to remain within the domain than direct perturbations to $V_t$. Therefore, we only perturb the \textit{structural} component, $\mathbf{A}_t = \mathrm{Softmax}({Q_t}{K^T_t}/\sqrt{d}) \in \mathbb{R}^{hw \times hw}$, to eliminate the structural information while preserving the appearance information. This simple approach of replacing the selected self-attention map with an identity matrix $\mathbf{I}\in \mathbb{R}^{hw \times hw}$ can be defined as
\begin{equation}
\label{equ:psa_attention}
\mathrm{PSA}(Q_t, K_t, V_t) = 
\mathbf{I}V_t=
V_t,
\end{equation}
where we call perturbed self-attention (PSA). More ablation studies on perturbing a self-attention map can be found in the Appendix~\ref{sec:sup:perturbation-ablations}.

\vspace{5pt}

\begin{wrapfigure}{r}{0.55\textwidth}
\vspace{-45pt}
    \begin{minipage}[t]{.55\textwidth}
      \begin{algorithm}[H]
        \caption{Sampling with PAG}
        \label{alg:pag}
        \begin{algorithmic}
            \State \( \mathbf{Model}(x_t), \mathbf{Model'}(x_t):
            \)
            \State \( \textrm{Diffusion model with self-attention and }\) 
            \State \( \textrm{perturbed self-attention (PSA), respectively.} \)
            \State \(s \textrm{: guidance scale,}\, \Sigma_t \textrm{: variance}\)
            \State \( x_T \sim \mathcal{N}(0, I) \)
            \For{\( t \) in \( T, T-1, ..., 1 \)}
                \State \( {\epsilon_t} \leftarrow  \mathbf{Model}(x_t), {\hat{\epsilon_t}} \leftarrow \mathbf{Model'}(x_t) \)
                \State \( \tilde{\epsilon}_t \leftarrow \epsilon_t + s(\epsilon_t - \hat{{\epsilon_t}}) \) \Comment{Eq.~\ref{eq:PAG-derivation}}
                \State \( x_{t-1} \sim \mathcal{N}(\frac{1}{\sqrt{\overline{\alpha}_t}} (x_t - \frac{1-\alpha_t}{\sqrt{1-\overline{\alpha}_t}}\tilde{\epsilon}_t), \Sigma_t) \)\Comment{Eq.~\ref{eq:ddpm_reverse_onestep}}
            \EndFor
            \State \textbf{return} \( x_0 \)
        \end{algorithmic}
      \end{algorithm}
    \end{minipage}
\vspace{-30pt}
\end{wrapfigure}

By using $\mathrm{SA}$ and $\mathrm{PSA}$ module, we implement $\epsilon_{\theta}(x_t)$ and ${\hat{\epsilon}}_{\theta}(x_t)$, respectively. Fig.~\ref{fig:overall} illustrates the overall pipeline of our method, dubbed \textbf{Perturbed-Attention Guidance (PAG)}, as a special case of perturbation guidance. The input image $x_t$ is fed into $\epsilon_{\theta}(\cdot)$ and ${\hat{\epsilon}}_{\theta}(\cdot)$ and the output of the two networks are linearly combined to get the final noise prediction $\tilde{\epsilon}_\theta(x_t)$ as in Eq.~\ref{eq:PAG-derivation}. The pseudo-code is provided in Alg.~\ref{alg:pag}. Note that in our general perturbation guidance framework, the perturbation is not limited to PSA and can be replaced with other strategies. We provide several such examples in the ablation study (see Appendix~\ref{sec:sup:perturbation-ablations}).

\vspace{-5pt}
\subsection{Analysis on PAG}

In this section, we explore why our guidance method is effective. 
Fig.~\ref{fig:pag_path_visualization} shows the sampling process using PAG, with each row (except the last) depicting $\hat{x}_0$ at each timestep using the original epsilon prediction ${\epsilon}_{\theta}(x_t)$, the perturbed epsilon prediction $\hat{\epsilon}_{\theta}(x_t)$, and the guided epsilon $\tilde{\epsilon}_{\theta}(x_t)$. The last row in (b) shows the guidance signal $\hat{\Delta}_t = {\epsilon}_{\theta}(x_t) - \hat{\epsilon}_{\theta}(x_t)$. This figure highlights how our guidance term provides semantic cues. The red rectangle in Fig.~\ref{fig:pag_path_visualization} shows that the perturbed prediction (row 3 in (b)) misses key features like eyes, nose, and tongue due to a lack of global structure understanding. The difference $\hat{\Delta}_t$ focuses on these missing features (row 4 in (b)). Adding $\hat{\Delta}_t$ to the original prediction $\epsilon_\theta(x_t)$ strengthens the sample's structure, as shown in the first row of (b) in Fig.~\ref{fig:pag_path_visualization}. 

More analysis is in Appendix~\ref{sec:sup:cfg-further-analysis} and ~\ref{sec:sup:complementarity-between-pag-sag}. We also visualize CFG~\cite{ho2022classifier} in Stable Diffusion in Fig.~\ref{fig:cfg-ours-visualization}, showing how CFG uses an undesirable sampling path in the unconditional generation to enhance class conditioning. We also discuss the theoretical explanation for why replacing the attention map with an identity matrix is highly effective in Appendix~\ref{sec:sup:theorticial-insights}, drawing on the recent connection between the transformer's self-attention and Hopfield networks.

\section{Experiments}

\begin{table}[!t]
    \centering
    \captionsetup{skip=5pt}
    \setlength{\tabcolsep}{8pt} 
    \caption{\textbf{Quantitative results on ADM~\cite{dhariwal2021diffusion}}. The best values are in bold.}
    \scriptsize 
    \begin{tabular}{c|c|cccc}
        \toprule
        Model & Guidance & FID $\downarrow$ & IS $\uparrow$ & Precision $\uparrow$ & Recall $\uparrow$ \\
        \midrule
        \multirow{3}{*}{\parbox{2.7cm}{\centering ImageNet 256\(\times\)256 \\ Unconditional}} & \ding{55} & 26.21 & 39.70 & 0.61 & \textbf{0.63} \\
        & SAG & 20.08 & 45.56 & 0.68 & 0.59 \\
        & \textbf{PAG} & \textbf{16.23} & \textbf{88.53} & \textbf{0.82} & 0.51 \\
        \midrule
        \multirow{3}{*}{\parbox{2.7cm}{\centering ImageNet 256\(\times\)256 \\ Conditional}} & \ding{55} & 10.94 & 100.98 & 0.69 & 0.63 \\
        & SAG & 9.41 & 104.79 & \textbf{0.70} & 0.62 \\
        & \textbf{PAG} & \textbf{6.32} & \textbf{338.02} & 0.51 & \textbf{0.82} \\
        \bottomrule
    \end{tabular}
    \label{table:adm-quan}
\end{table}

\subsection{Experimental and Implementation Details}

Our work utilizes pretrained models, including ADM~\cite{dhariwal2021diffusion}, Stable Diffusion 1.5~\cite{rombach2022high}, and SDXL~\cite{podell2023sdxl}. We accessed all necessary weights from their publicly available repositories and used the same evaluation metrics as in ADM~\cite{dhariwal2021diffusion}. For additional experimental details, please refer to Appendix~\ref{sec:sup:implementation-details}.

\vspace{-10pt}
\subsection{Pixel-Level Diffusion Models}
\begin{figure}[t]
    \centering
    \captionsetup{skip=5pt}
    \includegraphics[width=\linewidth]{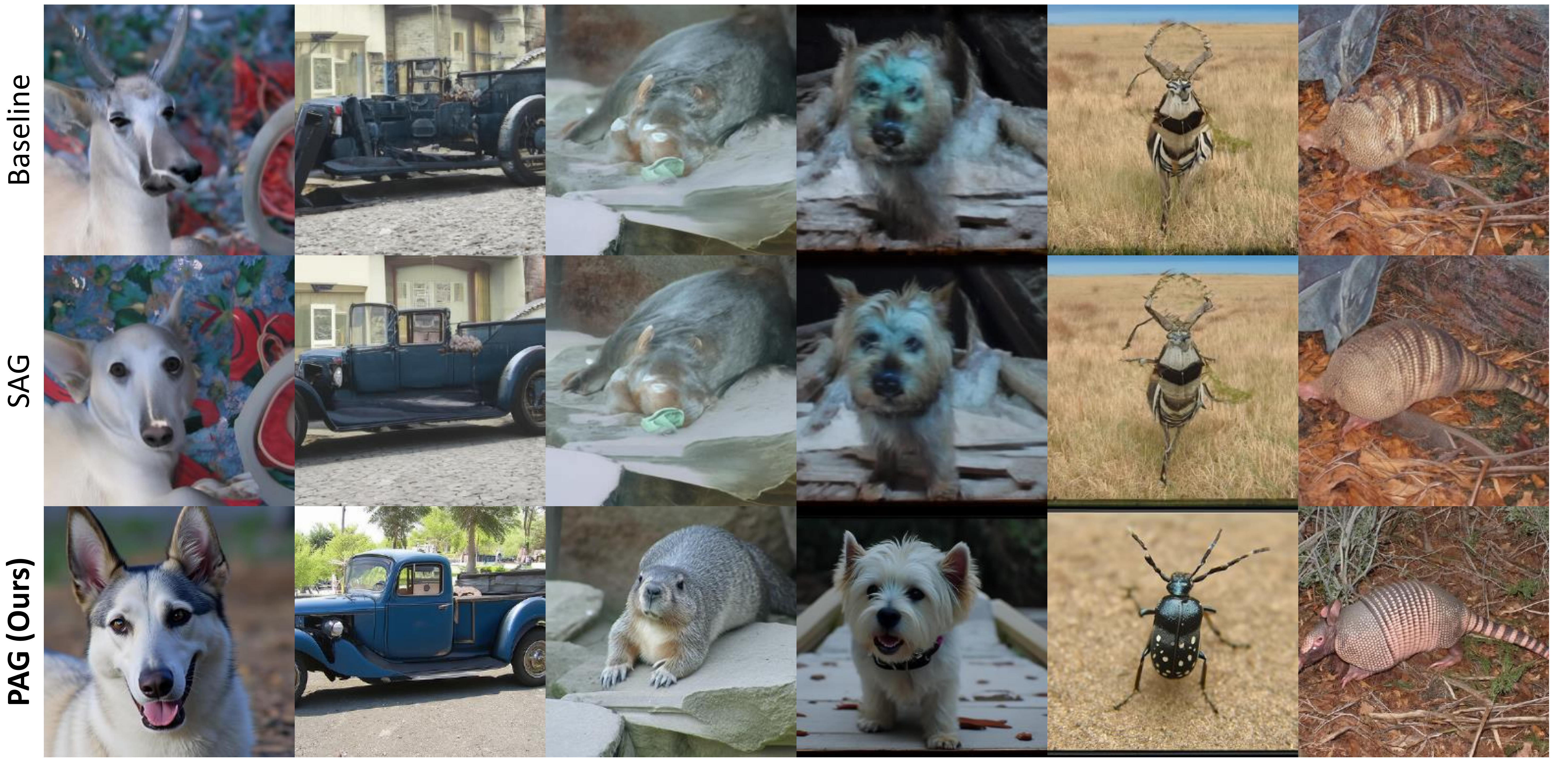} 
    \caption{\textbf{Qualitative comparison between SAG~\cite{hong2023improving} and PAG}. Images are sampled from the ImageNet 256\(\times\)256 unconditional model using the same seed sequence. Compared to samples guided by SAG, those guided by PAG exhibit significantly improved semantic structures with artifacts removed.}
\label{fig:sagours}
\end{figure}

With pretrained ADM~\cite{dhariwal2021diffusion}, we generates 50K samples on ImageNet~\cite{deng2009imagenet} 256$\times$256 to evaluate metrics. In Table~\ref{table:adm-quan}, we compare ADM~\cite{dhariwal2021diffusion} with SAG~\cite{hong2023improving} and PAG in both conditional and unconditional generation. Table~\ref{table:adm-quan} shows that ADM~\cite{dhariwal2021diffusion} with PAG outperforms the others with large margin in FID~\cite{heusel2017gans}, IS~\cite{salimans2016improved}. The contrastive patterns of Improved Recall and Precision~\cite{kynkaanniemi2019improved} in unconditional and conditional generation in Table~\ref{table:adm-quan} are attributed to the trade-off between fidelity and diversity~\cite{dhariwal2021diffusion, ho2022classifier, hong2023improving}. Despite this trade-off, the samples illustrated in Fig.~\ref{fig:sagours} exhibit significant enhancements in quality, demonstrating PAG's capability to rectify the diffusion sampling path leveraging perturbed self-attention. A qualitative comparison with SAG~\cite{hong2023improving} is also presented in Fig.~\ref{fig:sagours}. For further exploration, additional samples from ADM~\cite{dhariwal2021diffusion} are available in Appendix~\ref{sec:sup:adm_qual}.

\vspace{5pt}

\vspace{-5pt}
\begin{figure}[!t]
    \centering
     \makebox[\textwidth]{\includegraphics[width=1.0\textwidth]{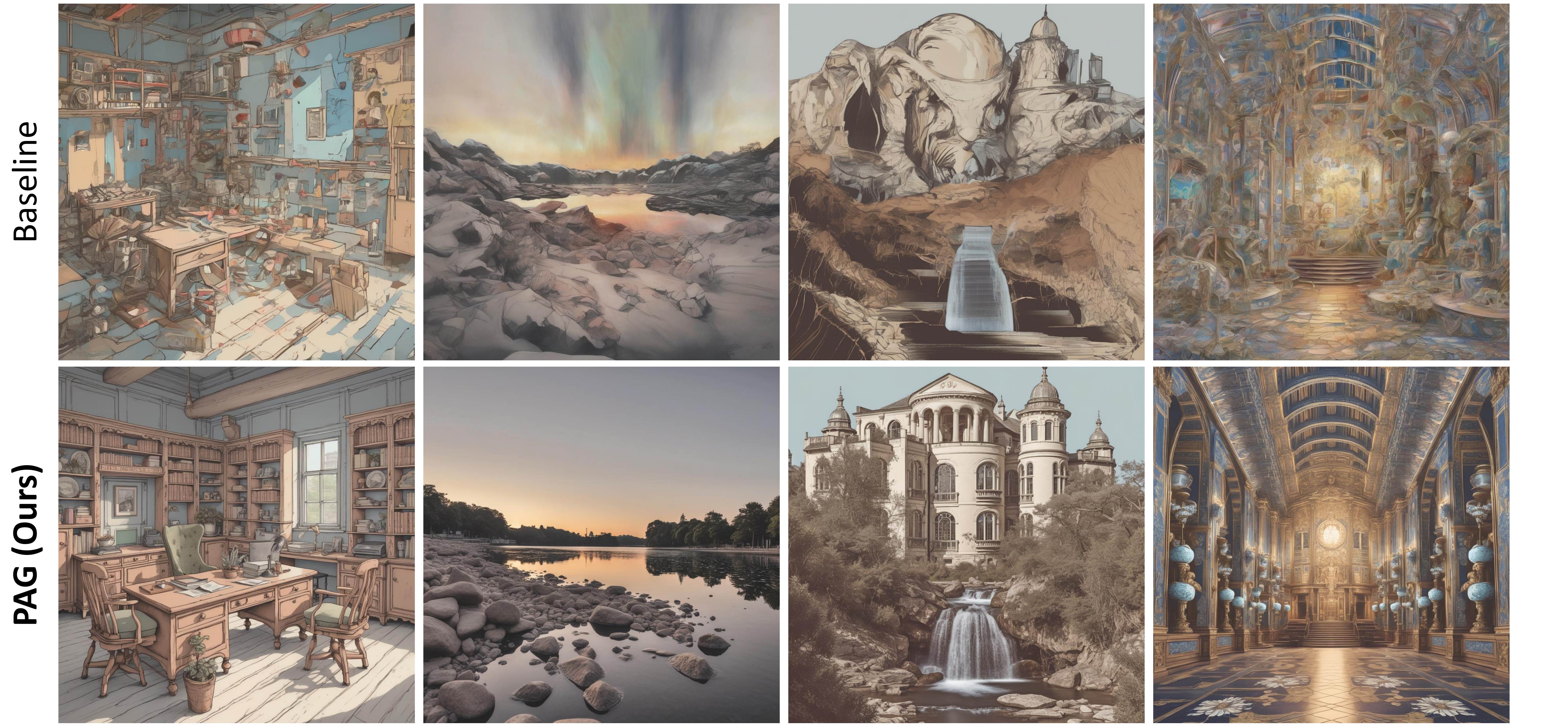}}
     \vspace{-15pt}
    \caption{\textbf{Unconditional generation samples w/o and w/ PAG.} 
    Figures display sampled images from Stable Diffusion XL~\cite{podell2023sdxl}. Each set of images shows sampling without  (\textbf{Top}) and with (\textbf{Bottom}) PAG. Samples guided by PAG appear high perceptual quality and demonstrate semantically coherent structures.}
    \vspace{-10pt}
    \label{fig:main_qual}
\end{figure}

\vspace{-10pt}
\subsection{Latent Diffusion Models}
\begin{table}[!t]
    \centering
    \caption{\textbf{Quantitative results on Stable Diffusion~\cite{rombach2022high}}. The results were obtained using Stable Diffusion v1.5. Sampling was conducted for each with 30K images, and the results were measured accordingly. For text-to-image tasks, 30k prompts were randomly selected from the MS-COCO 2014 validation set~\cite{lin2014microsoft}.}
    \captionsetup{skip=5pt}
    \setlength{\tabcolsep}{6pt} 
    \scriptsize 
    \begin{tabular}{c|c|cc|cc}
        \toprule
        Type & Condition &  \textbf{PAG} & CFG & FID $\downarrow$ & IS $\uparrow$ \\
        \midrule
        \multirow{2}{*}{Unconditional} & \multirow{2}{*}{\ding{55}} & \ding{55} & - & 53.13 & 16.26 \\
        & & \ding{51} & - & \textbf{47.57} & \textbf{21.38} \\
        \midrule
        \multirow{4}{*}{Text-to-Image} & \multirow{4}{*}{\ding{51}} & \ding{55} & \ding{55} & 25.20 & 22.97 \\
        & &  \ding{55} & \ding{51} & 15.00 & \textbf{40.43} \\
        & & \ding{51} & \ding{55} & 10.08 & 33.02 \\
        & & \ding{51} & \ding{51} & \textbf{8.73} & 36.99 \\
        \bottomrule
    \end{tabular}
    \label{table:sdquan}
    \vspace{-5pt}
\end{table}

\vspace{-5pt}
\subsubsection{Unconditional generation on Stable Diffusion.}
We further explored the application of our guidance to Stable Diffusion~\cite{rombach2022high}. In the ``Unconditional'' part of Table~\ref{table:sdquan}, we compared the baseline without PAG to that with PAG for unconditional generation without prompts. The use of PAG resulted in improved FID~\cite{heusel2017gans} and IS~\cite{salimans2016improved}. Samples from Stable Diffusion's unconditional generation with and without PAG are presented in the right column of Fig.~\ref{fig:main_qual} and in the top row of Fig.~\ref{fig:teaser}. Without PAG, the majority of images tend to exhibit semantically unusual structures or lower quality. In contrast, the application of PAG leads to the generation of geometrically coherent objects or scenes, significantly enhancing the visual quality of the samples compared to the baseline.

\vspace{-10pt}
\subsubsection{Text-to-image synthesis on Stable Diffusion.}

Results for text-to-image generation using prompts are presented in the ``Text-to-Image'' part of Table~\ref{table:sdquan}. In this case, since CFG~\cite{ho2022classifier} can be utilized, we conducted sampling in four different scenarios: without applying guidance as a baseline, using CFG, using PAG, and combining both guidance methods with an appropriate scale.
\begin{figure*}[t]
    \centering
    \captionsetup{skip=5pt}
    \includegraphics[width=\linewidth]{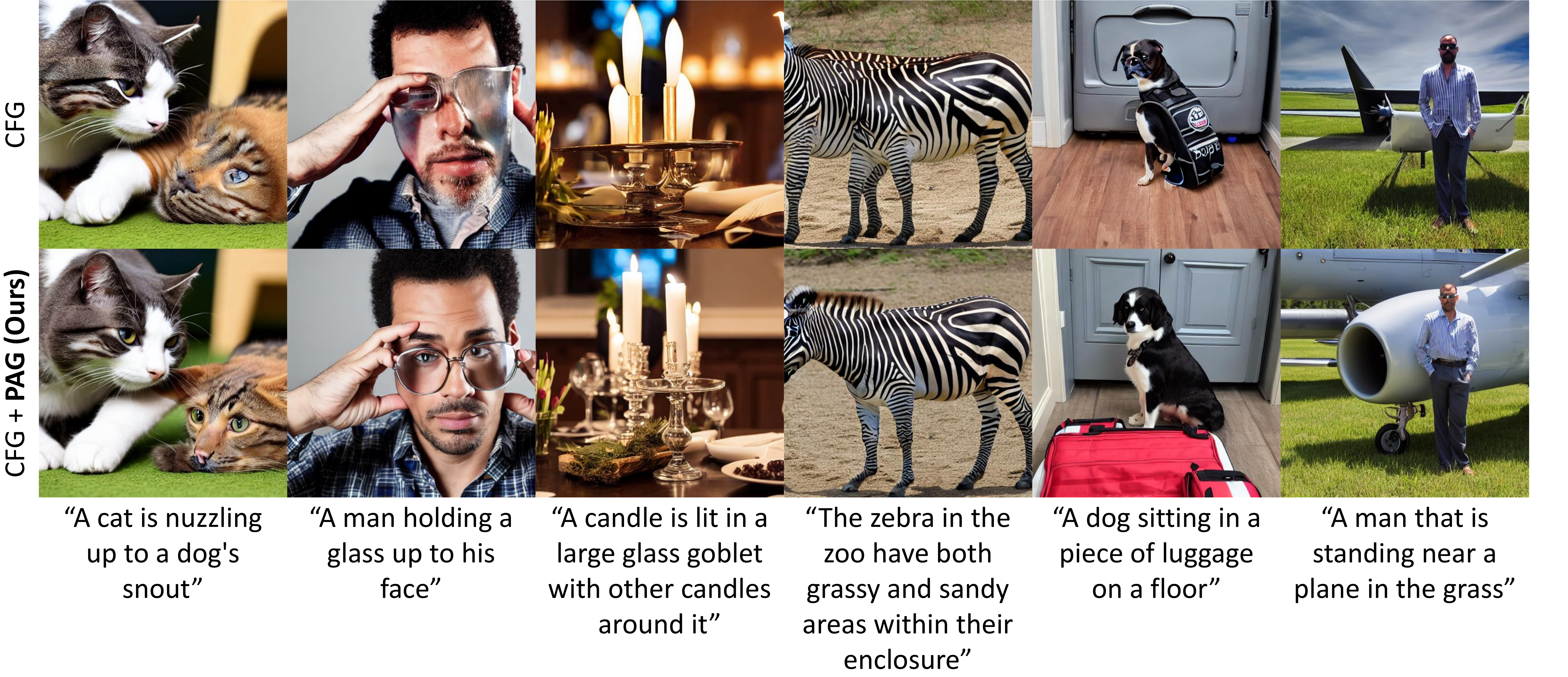} 
    \caption{\textbf{Qualitative comparison between CFG~\cite{ho2022classifier} and CFG + PAG}. Compared to using CFG alone, incorporating PAG alongside CFG noticeably improves the semantic coherence of the structures within the samples. This combination effectively rectifies errors in existing samples, such as adding a missing eye to a cat or eliminating extra legs from a zebra.}
\label{fig:cfgours}
\vspace{-10pt}
\end{figure*}

Interestingly, combining PAG and CFG~\cite{ho2022classifier} with an appropriate scale leads to a significant improvement in the FID  of the generated images. Fig.~\ref{fig:cfgours} offers a qualitative comparison between samples produced using solely CFG and those generated with both guidance methods. The synergy of CFG's effectiveness in aligning images with text prompts and PAG's enhancement of structural information culminates in visually more appealing images when these methods are applied together. Further analysis on the complementarity between PAG and CFG is provided in the Appendix~\ref{sec:sup:complementarity-between-pag-sag}.

\begin{wraptable}{r}{0.4\textwidth}
\vspace{-30pt}
    \footnotesize
    \caption{\textbf{Diversity comparison in samples generated by CFG~\cite{ho2022classifier} and PAG.}}
    \centering
    \captionsetup{skip=5pt}
    \setlength{\tabcolsep}{9pt}
    \scriptsize
    \vspace{5pt}
    \begin{tabular}{ccc}
        \toprule
         & IS $\uparrow$ & LPIPS $\uparrow$ \\
        \midrule
        CFG & 1.82 & 0.64 \\
         \textbf{PAG} & \textbf{2.32} & \textbf{0.68} \\
        \bottomrule
    \end{tabular}
    \label{tab:diversity}
\vspace{-20pt}
\end{wraptable}



To examine the trade-off between sample quality and diversity when using CFG, we initially define per-prompt diversity as ``\textit{the capacity to generate a variety of samples for a given prompt}''. In text-to-image synthesis, this involves generating multiple images from different latents for a single prompt, forming a batch of generated samples. Assessing metrics on such a batch may not effectively measure per-prompt diversity. Thus, to compare the per-prompt diversity of CFG and PAG, we conduct samplings using various latents for a single prompt. For this comparison, the Inception Score (IS)~\cite{salimans2016improved} is calculated over 1000 generated samples, and the LPIPS~\cite{zhang2018unreasonable} metric is averaged across pairwise comparisons of 100 samples (yielding 4950 pairs). The values presented in Table~\ref{tab:diversity} are averages from experiments conducted on 20 prompts, chosen not by selection but by using the first 20 prompts based on the IDs from the MS-COCO 2014 validation set~\cite{lin2014microsoft}. Further samples from Stable Diffusion are available in Appendix~\ref{sec:sup:sd-qual} for additional reference.

\subsection{Downstream Tasks}

\subsubsection{Inverse problems.} Inverse problem is one of the major tasks in the unconditional generation, which aims to restore $x$ from the noisy measurement $y = \mathcal{A}(x) + n$, where $\mathcal{A}(\cdot)$ denotes measurement operator (\eg, Gaussian blur) and $n$ represents a vector of noise. In this task, where text prompts are not available, PAG can operate properly to improve sample quality without prompts, whereas it is challenging to utilize existing guidance methods that require prompts. We test PAG using a subset of FFHQ~\cite{ffhq} 256$\times$256 on PSLD~\cite{rout2024solving} which leverages DPS~\cite{chung2022diffusion} and LDM~\cite{rombach2022high} to solve linear inverse problems. More details about experimental settings are provided in Appendix~\ref{sec:sup:implementation-details}. 

\begin{table}[!t]
  \centering
  \caption{\textbf{Quantitative results of PSLD~\cite{rout2024solving} on FFHQ~\cite{ffhq} 256$\times$256 1K validation set.}}
  \captionsetup{skip=5pt}
  \setlength{\tabcolsep}{5pt}
  \scriptsize
  \resizebox{1.0\textwidth}{!}{
  \begin{tabular}{lcccccccc}
    \toprule
    {} & \multicolumn{2}{c}{Box Inpainting} & \multicolumn{2}{c}{{\color{black}SR ($8\times$)}}    & \multicolumn{2}{c}{{\color{black}Gaussian Deblur}}    &
    \multicolumn{2}{c}{{\color{black}Motion Deblur}}\\
    \cmidrule(r){2-3}   \cmidrule(r){4-5}   \cmidrule(r){6-7} \cmidrule(r){8-9}
    Method & FID $\downarrow$ & LPIPS $\downarrow$ & FID $\downarrow$ & LPIPS $\downarrow$ & FID $\downarrow$ & LPIPS $\downarrow$ & FID $\downarrow$ & LPIPS $\downarrow$\\
    \midrule
    PSLD & 43.11 &  0.167  & {42.98}& {0.360}& 41.53 & \textbf{0.221} & {93.39} & {0.450} \\
    PSLD $+$ \textbf{PAG (Ours)}   & \textbf{21.13} & \textbf{0.149}   & \textbf{38.57} &\textbf{0.354}& \textbf{37.08}  & 0.343 & \textbf{40.26}  &  \textbf{0.397} \\
    \bottomrule
  \end{tabular}%
  }
  \label{tab:ffhq-sd-fid-lpips}
\vspace{-5pt}
\end{table} 

\begin{figure*}[!t]
    \centering
    \includegraphics[width=1.0\linewidth]{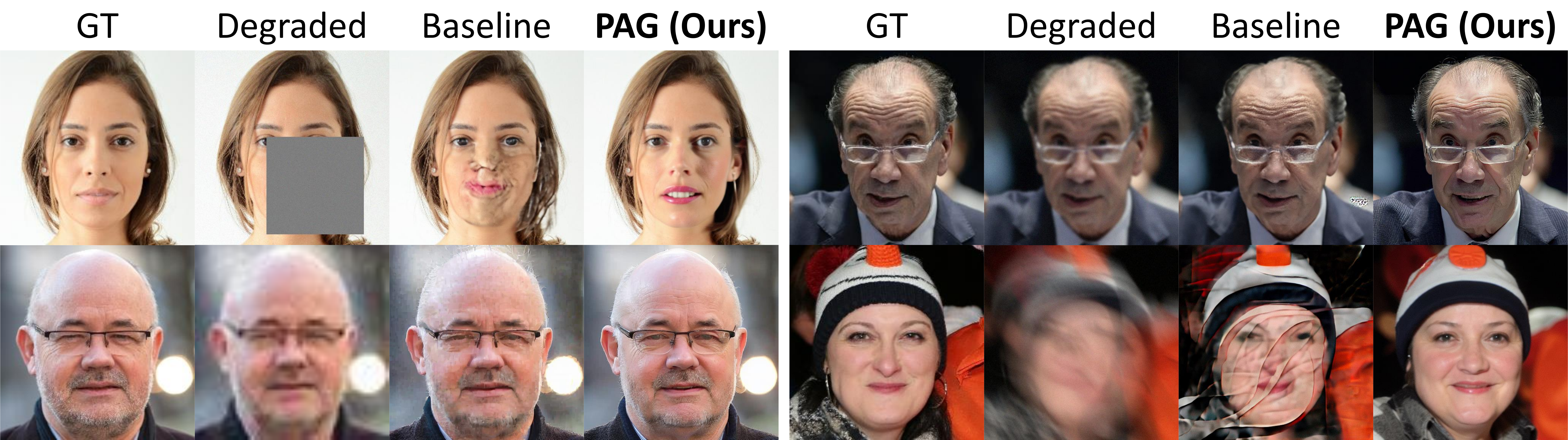} 
    \vspace{-15pt}
    \caption{\textbf{Qualitative results of PSLD~\cite{rout2024solving} with our PAG on FFHQ~\cite{ffhq} dataset.} \textbf{Left Top:} Box inpainting. \textbf{Left Bottom:} Super-resolution ($\times$8). \textbf{Right Top:} Gaussian deblur. \textbf{Right Bottom:} Motion deblur. Using PAG leads to the removal of artifacts and blurriness, resulting in more realistic restorations.}
\label{fig:psld-qual}
\vspace{-10pt}
\end{figure*}

Table~\ref{tab:ffhq-sd-fid-lpips} shows the quantitative results of PSLD with PAG on box inpainting, super-resolution ($\times$8), gaussian deblur, and motion deblur. The performance of PSLD with PAG outperforms all of the tasks in FID~\cite{heusel2017gans}, and mostly in LPIPS~\cite{zhang2018unreasonable}. Fig.~\ref{fig:psld-qual} highlights a considerable improvement in the quality of restored samples using PAG, with a notable reduction of artifacts present in the original method. Importantly, PAG can be adopted to any other restoration model based on diffusion models, shown in Appendix~\ref{sec:sup:additioanl_applications}.

\vspace{-10pt}

\subsubsection{ControlNet.}
\begin{figure*}[!t]
    \centering
    \includegraphics[width=0.9\textwidth]{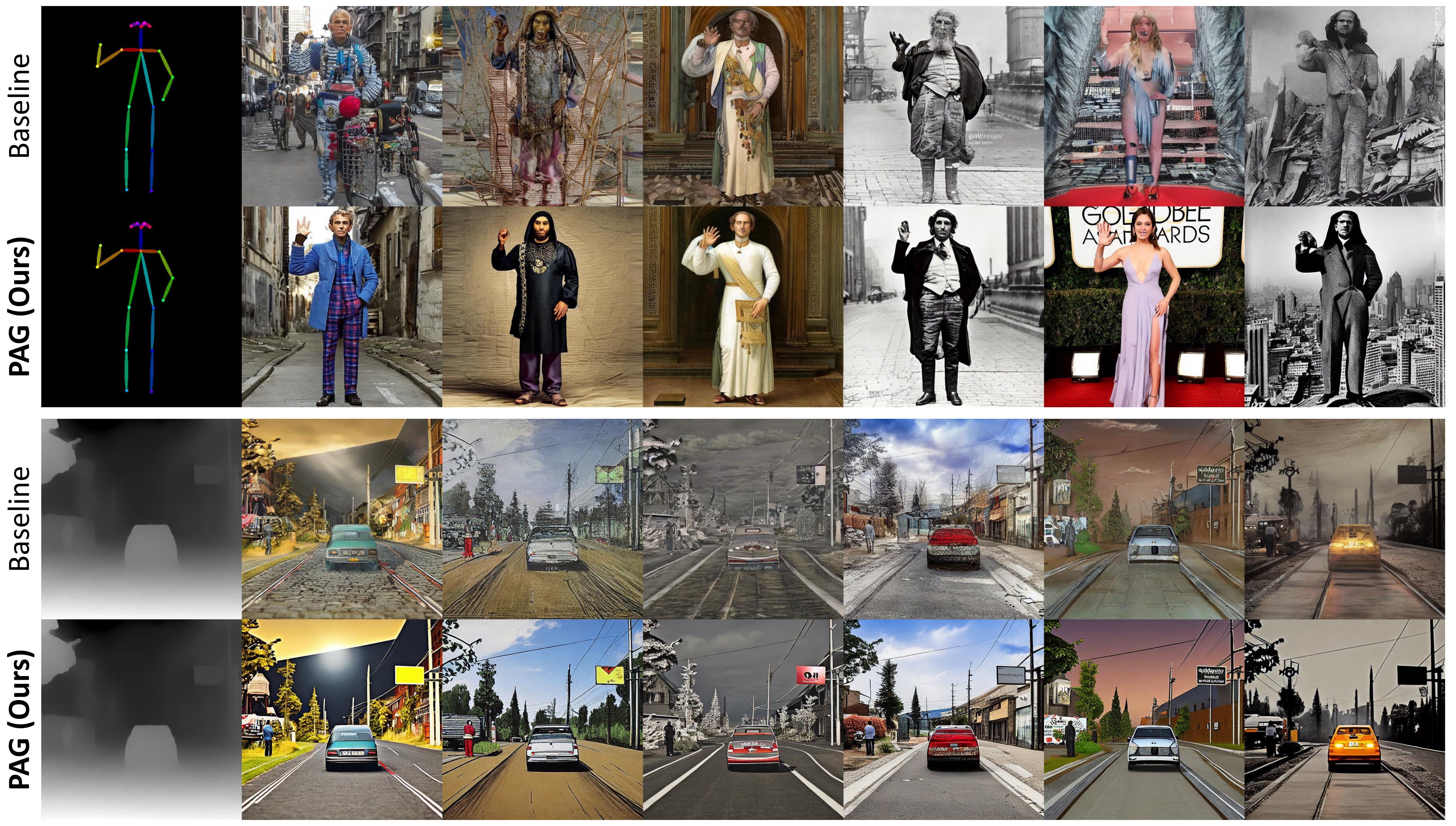}
    \vspace{-5pt}
    \caption{\textbf{ControlNet~\cite{zhang2023adding} sample images conditioned by pose and depth without text prompt}. Samples guided by PAG appear more realistic, exhibiting fewer artifacts and semantically coherent structure.}
    \vspace{-10pt}
    \label{fig:controlnet}
\end{figure*}
ControlNet~\cite{zhang2023adding}, a method for introducing spatial conditioning controls in pretrained text-to-image diffusion models, sometimes struggles to produce high-quality samples under unconditional generation scenarios, particularly when the spatial control signal is sparse, such as pose conditions. However, as demonstrated in Fig.~\ref{fig:controlnet}, PAG enhances sample quality in these instances. This enables the generation of plausible samples conditioned solely on spatial information without the need for specific prompts, making it useful for crafting training datasets tailored to specific goals and allowing artists to test diverse, imaginative works without relying on detailed prompts.

\vspace{-10pt}
\subsection{Ablation Studies}
\label{sec:ablations}

We provide ablations studies on self-attention perturbation strategy and effects of guidance scales on qualitative and quantitative results on Appendix~\ref{sec:sup:ablations}.

PAG, like CFG, can parallelize the two denoising passes in Fig.~\ref{fig:overall} by duplicating the input of the Diffusion U-Net and making a batch. As a result, the computational cost is nearly identical to that of CFG, and details on time and memory consumption are provided in the Appendix~\ref{sec:sup:commputational_cost}.
\vspace{-5pt}
\section{Conclusion and Discussion}
\vspace{-5pt}
In this work, we propose a novel guidance framework, termed \textbf{perturbation guidance}, to improve sample quality by guiding the sampling trajectory away from a “perturbed” forward pass. Building upon this idea, we introduce a specific implementation, \textbf{Perturbed-Attention Guidance (PAG)}, which leverages structural perturbations to enhance image generation. Starting with an elucidation of how CFG~\cite{ho2022classifier} refines sample realism, by replacing the diffusion U-Net's self-attention map with an identity matrix, we effectively guide the generation process away from structural degradation. Crucially, PAG achieves superior sample quality in both conditional and unconditional settings, requiring no additional training or external modules.  Furthermore, we demonstrate the versatility of PAG by showing its effectiveness in downstream tasks such as image restoration. 

In later studies, several perturbation-based and weak-model-based guidance methods~\cite{hong2024smoothed,karras2024guiding,hyung2025spatiotemporal,li2024self} have been proposed. Karras et al.\cite{karras2024guiding} illustrate how guidance with a ``bad'' model can be effective using toy examples, and suggest employing an under-trained or capacity-limited model for this purpose. Hong et al.\cite{hong2024smoothed} propose applying blur to self-attention maps to mitigate overly strong perturbations, supported by theoretical analysis. Hyung et al.~\cite{hyung2025spatiotemporal} explore the use of perturbed self-attention and layer-skip perturbations in video diffusion models.

We believe that our exploration enriches the understanding of sampling guidance methods and diffusion models, and illuminates the applicability of unconditional diffusion models, liberating diffusion models from reliance on text prompts and CFG.
\section*{Acknowledgements}
This research was supported by the MSIT, Korea (IITP-2024-2020-0-01819, RS-2023-00227592), Culture, Sports, and Tourism R\&D Program through the Korea Creative Content Agency grant funded by the Ministry of Culture, Sports and Tourism (Research on neural watermark technology for copyright protection of generative AI 3D content, RS-2024-00348469, RS-2024-00333068) and National Research Foundation of Korea (RS-2024-00346597). Thank you to Susung Hong for providing feedback on our research and manuscript.
%
%


\bibliographystyle{splncs04}
\bibliography{main}

\newpage
\appendix

\begin{center}
\Large
\textbf{Appendix}
\end{center}

In the following, we provide detailed information on the implementation of all experiments (Sec.~\ref{sec:sup:implementation-details}), along with a broader range of qualitative results from samples enhanced by the Perturbed-Attention Guidance (PAG), which includes human evaluations and results from downstream tasks (Sec.~\ref{sec:sup:more_qual}). Additionally, we highlight intriguing applications where PAG proves beneficial, such as DPS~\cite{chung2022diffusion}, the Stable Diffusion~\cite{rombach2022high} super-resolution/inpaint pipeline, and text-to-3D~\cite{poole2022dreamfusion} (Sec.~\ref{sec:sup:additioanl_applications}). We also present ablation studies focusing on perturbation methods and layer selection (Sec.~\ref{sec:sup:ablations}). A comprehensive analysis of CFG and PAG, including the dynamics of using CFG and PAG concurrently, is provided (Sec.~\ref{sec:sup:discussion}). Discussion on limitations is also included (Sec.~\ref{sec:sup:limitations}).

\section{Implementation Details}
\label{sec:sup:implementation-details}
In this section, we provide detailed descriptions of the implementation and hyperparameter settings for all experiments in the paper.

\subsection{Experiments on ADM}
\subsubsection{Quantitative results.}
For the main quantitative result presented in the main paper involving the ADM~\cite{dhariwal2021diffusion} ImageNet~\cite{deng2009imagenet} 256$\times$256 conditional and unconditional models, we utilized the official GitHub repository\footnote{https://github.com/openai/guided-diffusion} of ADM along with its publicly available pretrained weights. Our work builds upon the SAG~\cite{hong2023improving} repository\footnote{https://github.com/KU-CVLAB/Self-Attention-Guidance}, which is derived from the ADM official repository, to ensure precise comparison. We configured the PAG scale $s=1.0$ and defined the perturbation to the self-attention mechanism as substituting $\mathrm{Softmax}({Q_tK^T_t}/{\sqrt{d}}) \in \mathbb{R}^{hw \times hw}$ with an identity matrix $\mathbf{I} \in \mathbb{R}^{hw \times hw}$. Here, $Q_t$, and $K_t$ represent the query and key at timestep $t$ and h, w, and d refer to the height, width, and channel dimensions, respectively. 
The specific layers for applying perturbed self-attention are as follows: \texttt{input\_blocks.14.1}, \texttt{input\_blocks.16.1}, \texttt{input\_blocks.17.1}, \texttt{middle\_block.1} for unconditional models and \texttt{input\_blocks.14.1} for conditional models.  We follow the same evaluation protocol as SAG~\cite{hong2023improving}, utilizing the DDPM sampler with 250 steps and employing the same evaluation code as provided by the official repository of ADM.

\subsubsection{Qualitative results.}
For the qualitative results in the main paper, we configured the PAG scale $s=3.0$. 
This choice of a higher $s$ value stems from our observations in the ablation study on guidance scale. It shows that although sample quality improves with an increasing guidance scale the FID~\cite{heusel2017gans} score worsens. This may be due to the misalignment between FID and human perception~\cite{jayasumana2023rethinking}. Consequently, we increase the guidance scale to prioritize perceived quality improvement.
We applied the same identity matrix substitution and the same layers for perturbed self-attention as in the quantitative experiments.

\subsubsection{Visualization of diffusion sampling path.}
For the visualization of the reverse process in the Fig.~\ref{fig:pag_path_visualization}, we obtain
$\hat{\Delta}_t$ by calculating the absolute value of each channel, computing the channel-wise mean, and clipping outlier values to enhance clarity. The hyperparameters are consistent with those in the qualitative results with ADM~\cite{dhariwal2021diffusion}.

\subsection{Experiments on Stable Diffusion}
\label{sec:sup:experiments-on-stable-diffusion}
\subsubsection{Quantitative results.}
For all the quantitative experiments, we utilized Stable Diffusion v1-5\footnote{https://huggingface.co/runwayml/stable-diffusion-v1-5} implemented based on the pipeline provided by the Diffusers~\cite{von-platen-etal-2022-diffusers}. For the PAG guidance scale, $s=$ 2.0 is used for unconditional generation, while $s=$ 2.5 is used for text-to-image synthesis. In text-to-image synthesis, CFG~\cite{ho2022classifier} was set to the most commonly used value of $w=$ 7.5, and for experiments combining CFG and PAG, $w=$ 2.0 and $s=$ 1.5 were employed. For the diversity comparison in the main paper, $s=$ 4.5 and $w=$ 7.5 were used respectively. In all experiments, perturbed self-attention was applied to the middle layer \texttt{mid\_block.attentions.0.\-transformer\_blocks.0.attn1} of the U-Net, and sample images were generated through DDIM~\cite{song2020denoising} 50 step sampling method.

\subsubsection{Qualitative results.}
Stable Diffusion v1-5 and SDXL\footnote{https://huggingface.co/stabilityai/stable-diffusion-xl-base-1.0} are used for all qualitative generation results. For the main qualitative results, PAG guidance scale $s=$ 4.5 is used. Also, for CFG experiments, CFG guidance scale $w=$ 7.5 was applied, and for the CFG+PAG experiment, $w=$ 6.0 and $s=$ 1.5 were used. We used DDIM sampling~\cite{song2020denoising} with 200 steps for the teaser (Fig.~\ref{fig:teaser}), 50 steps for the main figure (Fig.~\ref{fig:main_qual}), and 25 steps for comparison between CFG and CFG + PAG (Fig.~\ref{fig:cfgours}). Perturbed self-attention was applied to the middle layer \texttt{mid\_block.attentions.0.transformer\_bloc\-ks.0.attn1} of the U-Net in all cases.

\subsubsection{Visualization of diffusion sampling path.}
For the visualization experiment of reverse process in the main figure (Fig.~\ref{fig:cfg-ours-visualization}), CFG~\cite{ho2022classifier} scale $w=$ 7.5 is used, and perturbed self-attention was applied to the middle layer \texttt{mid\_block.attentions.0.\-trans\-former\_blocks.0.attn1}, representing the initial 12 steps of DDIM 25 step sampling.

\subsubsection{Combination of CFG and PAG.}
To apply CFG~\cite{ho2022classifier} and PAG together in text-to-image synthesis, we produced $\Tilde{\epsilon}_{\theta}(x_t,c)$ using the following equation:
\begin{equation}
    \Tilde{\epsilon}_{\theta}(x_t,c) = {\epsilon}_{\theta}(x_t,c) + w ( \epsilon_{\theta}(x_t, c) - \epsilon_{\theta}(x_t,\phi) ) + s ( \epsilon_{\theta}(x_t,c) - \hat{\epsilon}_{\theta}(x_t,c) ),
\end{equation}
where $w$ and $s$ are guidance scale.
These estimations involve adding the deltas of CFG and PAG, each weighted by each guidance scale $w$ and $s$. To achieve this, we computed three estimations, ${\epsilon}_{\theta}(x_t,c)$, ${\epsilon}_{\theta}(x_t,\phi)$, and $\hat{\epsilon}_{\theta}(x_t,c)$ simultaneously, in the denoising U-Net.

\subsection{Experiments with PSLD}
We use Stable Diffusion v1.5 used in PSLD~\cite{rout2024solving}. The measurement operators for inverse problems are from DPS~\cite{chung2022diffusion}, as used in PSLD~\cite{rout2024solving}. PSLD~\cite{rout2024solving} leverages the loss term of DPS~\cite{chung2022diffusion} and further implements the gluing objective to enhance fidelity, multiplied with step size $\eta$ and $\gamma$ respectively for updating gradients. $\eta=1.0$ and $\gamma=0.1$ are used in experiments of PSLD~\cite{rout2024solving} without PAG as same as PSLD~\cite{rout2024solving}. Practically, we find that it is better to use unconditional score $\epsilon_\theta(\boldsymbol{z}_t)$ instead of guided score $\Tilde{\epsilon}_\theta(\boldsymbol{z}_t)$ when predicting $\hat{\boldsymbol{z}}_0$ to update gradients. Furthermore, we conduct more experiments with ImageNet~\cite{deng2009imagenet} dataset, which are provided in Sec.~\ref{sec:psld-imagenet}. All experiments with PSLD~\cite{rout2024solving} use DDIM~\cite{song2020denoising} sampling and all hyperparameters with PAG are in Table~\ref{tab:psld-hyparam}.  Perturbed self-attention is applied to the same layer, \texttt{input\_block.8.1.transformer\_blocks.0.attn1}, for both FFHQ~\cite{ffhq} and ImageNet~\cite{deng2009imagenet} dataset.

\begin{table}[]
\centering
\captionsetup{skip=5pt}
\setlength{\tabcolsep}{2pt}
\resizebox{0.8\textwidth}{!}{%
\begin{tabular}{lllllllll}
\toprule
{} & \multicolumn{4}{c}{\textbf{FFHQ}} & \multicolumn{4}{c}{\textbf{ImageNet}}\\
\cmidrule(r){2-5}
\cmidrule(r){6-9}
{} & Inpaint & SR$\times8$ & Gauss & Motion  & Inpaint & SR$\times8$ & Gauss & Motion \\
$\eta$ & 0.15 & 0.7 & 0.1 & 0.15 & 0.5 & 0.7 & 0.1 & 0.3 \\
$\gamma$ & 0.015 & 0.07 & 0.01 & 0.015 & 0.05 & 0.07 & 0.01 & 0.03 \\
$s$ & 4.0 & 4.0 & 5.0 & 4.0 & 4.0 & 4.0 & 5.0 & 5.0  \\
\bottomrule
\end{tabular}
}
\caption{\textbf{Hyperparameters for PSLD~\cite{rout2024solving} with \textbf{PAG} on FFHQ~\cite{ffhq} dataset and ImageNet~\cite{deng2009imagenet} dataset.} Here, $\eta$ and $\gamma$ are the step size for gradients of PSLD~\cite{rout2024solving} and $s$ is the scale for PAG from Eq.~\ref{eq:PAG-derivation} of main paper.} 
\label{tab:psld-hyparam}
\end{table}

\subsection{Experiments with ControlNet}
For the ControlNet~\cite{zhang2023adding} experiment in Fig. A, Stable Diffusion v1.5 was utilized, implemented based on the ControlNet pipeline from Diffusers. For pose conditional generation, PAG guidance scale 2.5 is used, while for depth conditional generation, 1.0 was employed. Sampling was conducted using the DDIM 50 steps method, and perturbed self-attention was applied to the middle layer \texttt{mid\_block.\-attentions.0.transformer\_blocks.0.attn1} of the U-Net.

\subsection{Ablation Study}
For the ablation study on the guidance scale and perturbation strategy, we generated 5k images using the ADM~\cite{dhariwal2021diffusion} ImageNet 256$\times$256 unconditional model with DDIM 25 step sampling and applied perturbed self-attention to the \texttt{input.13} layer. In the guidance scale ablation, identity matrix replacement was used consistently across other qualitative and quantitative experiments. For qualitative results with varying guidance scales on Stable Diffusion v1.5 (Fig.~\ref{fig:scalequal}), DDIM 50-step sampling was utilized with perturbed self-attention applied to \texttt{mid\_block.attentions.0.transformer\_blocks.0.attn1}, aligning with the approach used for Stable Diffusion qualitative samples in the bottom right of the main qualitative figure.

\subsection{Computational Cost}
\label{sec:sup:commputational_cost}
\begin{table}[!t]
    \caption{\textbf{Comparison of computational costs in Stable Diffusion.}}
    \centering
    \captionsetup{skip=5pt}
    \setlength{\tabcolsep}{10pt}
    \begin{tabular}{ccc}
        \toprule
         & GPU Memory $\downarrow$ & Sampling Speed $\uparrow$ \\
        \midrule
        No Guidance & 3,147 MB & 19.16 iter/s \\
        \midrule
        CFG~\cite{ho2022classifier} & \textbf{3,193 MB} & 12.67 iter/s \\
        PAG & \textbf{3,193 MB} & \textbf{12.68 iter/s} \\
        \bottomrule
    \end{tabular}
    \label{tab:computational}
\end{table}
We measured the computational costs for sampling without guidance, using CFG, and using PAG in Stable Diffusion. We utilized one NVIDIA GeForce RTX 3090 GPU and conducted sampling with one batch. Firstly, we measured GPU memory usage, which appeared to be nearly identical across all three scenarios. Next, we measured the iteration speed in the denoising U-Net, showing that both CFG and PAG exhibited similar sampling speeds, albeit slightly slower when compared to not using guidance..

\newpage

\section{Additional Qualitative Results}
\label{sec:sup:more_qual}

\subsection{ADM Results}
\label{sec:sup:adm_qual}
\begin{figure*}[!h]
    \includegraphics[width=0.96\textwidth]{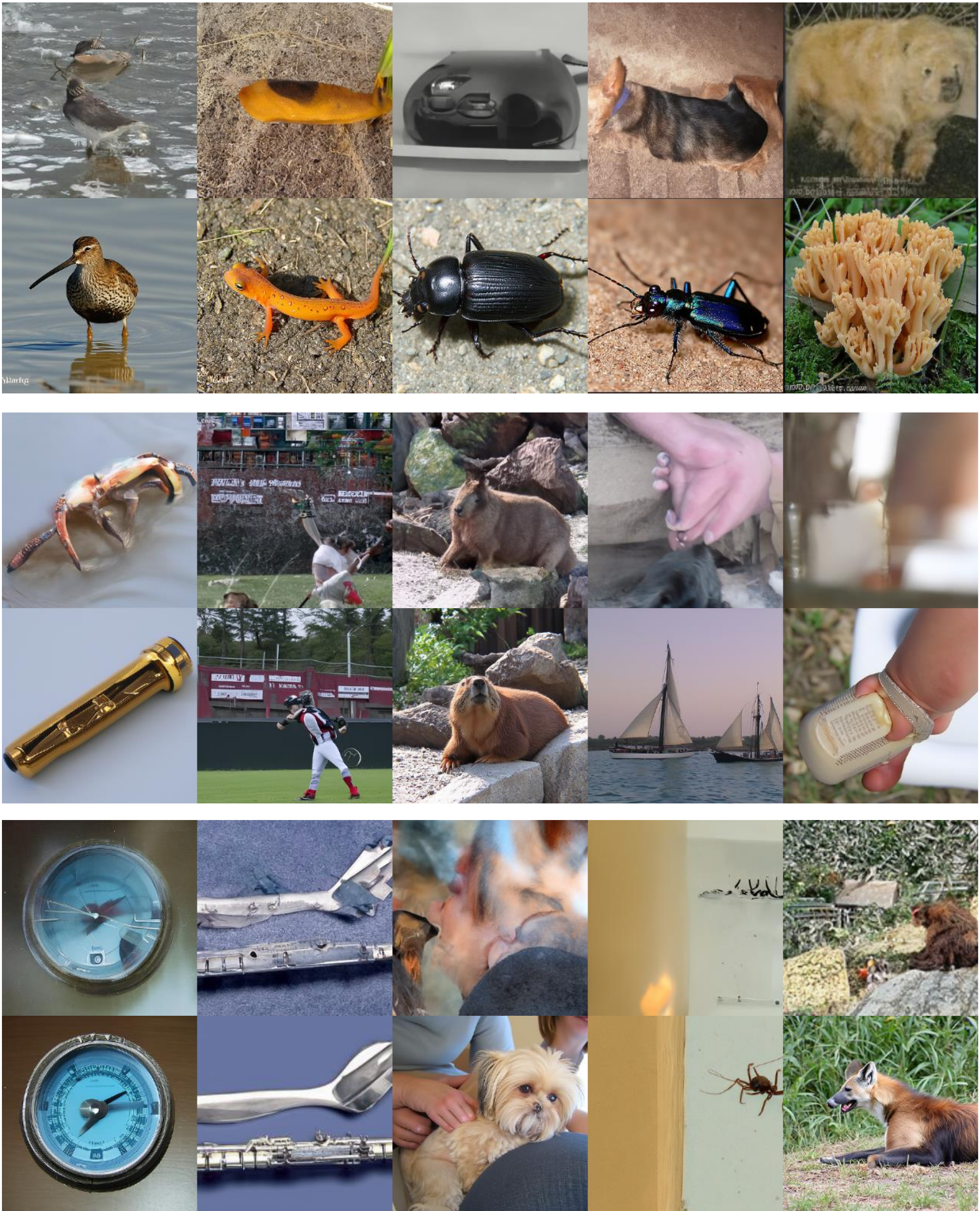}
    \caption{\textbf{Uncurated samples from ADM~\cite{dhariwal2021diffusion} ImageNet 256 \textit{unconditional} model w/o and w/ PAG.} In each image set, the images in the top row are samples without using guidance, and the images in the bottom row are samples using PAG. PAG guidance scale $s=$ 3.0 is used and perturbed layers are following: \texttt{i13},\texttt{i14},\texttt{i16},\texttt{m1}.}
    \label{fig:supple_adm_uncond_1}
    \vspace{-10pt}
\end{figure*}

\newpage
\begin{figure*}[!h]
    \includegraphics[width=0.96\textwidth]{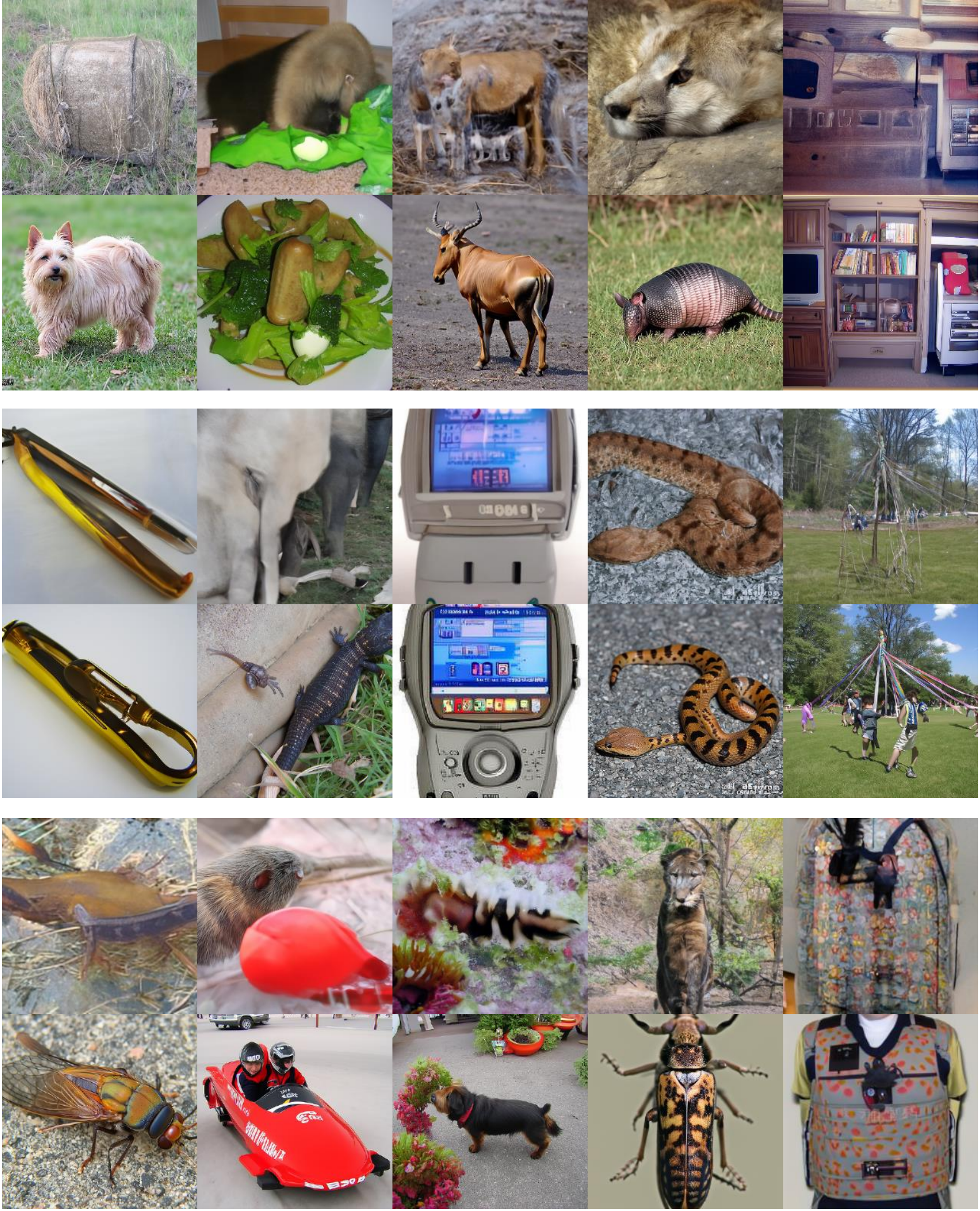}
    \caption{\textbf{Uncurated samples from ADM~\cite{dhariwal2021diffusion} ImageNet 256 \textit{unconditional} model w/o and w/ PAG.} In each image set, the images in the top row are samples without using guidance, and the images in the bottom row are samples using PAG. PAG guidance scale $s=$ 3.0 is used and perturbed layers are following: \texttt{i13},\texttt{i14},\texttt{i16},\texttt{m1}.}
    \label{fig:supple_adm_uncond_2}
    \vspace{-10pt}
\end{figure*}

\begin{figure*}[!h]
    \includegraphics[width=0.96\textwidth]{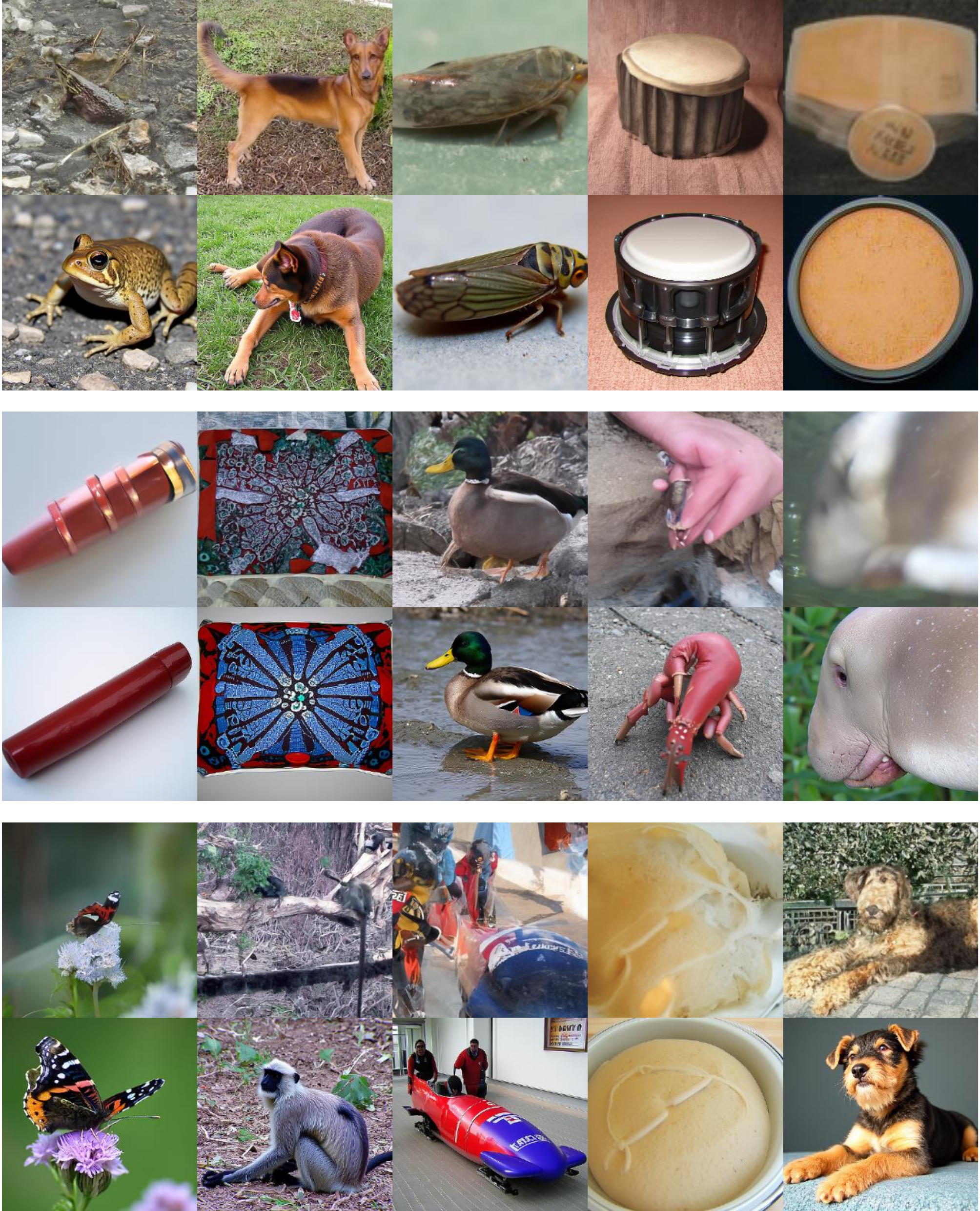}
    \caption{\textbf{Uncurated samples from ADM~\cite{dhariwal2021diffusion} ImageNet 256 \textit{conditional} model w/o and w/ PAG.} In each image set, the images in the top row are samples without using guidance, and the images in the bottom row are samples using PAG. PAG guidance scale $s=$ 3.0 is used and perturbed layers are following: \texttt{i13},\texttt{i14},\texttt{i16},\texttt{m1}.}
    \label{fig:supple_adm_cond_1}
    \vspace{-10pt}
\end{figure*}

\begin{figure*}[!h]
    \includegraphics[width=0.96\textwidth]{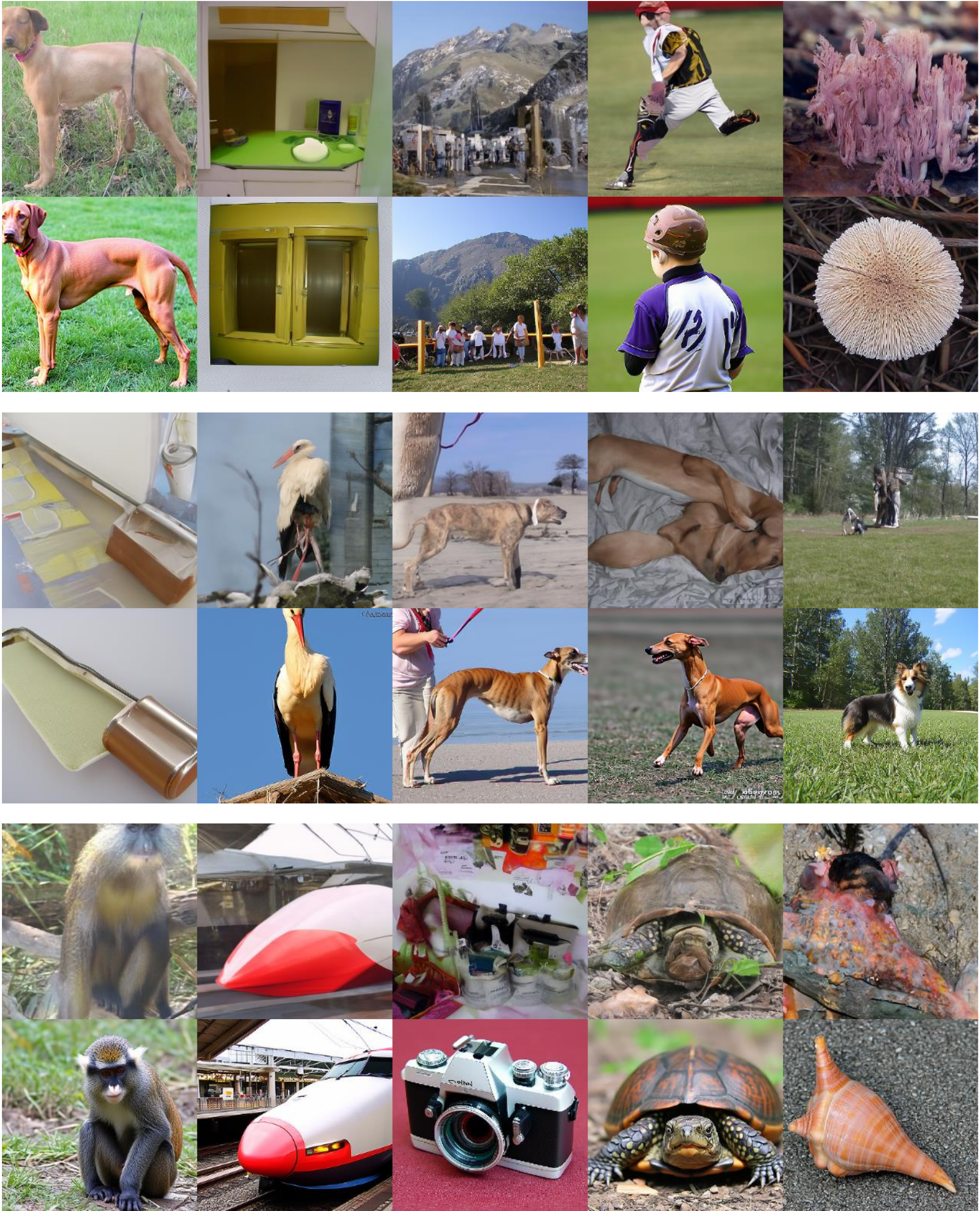}
    \caption{\textbf{Uncurated samples from ADM~\cite{dhariwal2021diffusion} ImageNet 256 \textit{conditional} model w/o and w/ PAG.} In each image set, the images in the top row are samples without using guidance, and the images in the bottom row are samples using PAG. PAG guidance scale $s=$ 3.0 is used and perturbed layers are following: \texttt{i13},\texttt{i14},\texttt{i16},\texttt{m1}.}
    \label{fig:supple_adm_cond_2}
    \vspace{100pt}
\end{figure*}

\clearpage
\subsection{Stable Diffusion Results}
\label{sec:sup:sd-qual}
\begin{figure*}[!h]
    \includegraphics[width=\textwidth]{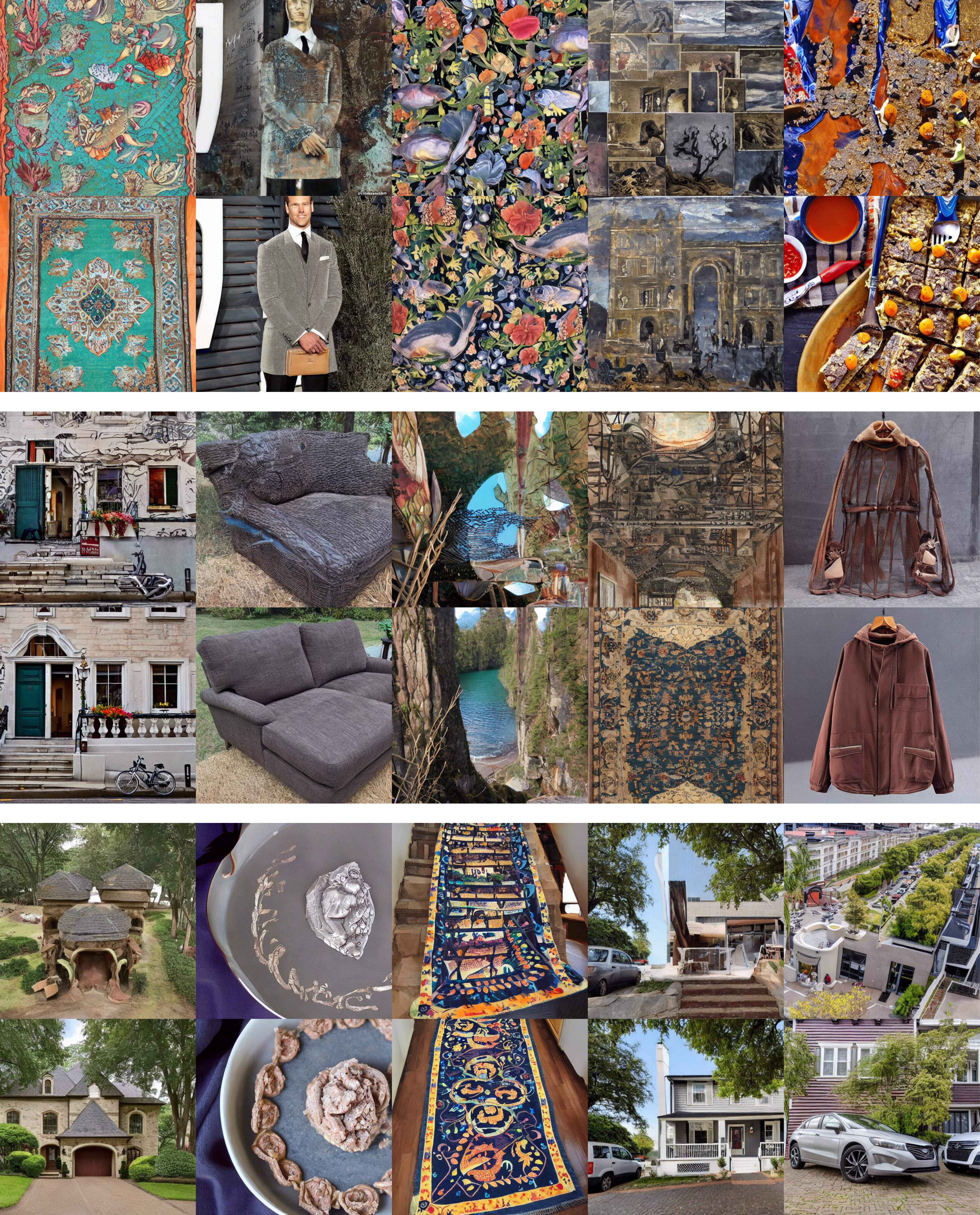}
    \caption{\textbf{Uncurated samples from SD~\cite{rombach2022high} in \textit{unconditional} generation w/o and w/ PAG.} In each image set, the images in the top row are samples without using guidance, and the images in the bottom row are samples using PAG. PAG guidance scale $s=$ 5.0 and perturbed layer \texttt{mid\_block.attentions.0.\-transformer\_blocks.0.attn1} are used.}
    \label{fig:supple_sd_uncond_1}
    \vspace{-10pt}
\end{figure*}

\newpage

\begin{figure*}[!h]
    \includegraphics[width=\textwidth]{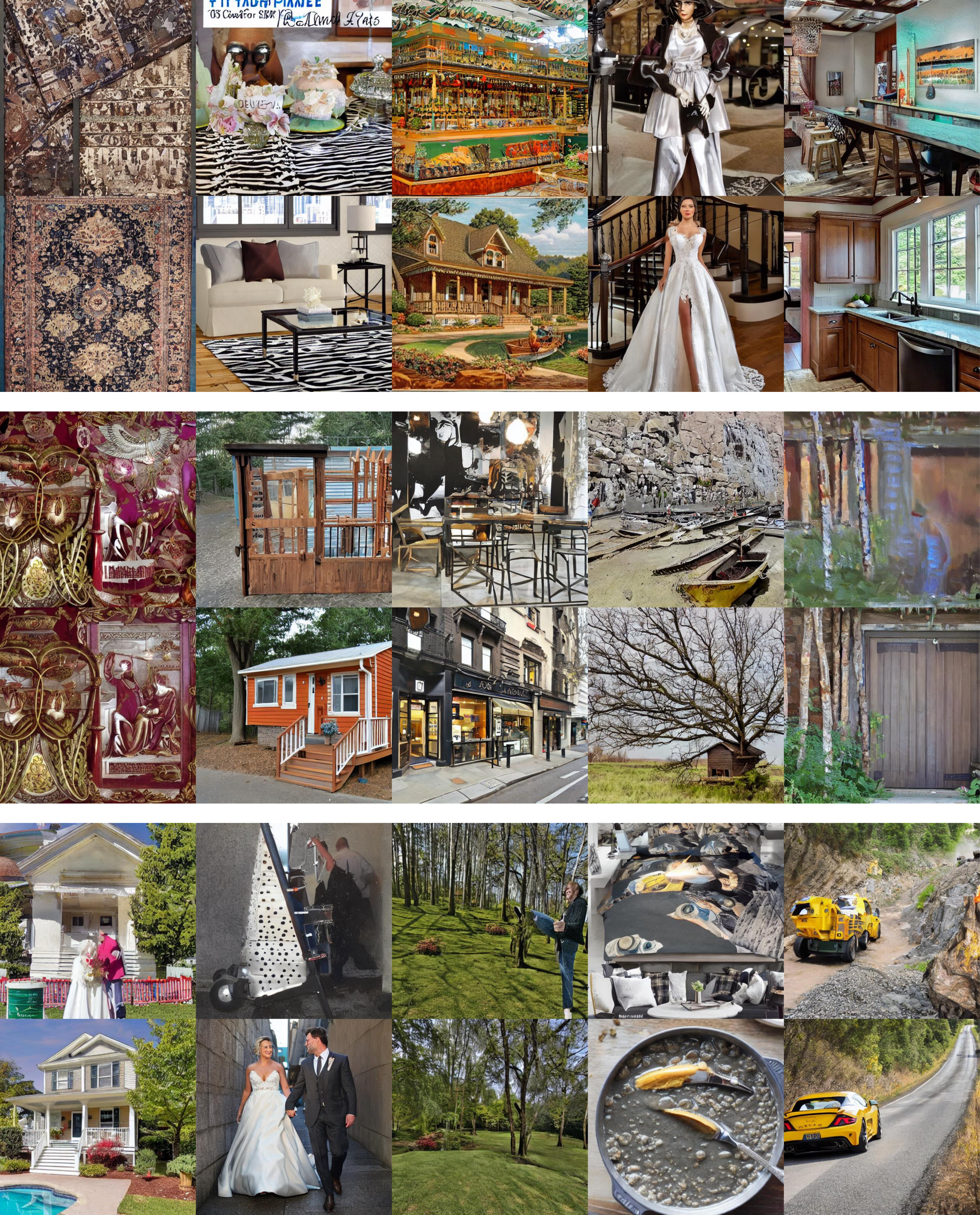}
    \caption{\textbf{Uncurated samples from SD~\cite{rombach2022high} in \textit{unconditional} generation w/o and w/ PAG.} In each image set, the images in the top row are samples without using guidance, and the images in the bottom row are samples using PAG. PAG guidance scale $s=$ 5.0 and perturbed layer \texttt{mid\_block.attentions.0.\-transformer\_blocks.0.attn1} are used.}
    \label{fig:supple_sd_uncond_2}
    \vspace{-10pt}
\end{figure*}

\newpage

\begin{figure*}[!h]
    \includegraphics[width=\textwidth]{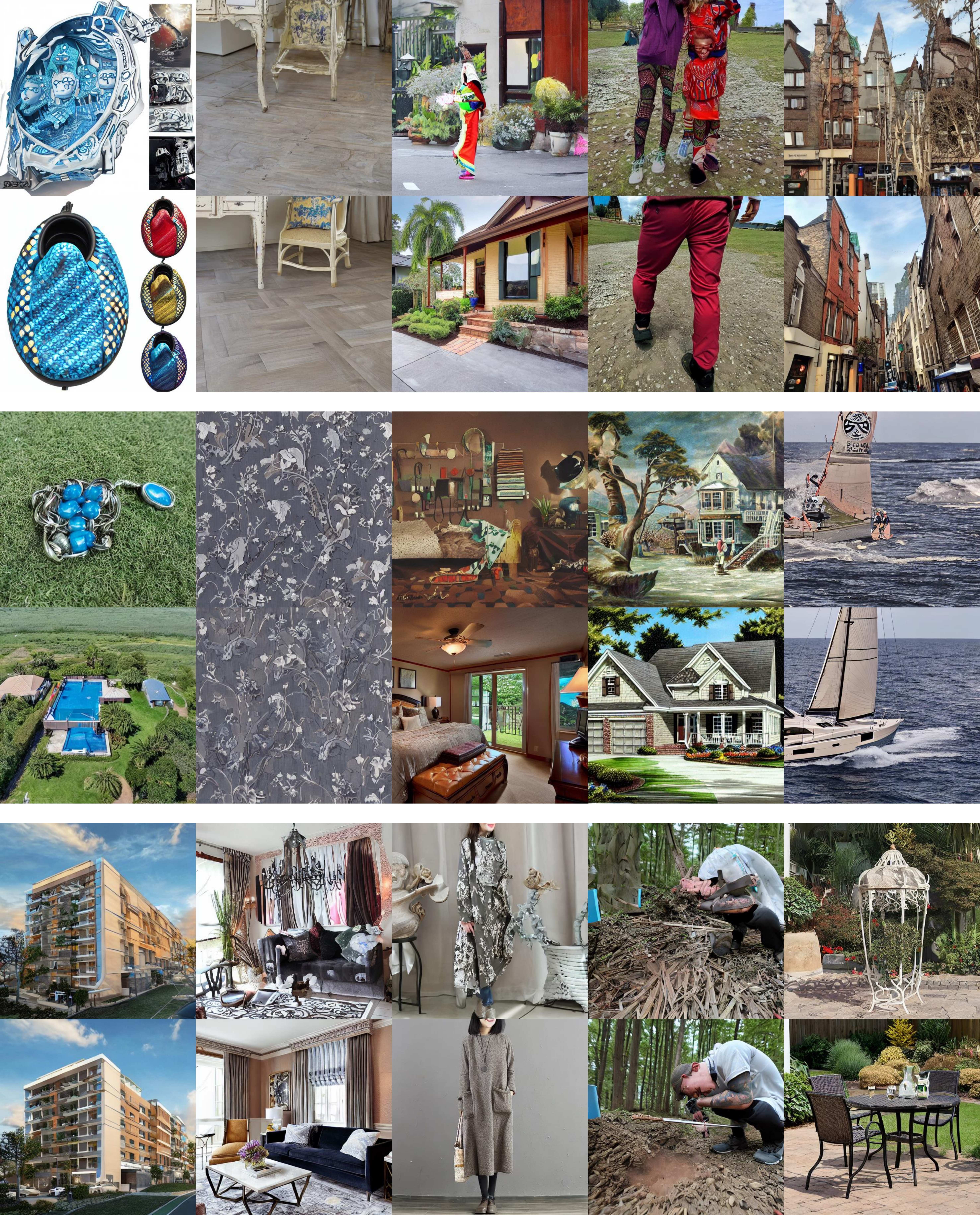}
    \caption{\textbf{Uncurated samples from SD~\cite{rombach2022high} in \textit{unconditional} generation w/o and w/ PAG.} In each image set, the images in the top row are samples without using guidance, and the images in the bottom row are samples using PAG. PAG guidance scale $s=$ 5.0 and perturbed layer \texttt{mid\_block.attentions.0.\-transformer\_blocks.0.attn1} are used.}
    \label{fig:supple_sd_uncond_3}
    \vspace{-10pt}
\end{figure*}


\subsection{PSLD Results}
\subsubsection{FFHQ.}
\begin{figure*}[!htbp]
    \centering
    \includegraphics[width=0.6\linewidth]{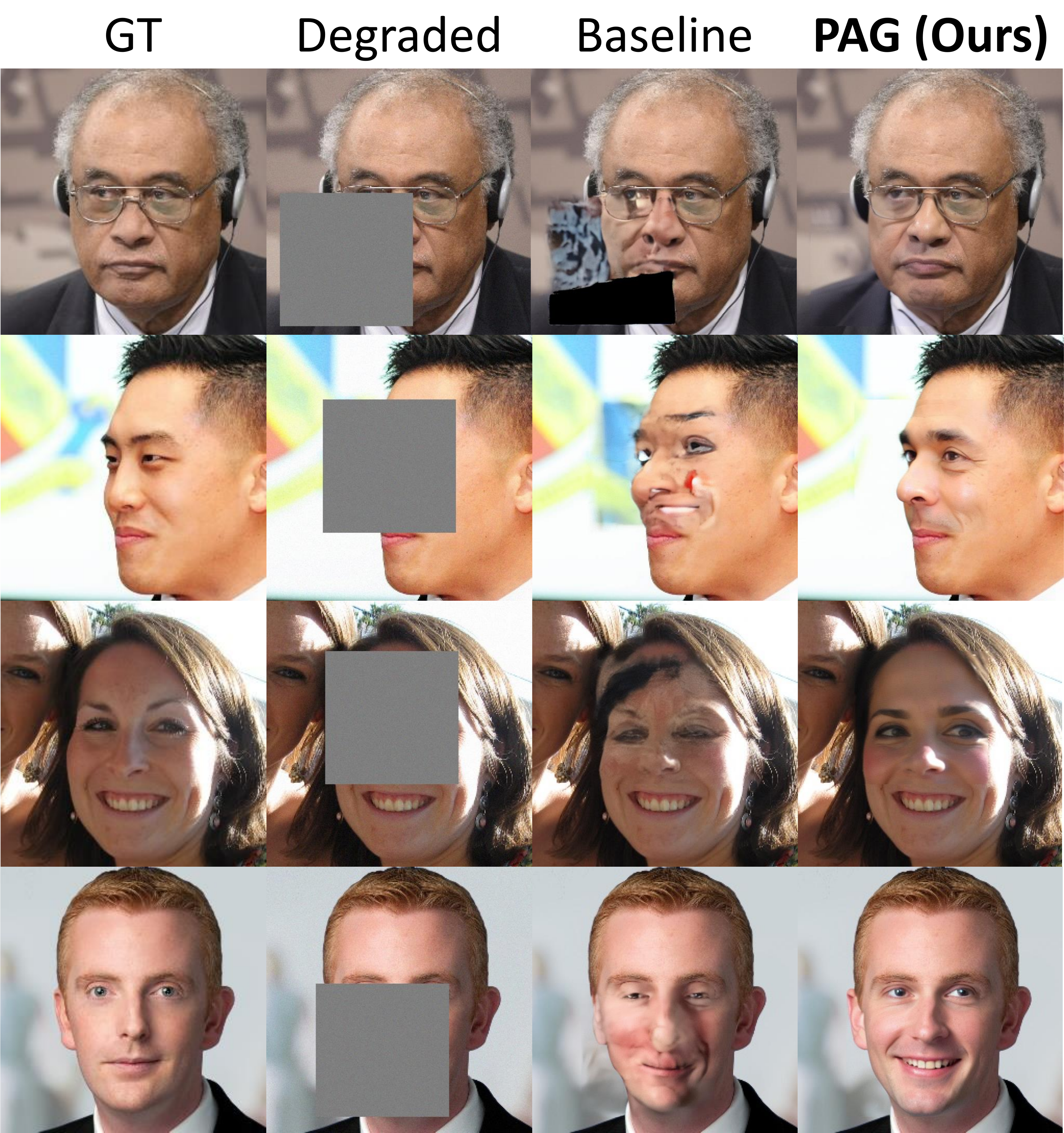} 
    \caption{\textbf{Box inpainting results of PSLD~\cite{rout2024solving} with PAG on FFHQ~\cite{ffhq} dataset.}}
\vspace{-10pt}
\label{fig:psld-bip-qual}
\vspace{-15pt}
\end{figure*}

\begin{figure*}[!htbp]
    \centering
    \includegraphics[width=0.6\linewidth]{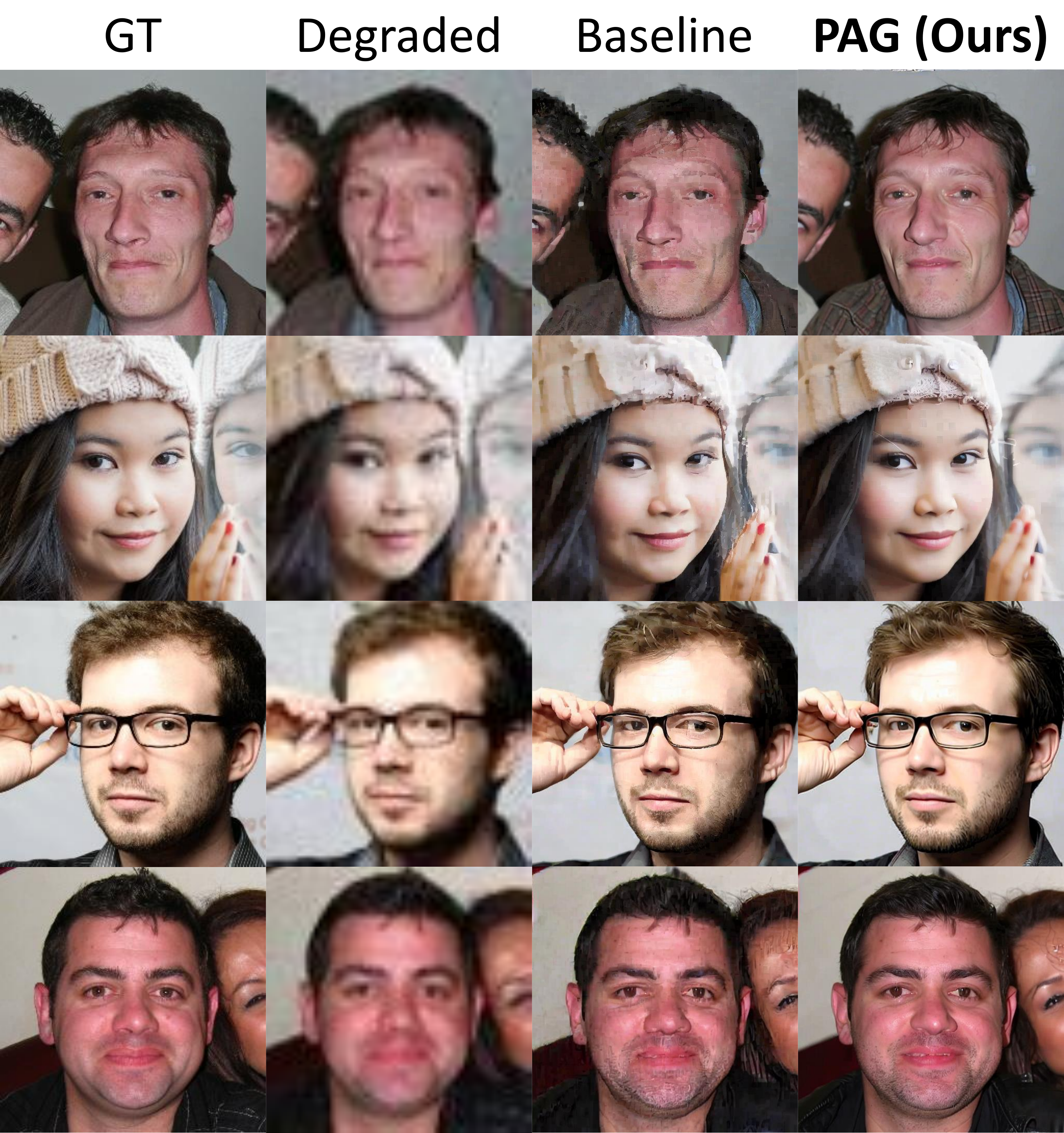} 
    \caption{\textbf{Super-resolution ($\times$8) results of PSLD~\cite{rout2024solving} with PAG on FFHQ~\cite{ffhq} dataset.}}
\vspace{-10pt}
\label{fig:psld-sr-qual}
\end{figure*}

\begin{figure*}[!htbp]
    \centering
    \includegraphics[width=0.6\linewidth]{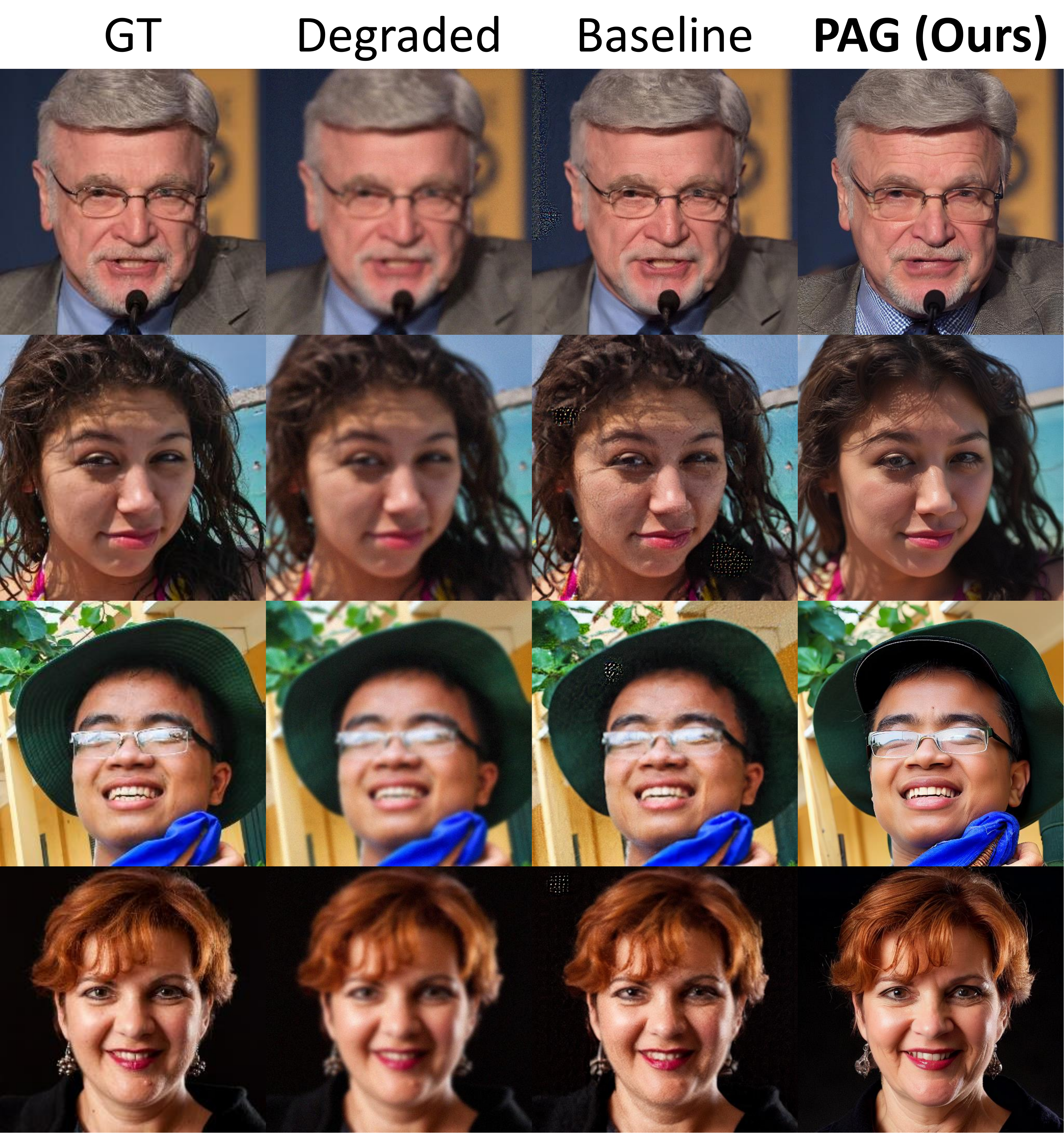} 
    \caption{\textbf{Gaussian deblur results of PSLD~\cite{rout2024solving} with PAG on FFHQ~\cite{ffhq} dataset.}}
\vspace{-10pt}
\label{fig:psld-gb-qual}
\vspace{-15pt}
\end{figure*}

\begin{figure*}[!hbp]
    \centering
    \includegraphics[width=0.6\linewidth]{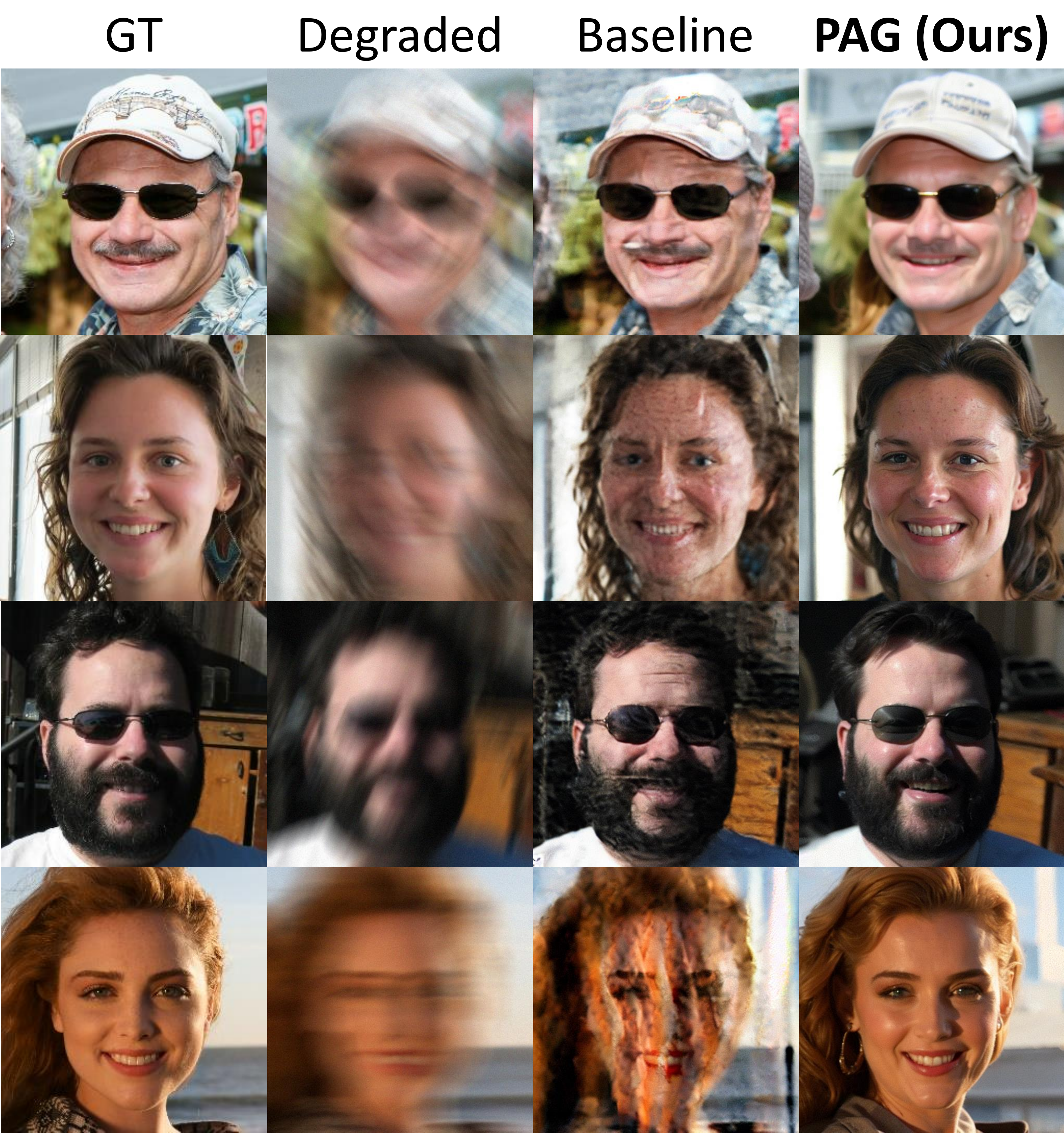} 
    \caption{\textbf{Motion deblur results of PSLD~\cite{rout2024solving} with PAG on FFHQ~\cite{ffhq} dataset.}}
\vspace{-10pt}
\label{fig:psld-mb-qual}
\end{figure*}

Since we use Stable Diffusion v1.5, we upsamle inputs to 512$\times$512 as PSLD~\cite{rout2024solving} does. Then, the outputs are downsampled to 256$\times$256 for evaluation. Further qualitative results are provided in Fig.~\ref{fig:psld-bip-qual}~\ref{fig:psld-sr-qual}~\ref{fig:psld-gb-qual}~\ref{fig:psld-mb-qual}.

\newpage

\clearpage
\newpage
\subsubsection{ImageNet.}
\label{sec:psld-imagenet}
\begin{figure*}[!htbp]
    \centering
    \includegraphics[width=0.6\linewidth]{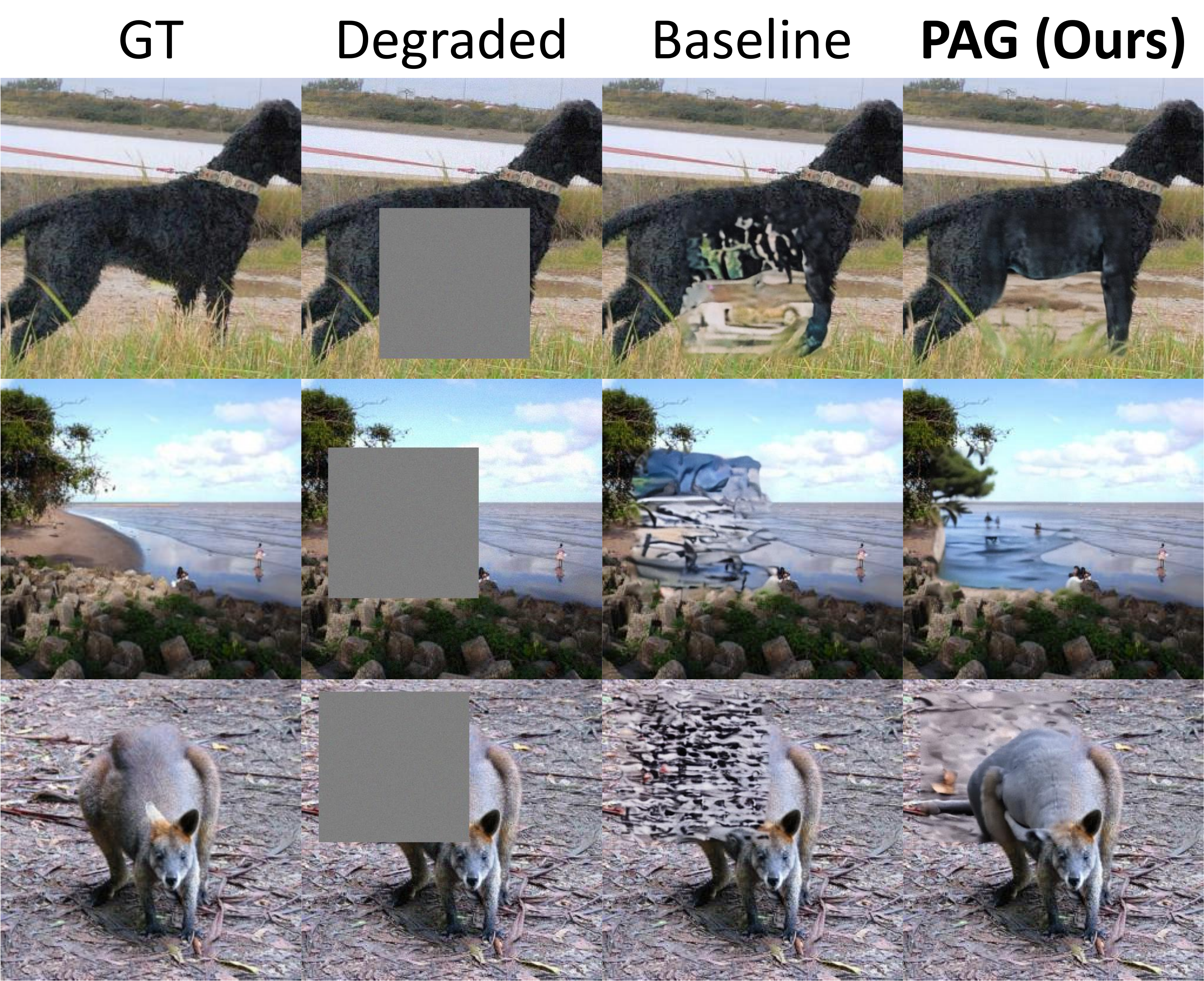} 
    \caption{\textbf{Box inpainting results of PSLD~\cite{rout2024solving} with PAG on ImageNet~\cite{deng2009imagenet} dataset.}}
\vspace{-10pt}
\label{fig:psld-bip-imagenet}
\vspace{-15pt}
\end{figure*}

\begin{figure*}[!htbp]
    \centering
    \includegraphics[width=0.6\linewidth]{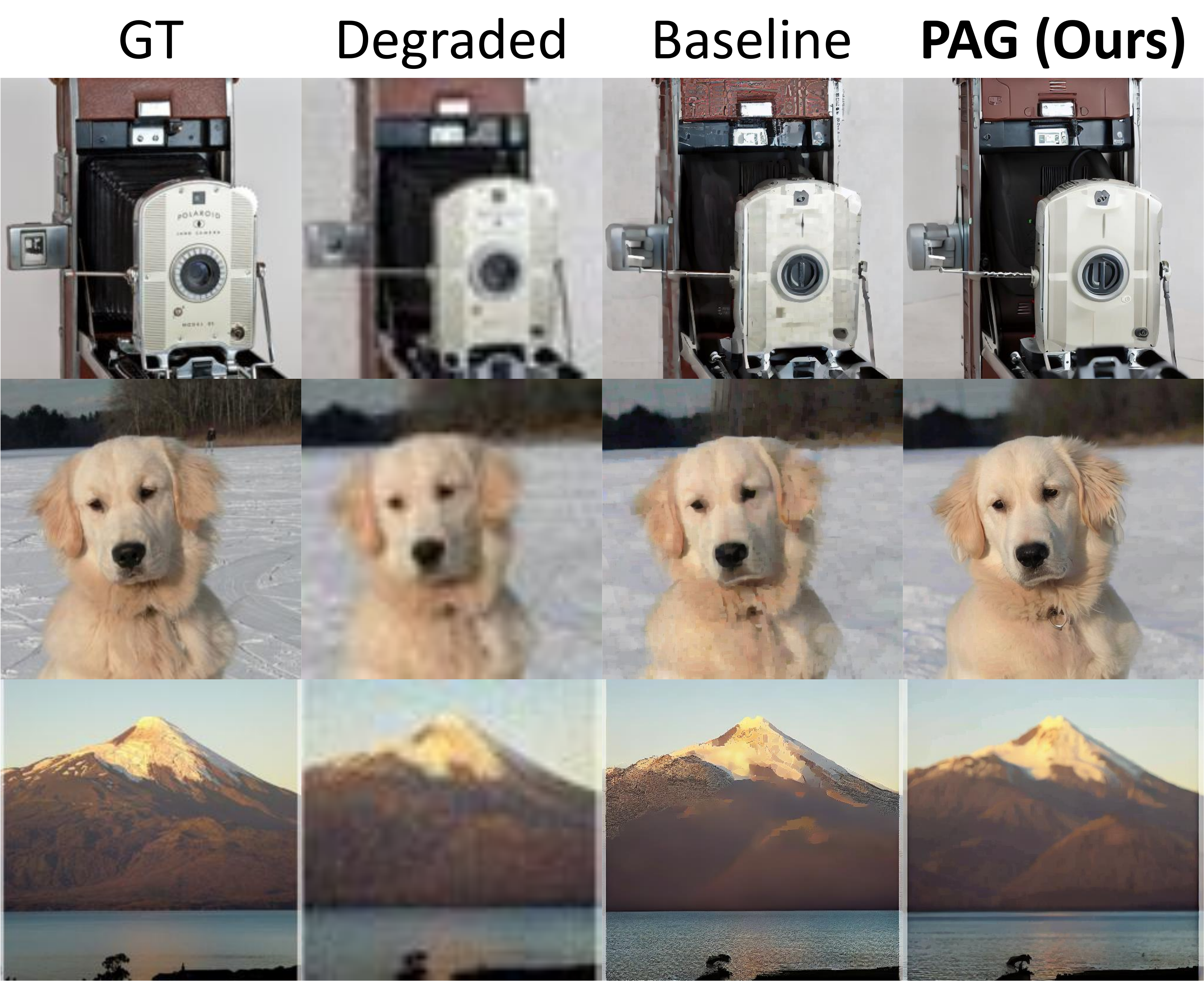} 
    \caption{\textbf{Super-resolution($\times$8) results of PSLD~\cite{rout2024solving} with PAG on ImageNet~\cite{deng2009imagenet} dataset.}}
\vspace{-10pt}
\label{fig:psld-sr-imagenet}
\end{figure*}

\begin{figure*}[!htbp]
    \centering
    \includegraphics[width=0.6\linewidth]{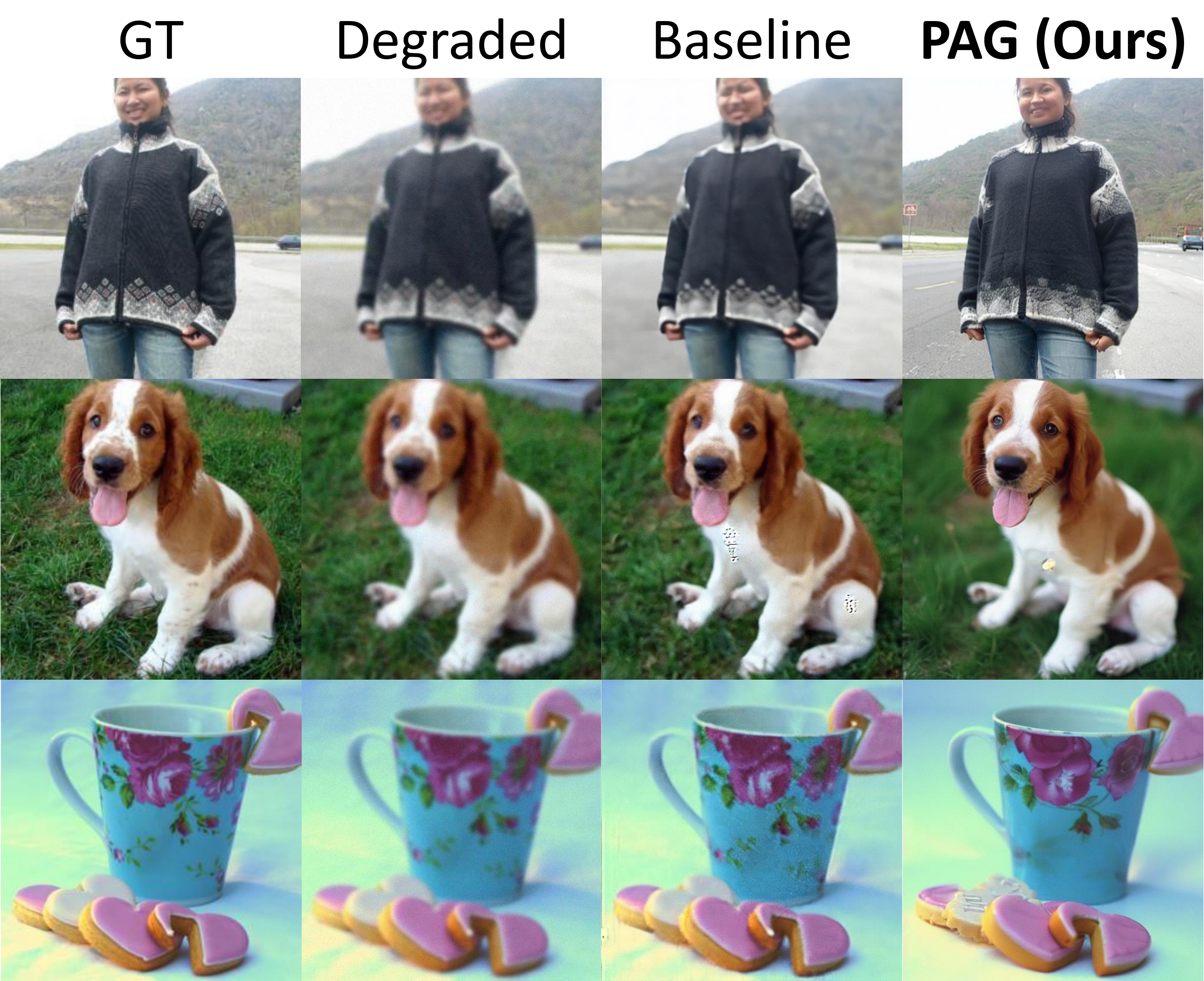} 
    \caption{\textbf{Gaussian deblur results of PSLD~\cite{rout2024solving} with PAG on ImageNet~\cite{deng2009imagenet} dataset.}}
\vspace{-10pt}
\label{fig:psld-gb-imagenet}
\vspace{-15pt}
\end{figure*}

\begin{figure*}[!htbp]
    \centering
    \includegraphics[width=0.6\linewidth]{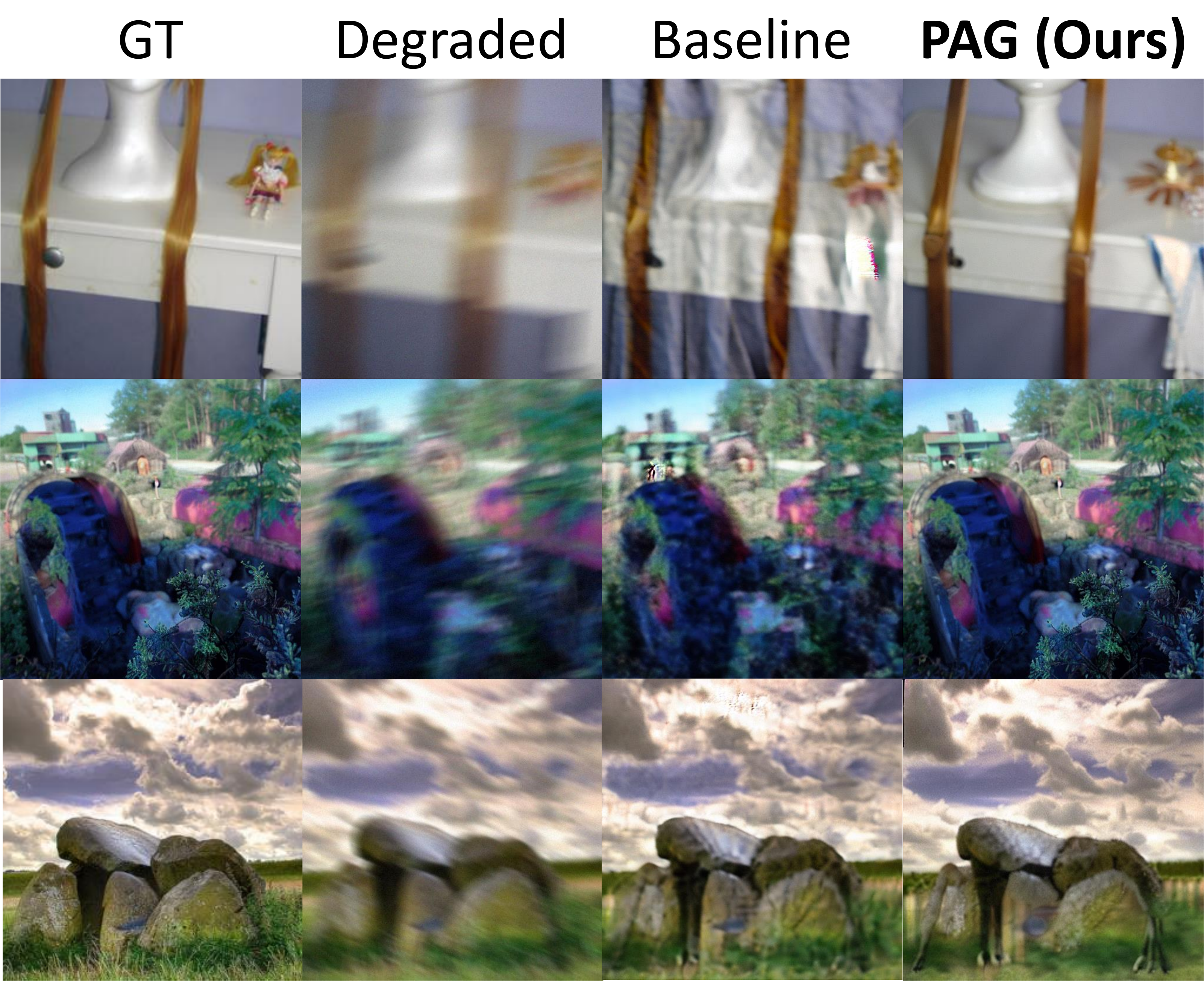} 
    \caption{\textbf{Motion deblur results of PSLD~\cite{rout2024solving} with PAG on ImageNet~\cite{deng2009imagenet} dataset.}}
\vspace{-10pt}
\label{fig:psld-mb-imagenet}
\end{figure*}

We use 1K ImageNet~\cite{deng2009imagenet} 256$\times$256 dataset which is used in~\cite{kawar2022denoising,chung2022diffusion, rout2024solving}. Qualitative results shows that PAG properly improves sample quality with more various classes of images, as provided in Fig.~\ref{fig:psld-bip-imagenet}~\ref{fig:psld-sr-imagenet}~\ref{fig:psld-gb-imagenet}~\ref{fig:psld-mb-imagenet}.

\newpage

\clearpage
\newpage
\section{Additional Applications}
\label{sec:sup:additioanl_applications}
\subsection{Diffusion Posterior Sampling}
We conduct additional experiments on another diffusion restoration model, DPS~\cite{chung2022diffusion}, which is based on ADM~\cite{dhariwal2021diffusion}. DPS~\cite{chung2022diffusion} updates the gradient of the loss term to perform sampling from the posterior distribution~\cite{chung2022diffusion}. The detailed hyperparameters for all the DPS~\cite{chung2022diffusion} experiments are presented in Table~\ref{tab:dps-hyparam}. Here, we only use the unconditional score $\epsilon_\theta(\boldsymbol{z}_t)$ for predicting $\hat{\boldsymbol{z}}_0$, consistent with PSLD~\cite{chung2022diffusion} experiments.

\begin{table}[]
\centering
\captionsetup{skip=5pt}
\setlength{\tabcolsep}{2pt}
\resizebox{1.0\textwidth}{!}{%
\begin{tabular}{llllllllll}
\toprule
{} & {} & \multicolumn{4}{c}{\textbf{FFHQ}} & \multicolumn{4}{c}{\textbf{ImageNet}}\\
\cmidrule(r){3-6}
\cmidrule(r){7-10}
{} & {} & Inpaint & SR$\times8$ & Gauss & Motion  & Inpaint & SR$\times8$ & Gauss & Motion \\
{DPS} & $\eta$ & 1.0 & 1.0 & 1.0 & 1.0 & 1.0 & 1.0 & 0.4 & 0.6 \\
\midrule
{DPS $+$ \textbf{PAG (Ours)}} & $\eta$ & 1.0 & 1.0 & 1.0 & 1.0 & 1.0 & 1.0 & 0.4 & 1.0 \\
{} & $s$ & 1.0 & 1.0 & 1.0 & 1.0 & 2.0 &2.0 & 1.0 & 2.0  \\
{} & layer & \multicolumn{4}{c}{\texttt{input9.1}} & \multicolumn{4}{c}{\texttt{input9.1} \texttt{middle.1} \texttt{output2.1}}\\
\bottomrule
\end{tabular}
}
\caption{Hyperparameters for DPS~\cite{chung2022diffusion} w/o and w/ \textbf{PAG} on FFHQ~\cite{ffhq} dataset and ImageNet~\cite{deng2009imagenet} dataset. $\eta$ is the step size for updating gradients of DPS~\cite{chung2022diffusion} and $s$ is the scale for PAG from Eq.~\ref{eq:PAG-derivation} of main paper.} 
\label{tab:dps-hyparam}
\vspace{-20pt}
\end{table}

\begin{figure*}[!htbp]
    \centering
    \includegraphics[width=0.6\linewidth]{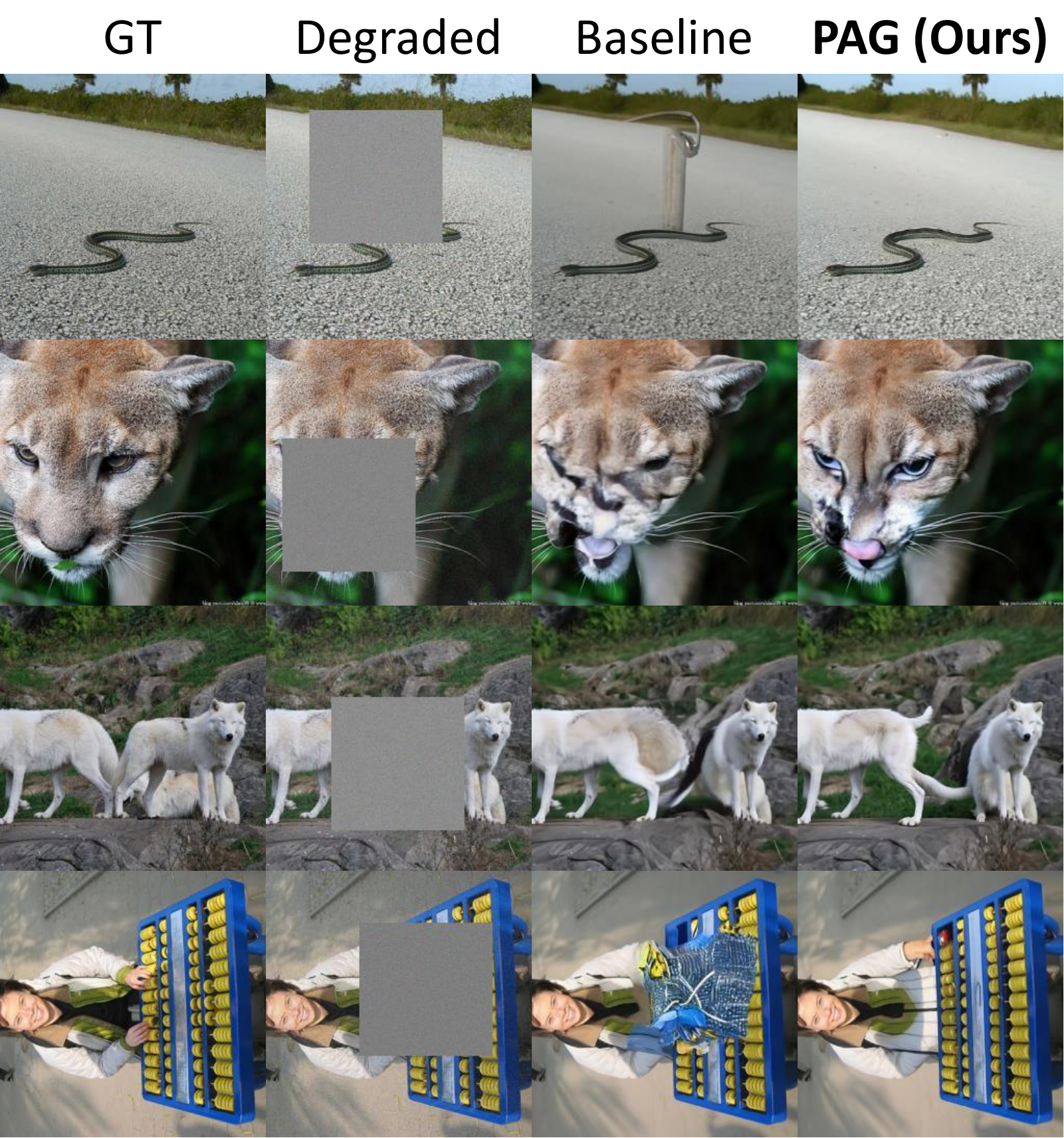} 
    \caption{\textbf{Box inpainting results of DPS~\cite{chung2022diffusion} with PAG on ImageNet~\cite{deng2009imagenet} dataset.}}
\vspace{-10pt}
\label{fig:dps-imagenet-bip}
\end{figure*}

\begin{figure*}[!htbp]
    \centering
    \includegraphics[width=0.6\linewidth]{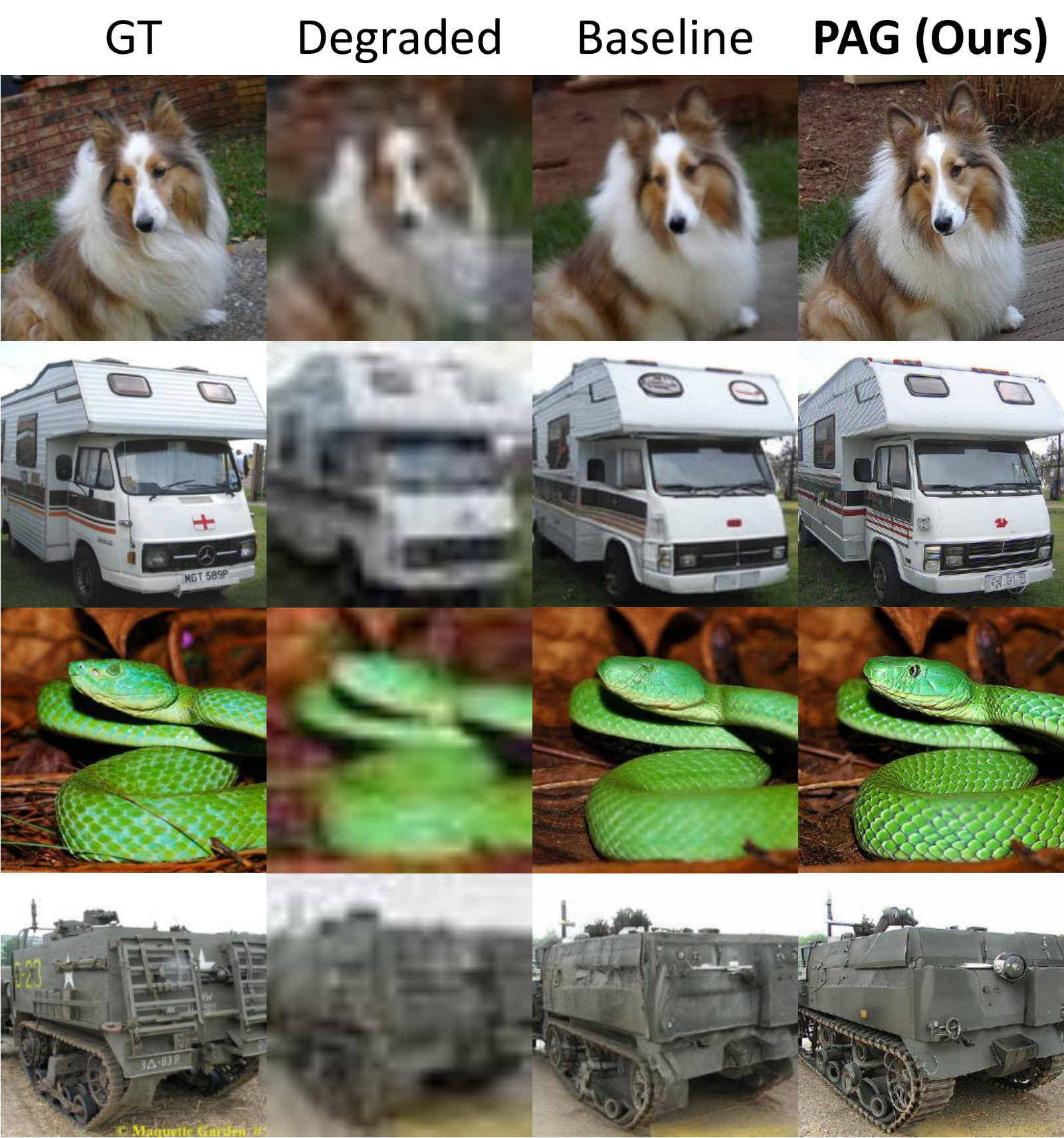} 
    \caption{\textbf{Super-resolution($\times$8) results of DPS~\cite{chung2022diffusion} with PAG on ImageNet~\cite{deng2009imagenet} dataset.}}
\vspace{-10pt}
\label{fig:dps-imagenet-sr}
\end{figure*}

\begin{figure*}[!htbp]
    \centering
    \includegraphics[width=0.6\linewidth]{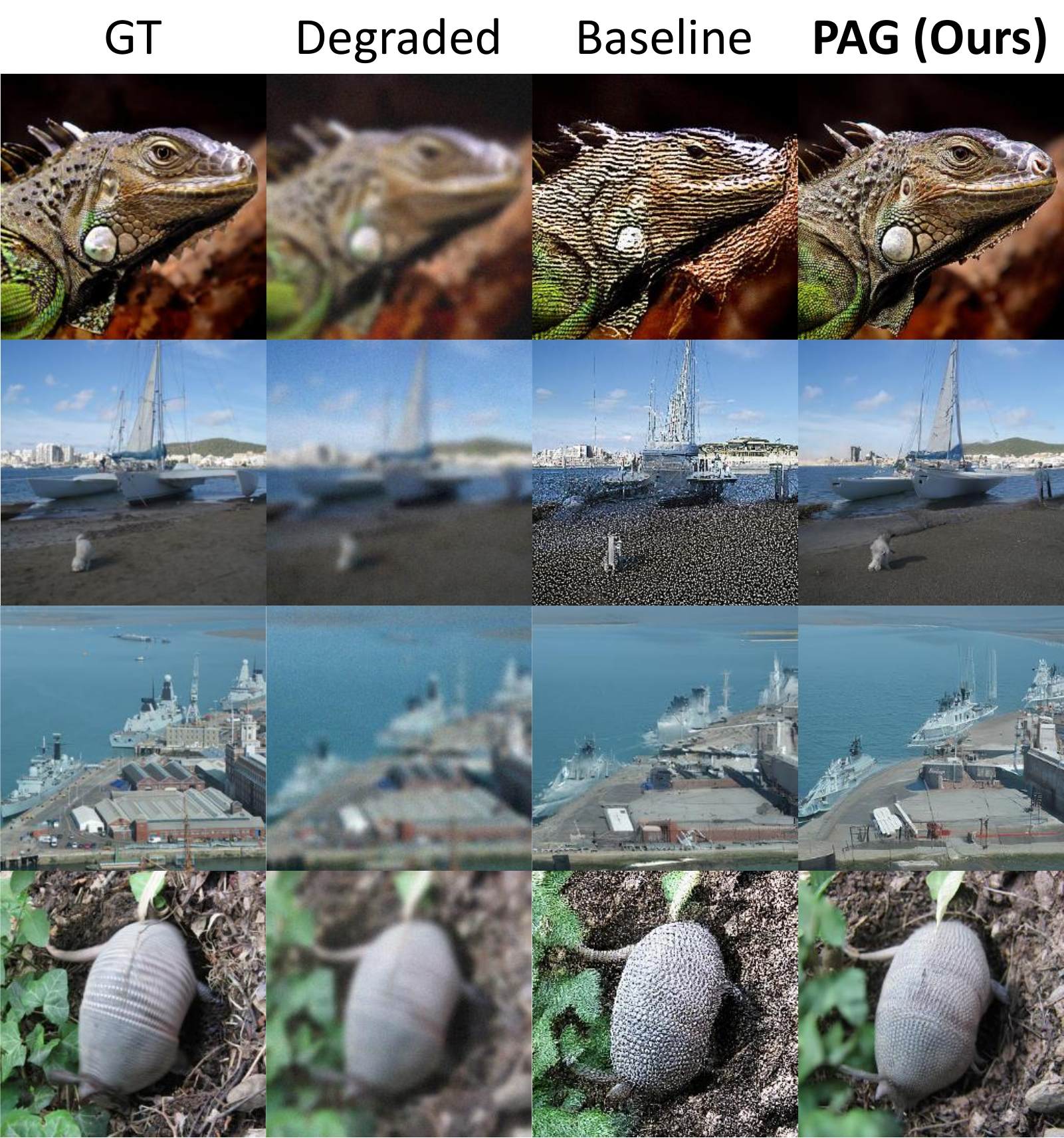} 
    \caption{\textbf{Gaussian deblur results of DPS~\cite{chung2022diffusion} with PAG on ImageNet~\cite{deng2009imagenet} dataset.}}
\vspace{-10pt}
\label{fig:dps-imagenet-gb}
\end{figure*}

\begin{figure*}[!htbp]
    \centering
    \includegraphics[width=0.6\linewidth]{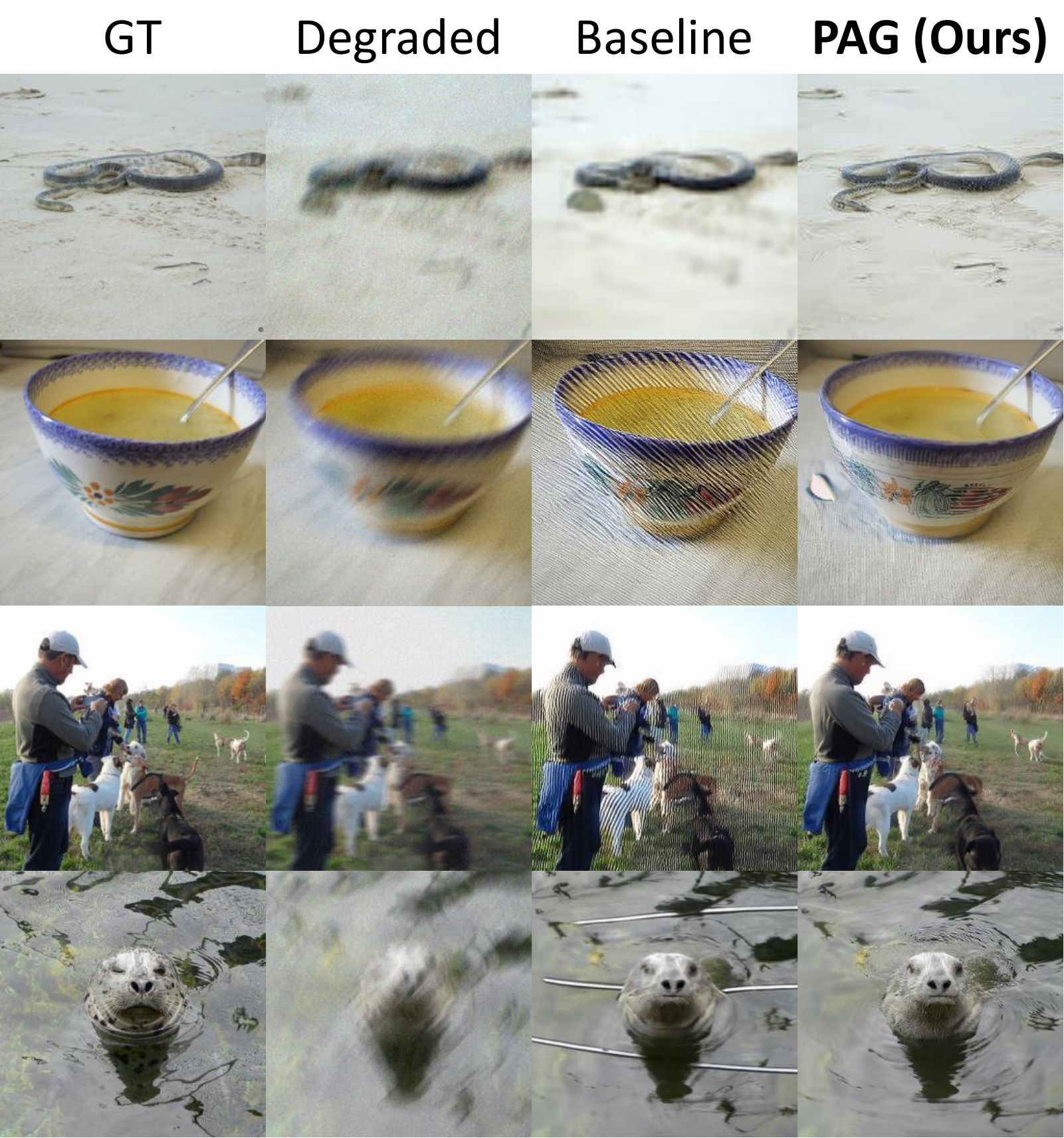} 
    \caption{\textbf{Motion deblur results of DPS~\cite{chung2022diffusion} with PAG on ImageNet~\cite{deng2009imagenet} dataset.}}
\vspace{-10pt}
\label{fig:dps-imagenet-mb}
\end{figure*}

\begin{figure*}[!htbp]
    \centering
    \includegraphics[width=0.6\linewidth]{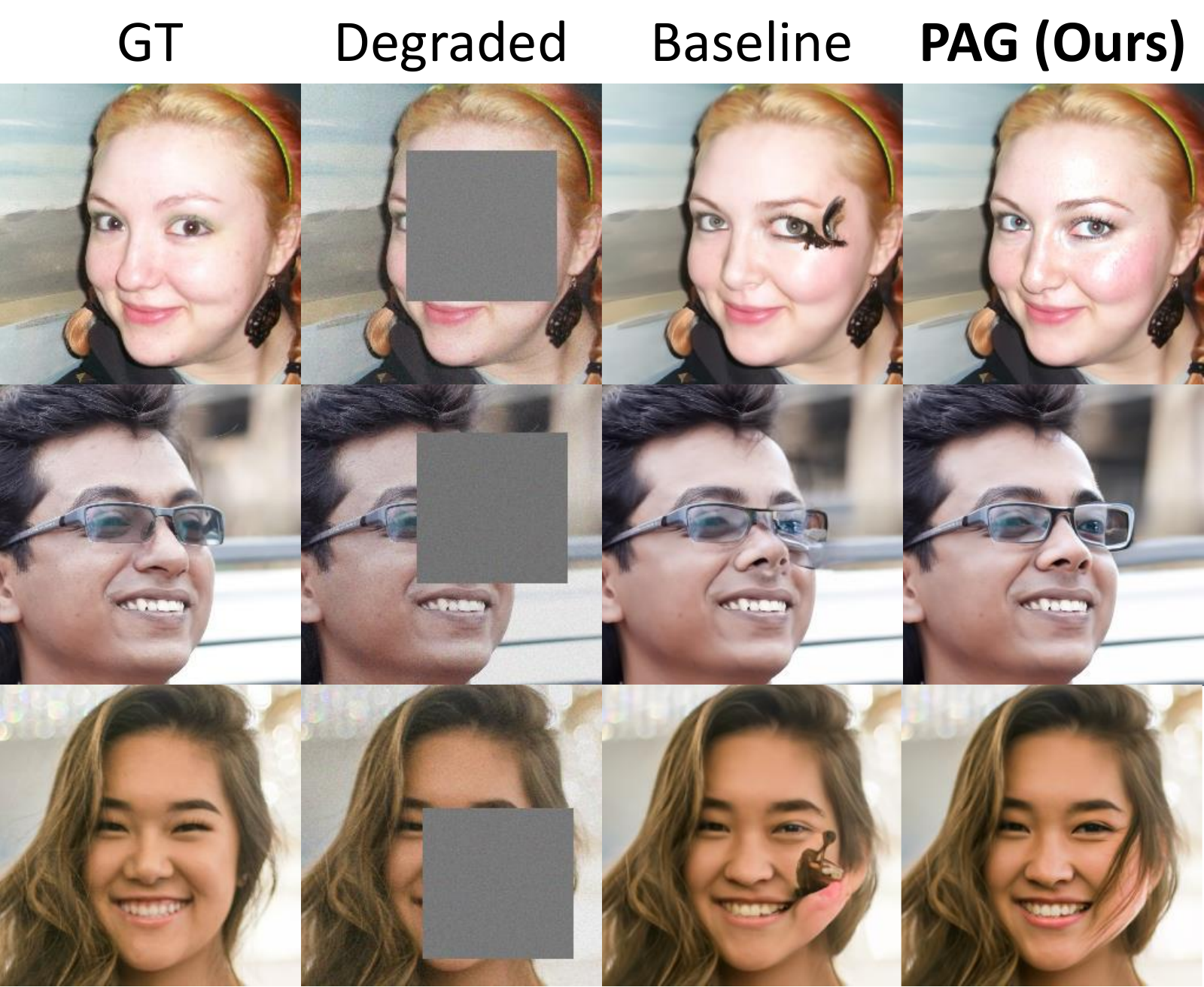} 
    \caption{\textbf{Box inpainting results of DPS~\cite{chung2022diffusion} with PAG on FFHQ~\cite{ffhq} dataset.}}
\vspace{-10pt}
\label{fig:dps-ffhq-bip}
\end{figure*}

All experiments with DPS~\cite{chung2022diffusion} use DDPM~\cite{ho2020denoising} sampling. Quantitative results on 1K 256 are provided in Table~\ref{tab:dps-quan}. PAG outperforms baseline on FID~\cite{heusel2017gans}, except for super-resolution($\times$8), where FID~\cite{heusel2017gans} is comparable. This result may be attributed to the point that sampling images with hard degradations can be regarded as generation rather than restoration, which emphasizes the importance of FID~\cite{heusel2017gans} over LPIPS~\cite{zhang2018unreasonable}. Additional qualitative results are in Fig.~\ref{fig:dps-imagenet-bip}~\ref{fig:dps-imagenet-sr}~\ref{fig:dps-imagenet-gb}~\ref{fig:dps-imagenet-mb} for ImageNet~\cite{deng2009imagenet} dataset and Fig.~\ref{fig:dps-ffhq-bip} for FFHQ~\cite{ffhq} dataset.

\vspace{-10pt}
\begin{table}[!htbp]
  \centering
  \caption{\textbf{Quantitative results of DPS~\cite{chung2022diffusion} on FFHQ~\cite{ffhq} 256$\times$256 1K validation set~\cite{ffhq}}.}
  \captionsetup{skip=5pt}
  \setlength{\tabcolsep}{5pt}
  \resizebox{1.0\textwidth}{!}{
  \begin{tabular}{lcccccccc}
    \toprule
    {} & \multicolumn{2}{c}{Box Inpainting} & \multicolumn{2}{c}{{\color{black}SR ($8\times$)}}    & \multicolumn{2}{c}{{\color{black}Gaussian Deblur}}    &
    \multicolumn{2}{c}{{\color{black}Motion Deblur}}\\
    \cmidrule(r){2-3}   \cmidrule(r){4-5}   \cmidrule(r){6-7} \cmidrule(r){8-9}
    Method & FID $\downarrow$ & LPIPS $\downarrow$ & FID $\downarrow$ & LPIPS $\downarrow$ & FID $\downarrow$ & LPIPS $\downarrow$ & FID $\downarrow$ & LPIPS $\downarrow$\\
    \midrule
    DPS & 33.12 &  \textbf{0.168}  & \textbf{34.00}& \textbf{0.320}& 44.05 & \textbf{0.257} & 39.92 & \textbf{0.242} \\
    DPS $+$ \textbf{PAG (Ours)}   & \textbf{26.74} & 0.212  & 34.05 & 0.327& \textbf{29.42}  & 0.259 & \textbf{30.57}  & 0.283 \\
    \bottomrule
  \end{tabular}%
  }
  \label{tab:dps-quan}
\vspace{-20pt}
\end{table}

\newpage
\subsection{Stable Diffusion Super-Resolution and Inpainting}
\begin{figure*}[!t]
    \centering
    \includegraphics[width=\linewidth]{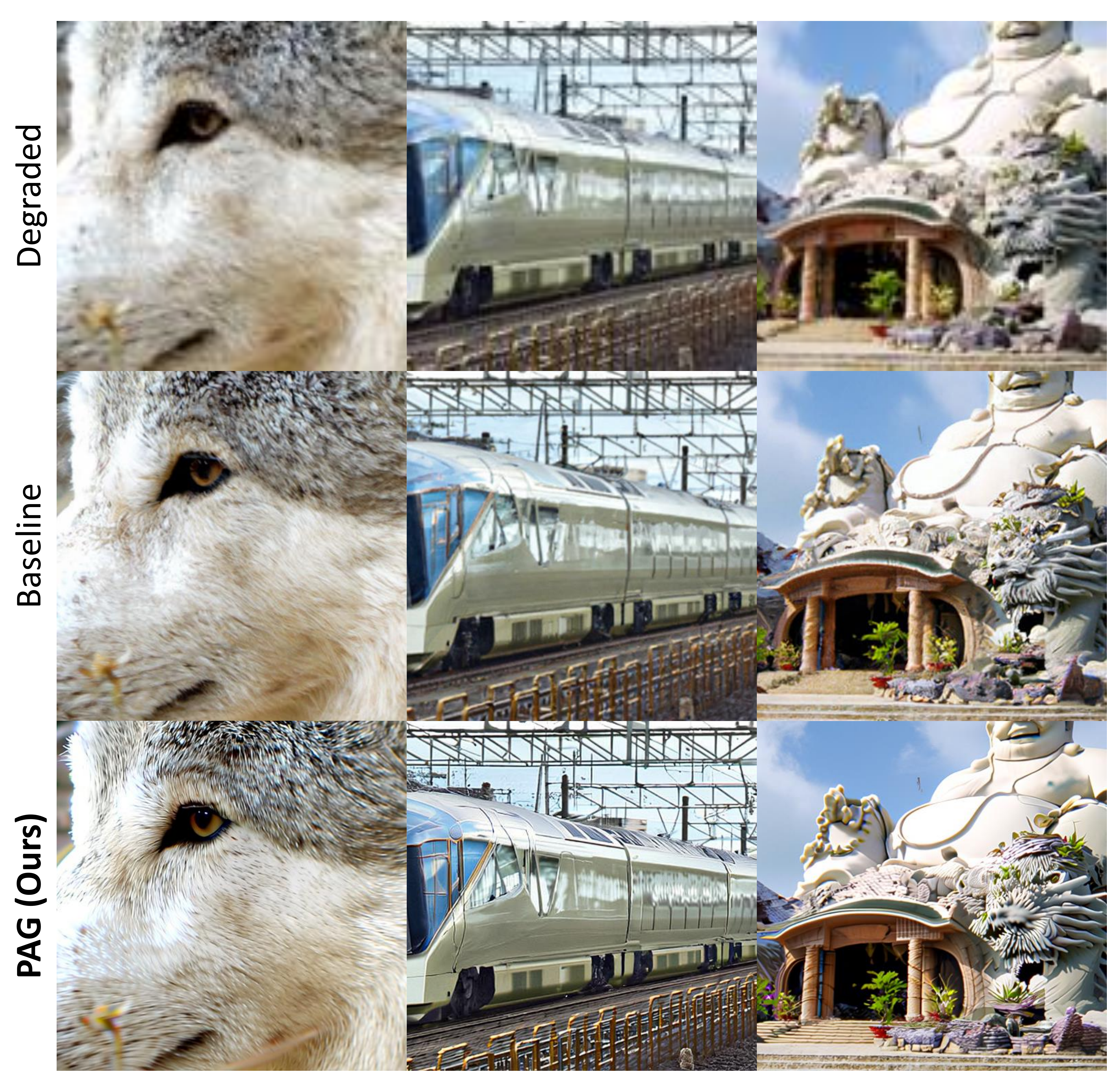} 
    \vspace{-15pt}
    \caption{\textbf{Comparison of Stable Diffusion~\cite{rombach2022high} super-resolution results between w/o and w/ PAG.} PAG applies guidance that enables the model to upscale images to high-quality renditions with clearer edges and finer details, even when using an empty prompt (3rd row). The guidance scales employed, from left to right, are sequentially 3.0, 2.0, and 1.0. The model upscaled a 256$\times$256 input image to 512$\times$512.}
\label{fig:sup:SD-upscale}
\vspace{-10pt}
\end{figure*}

\begin{figure*}[!t]
    \centering
    \includegraphics[width=\linewidth]{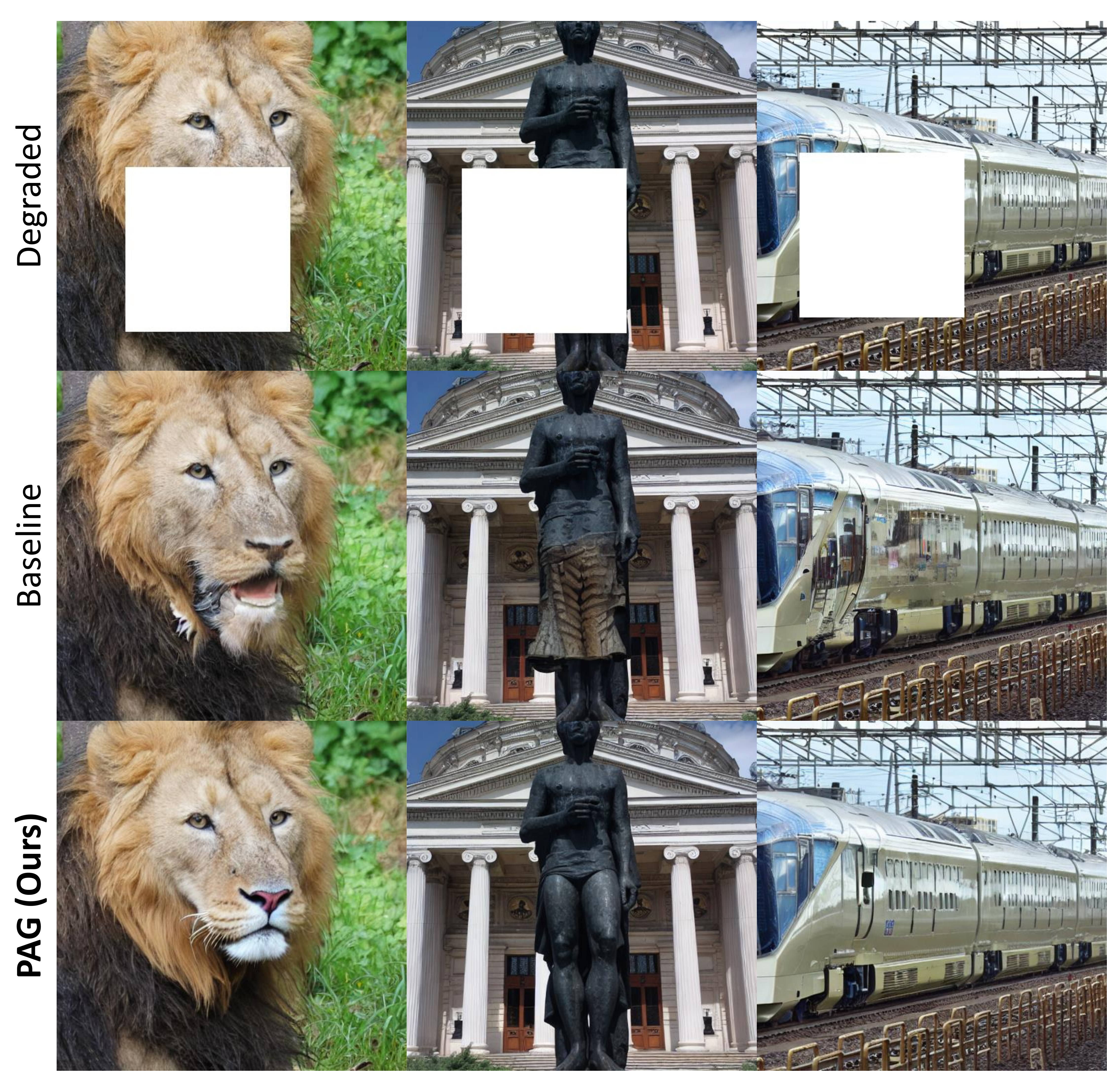} 
    \vspace{-5pt}
    \caption{\textbf{Comparison of Stable Diffusion~\cite{rombach2022high} inpainting results between w/o and w/ PAG.} PAG aids the model in inpainting images, improving their realism and diminishing artifacts, without the necessity for a prompt (3rd row). The guidance scale of 1.5 is employed for all.}
\label{fig:sup:SD-inpaint}
\vspace{-15pt}
\end{figure*}

Stable Diffusion~\cite{rombach2022high} extends beyond the text-to-image pipeline to also support tasks requiring image input, such as super-resolution\footnote{https://huggingface.co/stabilityai/stable-diffusion-x4-upscaler} and inpainting\footnote{https://huggingface.co/runwayml/stable-diffusion-inpainting}.
The model also requires text input alongside image input to leverage CFG~\cite{ho2022classifier}, yet there are instances where input prompts do not fit. For example, in a landscape photo, it may be more intuitive to specify only the area to be removed (such as a person in the background, shadows, or lens artifacts) and naturally fill it to match the surroundings, rather than providing a text prompt describing the entire content of the current image. Similarly, for super-resolution, it is more natural to input the image alone without having to describe it entirely in text, especially for real images. While synthetic images may have an associated creation prompt, real images do not, making it challenging to provide suitable text prompts. In contrast, PAG does not require text prompts, providing a natural way to enhance the quality of results in such pipelines. Fig.~\ref{fig:sup:SD-upscale} and ~\ref{fig:sup:SD-inpaint} present the outcomes of applying PAG to the Stable Diffusion super-resolution and inpainting pipelines, where the use of PAG produces sharper and more realistic results compared to those without it, offering a much more natural approach for these tasks. We select a subset of the DIV2K~\cite{Ignatov_2018_ECCV_Workshops} dataset downscaled by a factor of 2 using bicubic interpolation and then center cropped to adjust the images to a resolution of 512$\times$512.

\clearpage
\subsection{Text-to-3D}
\begin{figure*}[!htpb]
    \centering
    \vspace{20pt}
    \includegraphics[width=\linewidth]{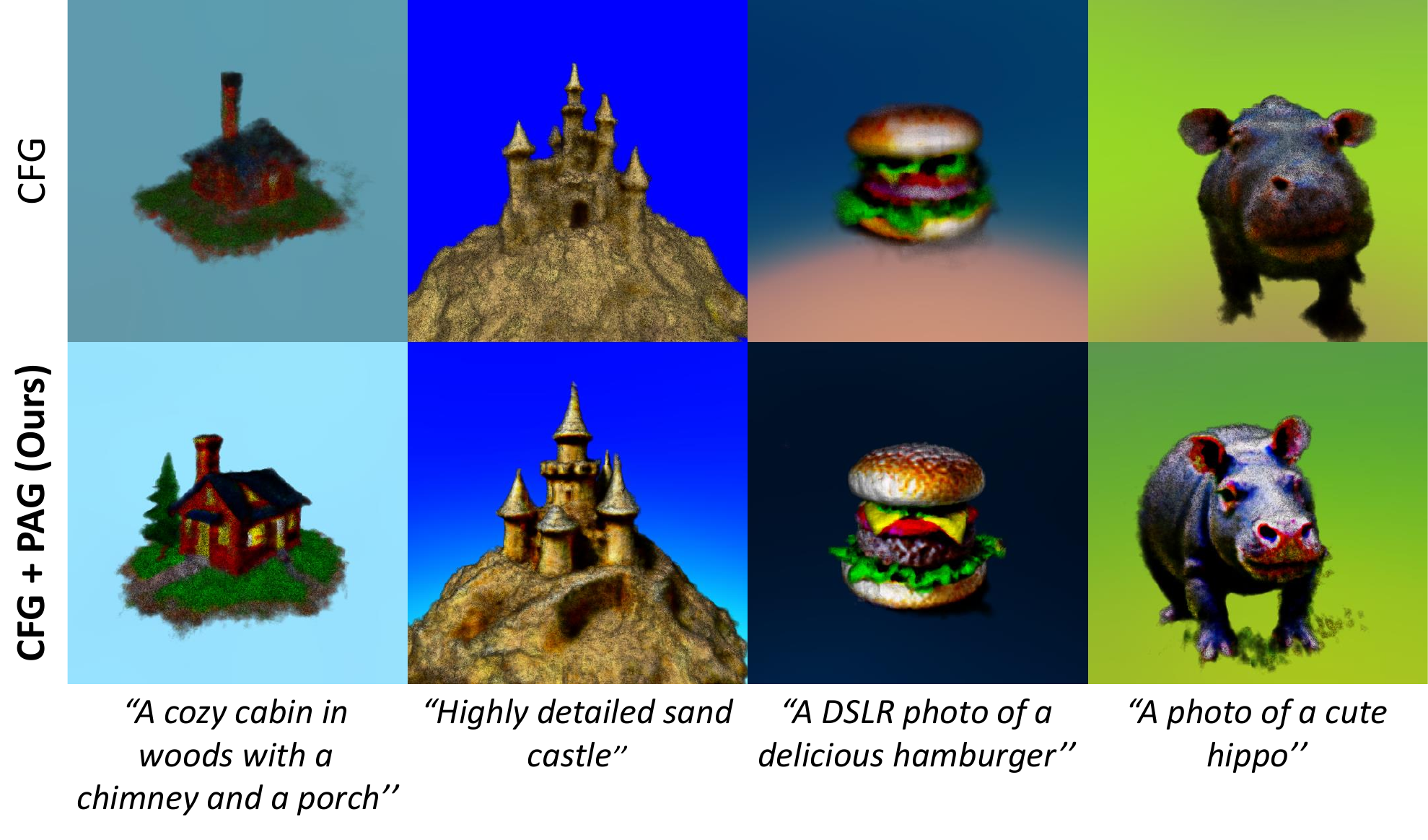} 
    \vspace{-5pt}
    \caption{\textbf{Comparison of text-to-3D results between CFG~\cite{ho2022classifier}, and CFG with PAG.} }
\vspace{-10pt}
\label{fig:3d}
\vspace{-10pt}
\end{figure*}

We integrated PAG with CFG for text-to-3D generation, utilizing the Dreamfusion~\cite{poole2022dreamfusion} implementation provided by Threestudio~\cite{threestudio2023} due to the unavailability of official code. We employed a scale of 100 for both CFG and PAG. As seen in Fig.~\ref{fig:3d}, combining CFG with PAG yields results with enhanced details and textures compared to using CFG alone.

\clearpage
\subsection{Human Evaluation}
\label{sec:human_evaluation}
\begin{figure*}[!hptb]
    \includegraphics[width=\textwidth]{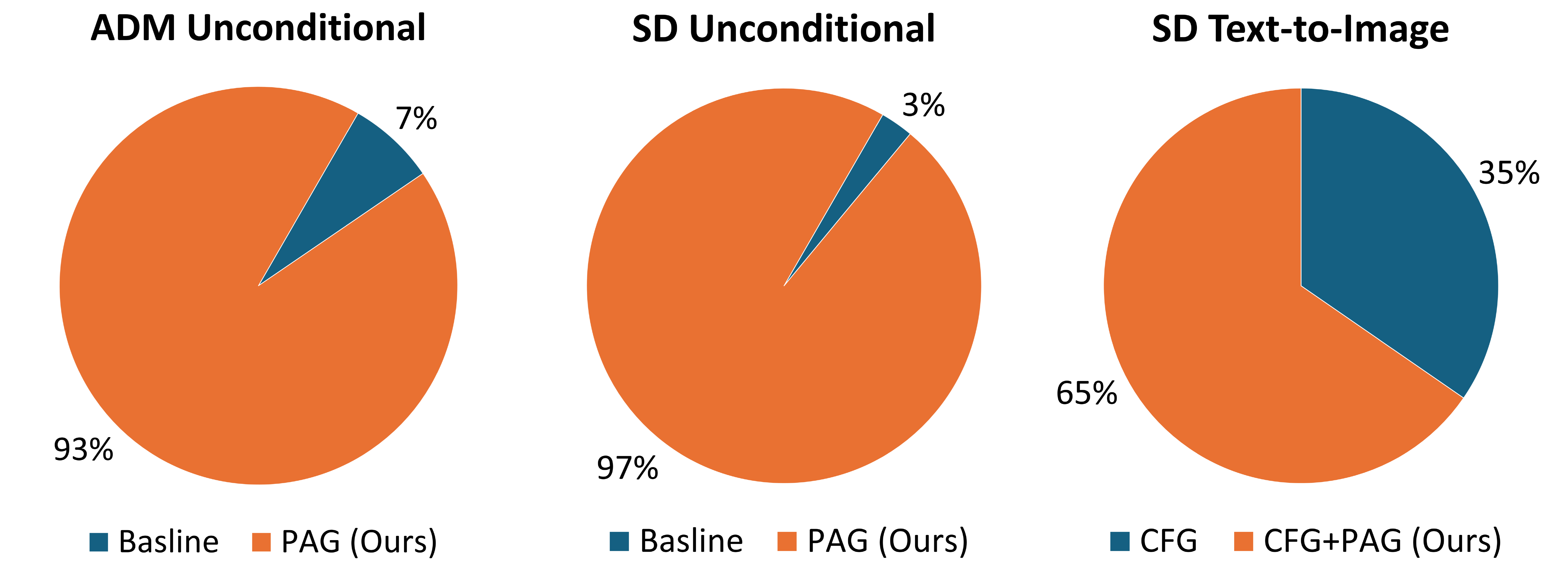}
    \caption{\textbf{The results of the user study.}}
    \label{fig:human_eval}
    \vspace{-10pt}
\end{figure*}
A user study (Fig.~\ref{fig:human_eval}) conducts to evaluate the quality of samples in ADM unconditional, Stable Diffusion unconditional, and Stable Diffusion text-to-image synthesis models. In the cases of unconditional generation, participants are presented with sets of four images sampled both with and without PAG and ask to identify the higher quality samples. For text-to-image synthesis, participants compare sets of four images generated using only CFG against those using both CFG and PAG. Each task comprises 10 questions, resulting in a total of 30 questions evaluated by 60 participants. The results show that the majority of unconditional generation questions prefer samples generated with PAG. Similarly, in the text-to-image synthesis task, samples generated with both CFG and PAG are frequently rated as higher quality.

\clearpage
\section{Ablation Studies}
\label{sec:sup:ablations}

\subsection{Guidance scale}

\begin{figure*}[!t]
    \centering
    \includegraphics[width=\textwidth]{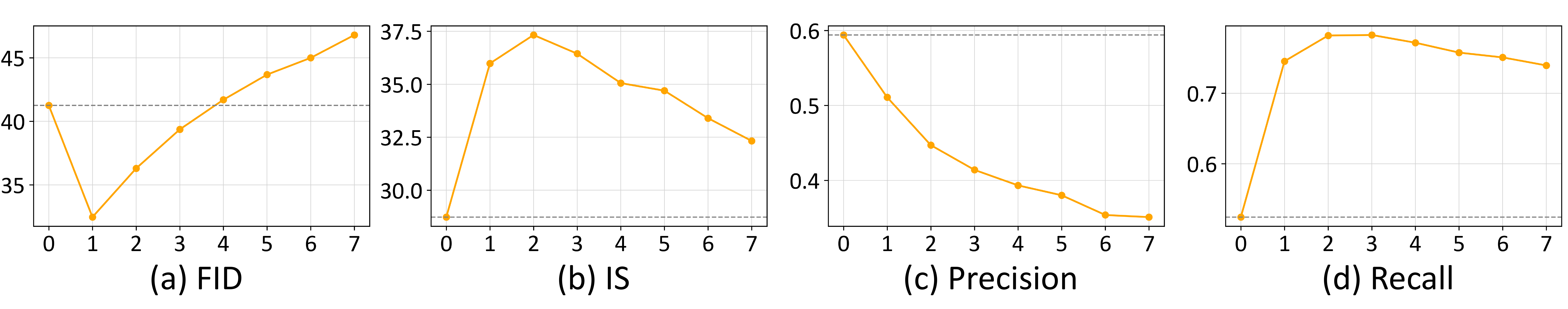}
    \vspace{-15pt}
    \caption{\textbf{Quantitative Analysis of Guidance Scale.}}
    \label{fig:scalequal}
    \vspace{-10pt}
\end{figure*}

\begin{figure*}[!t]
    \centering
    \includegraphics[width=0.9\linewidth]{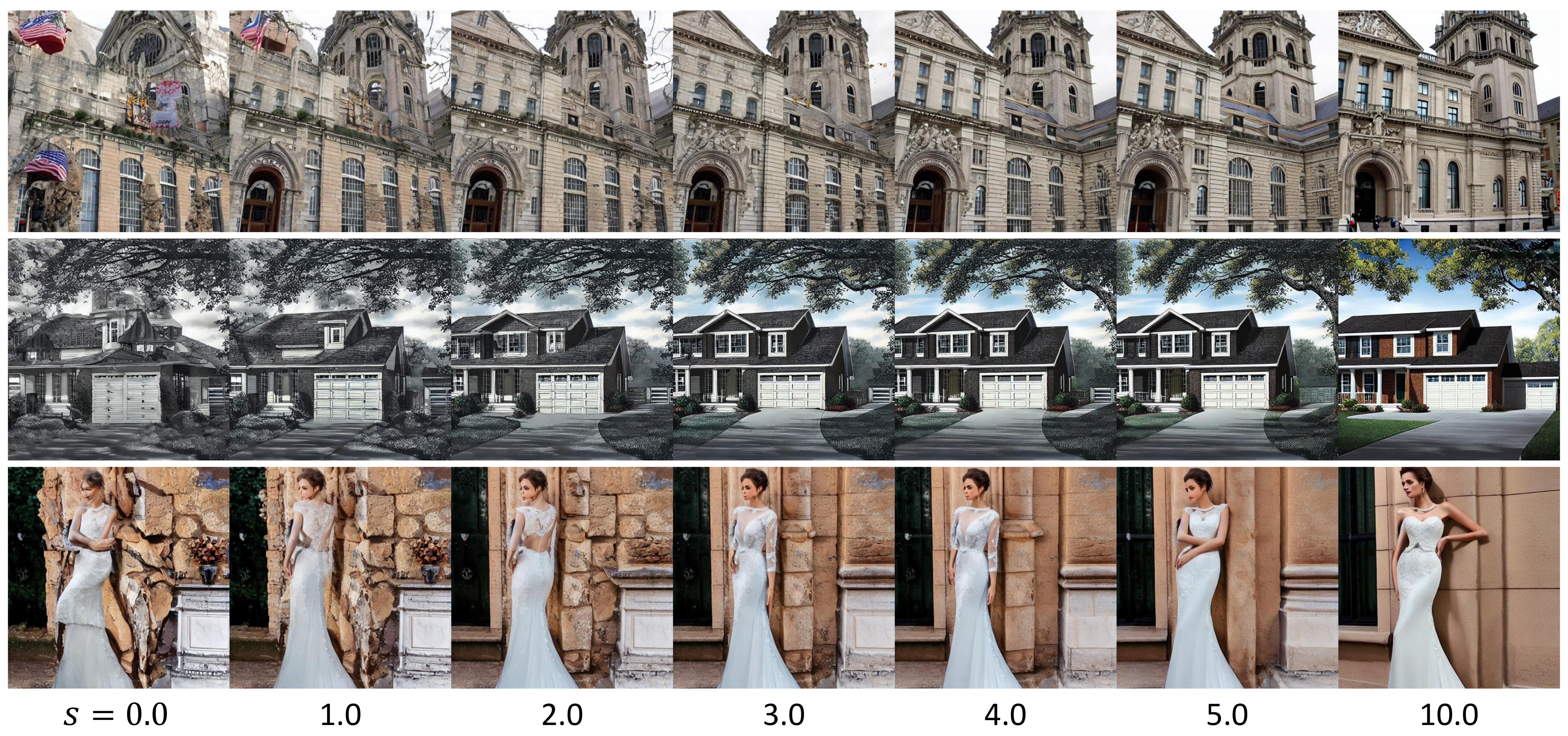} 
    \vspace{-5pt}
    \caption{\textbf{Effect of Guidance Scale on Image Quality.} Increasing the guidance scale $s$ results in images with more semantically coherent structures and fewer artifacts, thereby improving their overall quality. However, an excessively large guidance scale can lead to smoother textures and slight saturation in the images, similar to the effects observed with CFG~\cite{ho2022classifier}.}
    \label{fig:secondstage}
    \vspace{-5pt}
\end{figure*}

We conduct experiments to investigate the performance difference based on the guidance scale. Using scales set from 0.0 to 7.0 with intervals of 1.0, we sampled 5K images with ADM~\cite{dhariwal2021diffusion} and measured FID~\cite{heusel2017gans}, IS~\cite{salimans2016improved}, Precision, and Recall metrics~\cite{kynkaanniemi2019improved} for these images. The results can be seen in the graph in Fig.~\ref{fig:scalequal}. PAG showed the best FID at a guidance scale of 1.0 and the best IS at a guidance scale of 2.0.

Additionally, we conduct a qualitative comparison of the guidance scale for unconditional generation using Stable Diffusion~\cite{rombach2022high}. In Fig.~\ref{fig:secondstage}, it can be observed that as the guidance scale increases from 0.0, the structure of the sampled images improves, leading to more natural images with fewer artifacts.




\subsection{Perturbation on Self-Attention Maps}
\label{sec:sup:perturbation-ablations}
\begin{figure}[h]
    \centering
    \captionsetup[subfigure]{justification=centering, singlelinecheck=false} 
    \begin{subfigure}[t]{0.12\textwidth}
        \includegraphics[width=\textwidth]{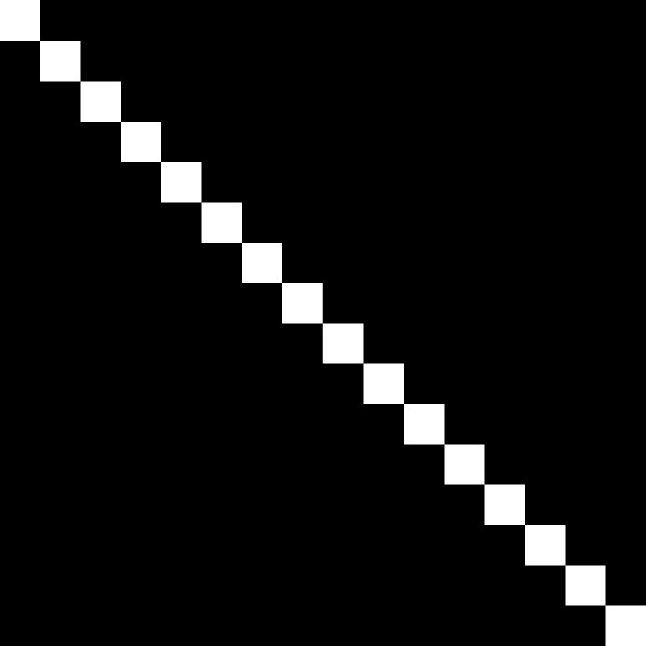}
        \caption{}
    \end{subfigure}
    \hspace{5mm} 
    \begin{subfigure}[t]{0.12\textwidth}
        \includegraphics[width=\textwidth]{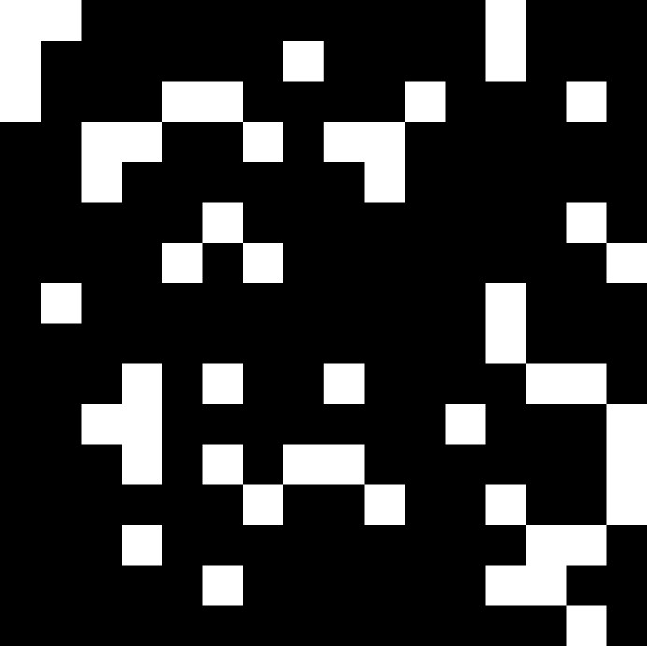}
        \caption{}
    \end{subfigure}
    \hspace{5mm}
    \begin{subfigure}[t]{0.12\textwidth}
        \includegraphics[width=\textwidth]{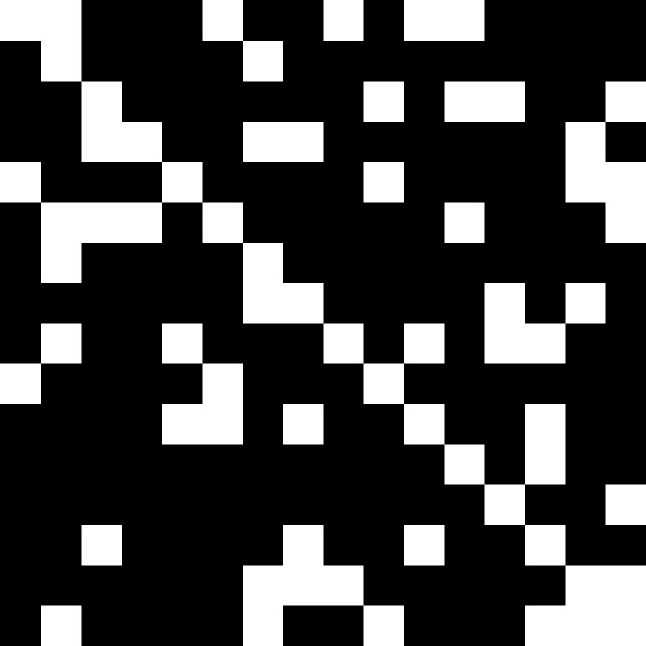}
        \caption{}
    \end{subfigure}
    \vspace{-5pt}
    \caption{\textbf{Visualization of self-attention map masking strategy}. For the evaluation of FID~\cite{heusel2017gans}, we sample 5K images from  ADM~\cite{dhariwal2021diffusion} ImageNet~\cite{deng2009imagenet} 256$\times$256 unconditional model for each method. Black entries indicate the masked (set to $-\infty$) elements of the self-attention map $\mathbf{A_t}$ in Eq.~\ref{equ:attention} before the $\mathrm{Softmax}$ operation is applied. \textbf{(a)} Replacing attention map with identity matrix. \textbf{FID: 32.34}, \textbf{(b)} Random masking (ratio: 0.25). \textbf{FID: 40.20}, \textbf{(c)} Random masking off-diagonal entries (ratio: 0.25). \textbf{FID: 39.49}.}
    \label{fig:pag_self-attention_mask}
    \vspace{-15pt}
\end{figure}

\begin{figure}
    \centering
    \includegraphics[width=\textwidth]{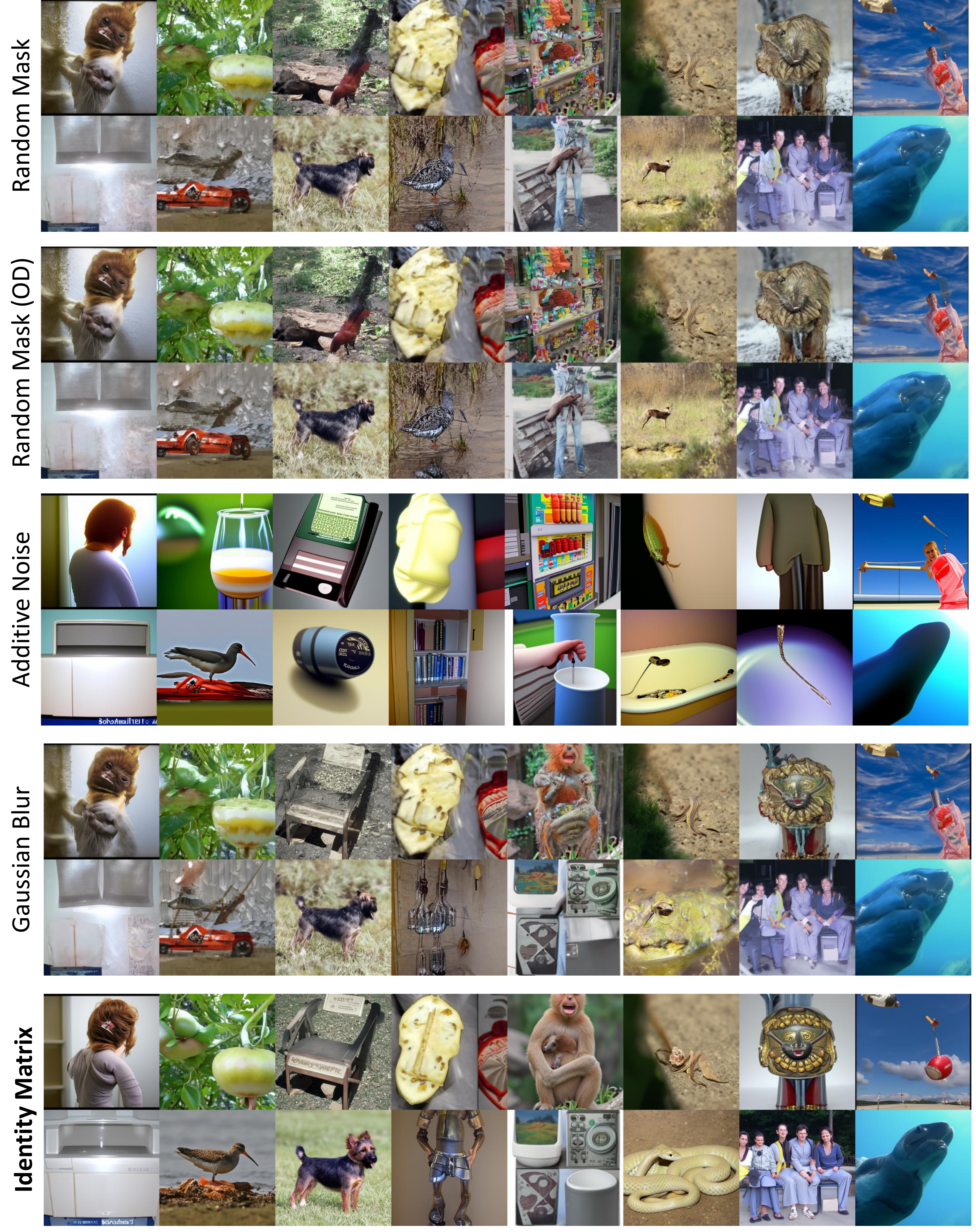}
    \caption{\textbf{Uncurated samples from ADM~\cite{dhariwal2021diffusion} with different perturbations on the self-attention map.
    } Random Mask (OD) means masking on off-diagonal entries of the self-attention map. Note that all samples are not curated and use same layer to perturb (\texttt{input.13}) and same guidance scale ($s=1.0$). The results clearly show that samples with identity matrix replacement generate plausible structures and semantics. In contrast, other perturbations often result in over-smoothed textures (additive noise) or introduce artifacts (other perturbations).}
    \label{fig:sup:perturbation-ablations}
\end{figure}

We explored various perturbation techniques that modify the structure part of self-attention, \(\mathrm{Softmax}({Q_t}{K^T_t}/\sqrt{d}) \in \mathbb{R}^{hw \times hw}\) in Eq.~\ref{equ:attention}. These methods include replacing the attention map with an identity matrix, applying random masking, and selectively masking off-diagonal entries, as illustrated in Fig.~\ref{fig:pag_self-attention_mask}. We also tried additional perturbations, including applying Gaussian blur to the self-attention map and adding Gaussian noise to it. The quantitative results are detailed in Table.~\ref{tab:sup:perturbation-ablations}. The qualitative outcomes are depicted in Fig.~\ref{fig:sup:perturbation-ablations}, illustrating that substituting the self-attention map with an identity matrix enhances image realism by minimizing artifacts and making the objects' structure semantically plausible. For additive noise, we use \(\sigma=0.1\) for Gaussian noise, and for Gaussian blur, we apply a blur kernel with a kernel size of 5 and a blur sigma of 1.0.

\begin{table}[h]
  \centering
  \vspace{40pt}
  \caption{\textbf{Ablation study on perturbations.} We sampled 5K images from the ADM~\cite{dhariwal2021diffusion} ImageNet~\cite{deng2009imagenet} 256$\times$256 unconditional model. Perturbations are applied to the same layer (\texttt{input.13}) and the same guidance scale ($s=1.0$) is used.}
  \captionsetup{skip=5pt}

  \resizebox{0.4\textwidth}{!}{
  \begin{tabular}{ll}
    \toprule
    Perturbation strategy & FID $\downarrow$ \\
    \midrule
    Random Mask & 40.20 \\
    Random Mask (off-diag) & 39.49 \\
    Additive Noise & 62.83 \\
    Gaussian Blur & 35.48 \\
    \textbf{Identity Matrix} & \textbf{32.34} \\
    \bottomrule
  \end{tabular}}
  \label{tab:sup:perturbation-ablations}
\end{table}

\newpage

\subsection{Layer Selection}
\begin{figure*}[!h]
    \centering
    \includegraphics[width=0.8\textwidth]{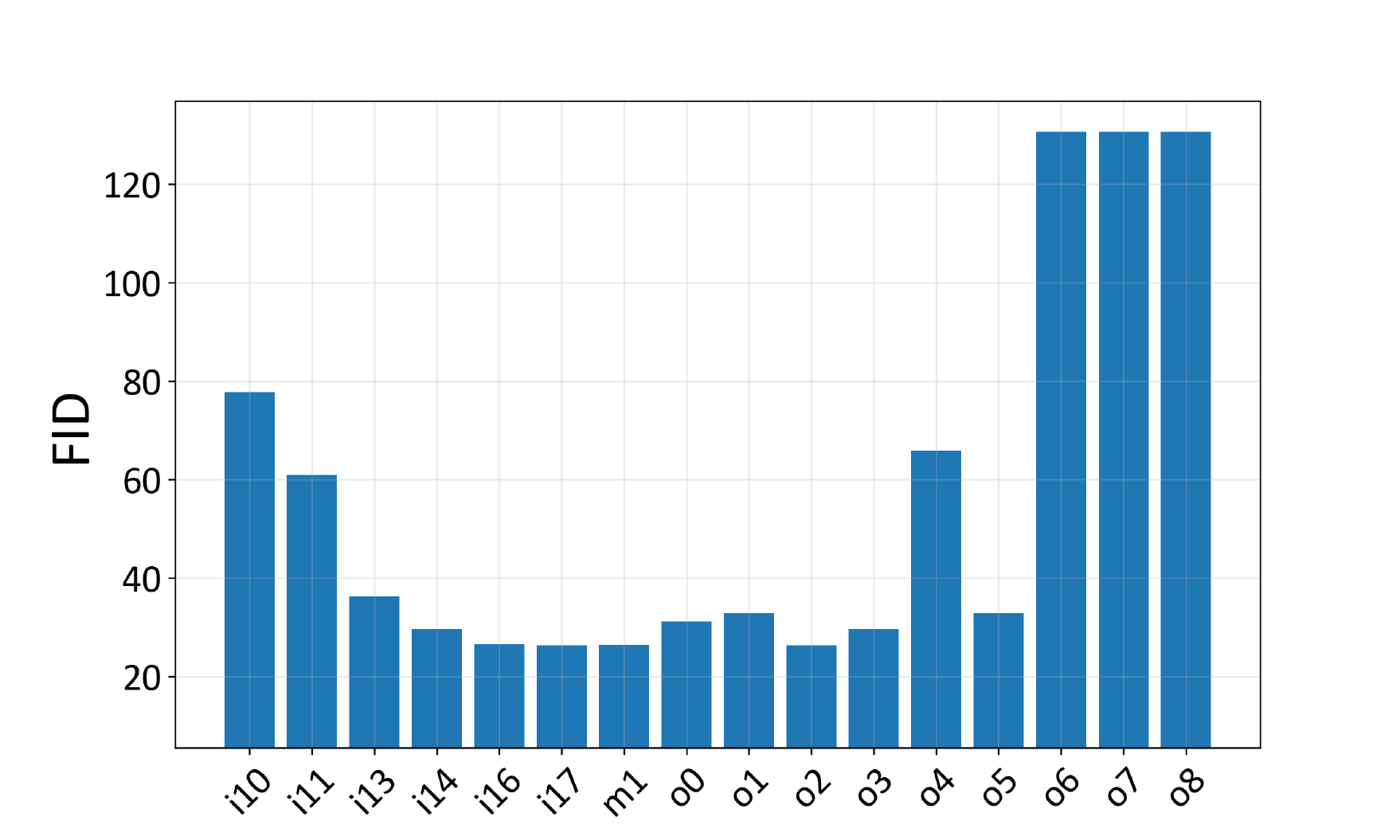}
    \caption{\textbf{Ablation study on which layer to apply perturbation with ADM~\cite{dhariwal2021diffusion}.}}
    \label{fig:sup:layer-ablations}
    \vspace{-15pt}
\end{figure*}

\begin{figure*}[!h]
    \centering
    \includegraphics[width=0.8\textwidth]{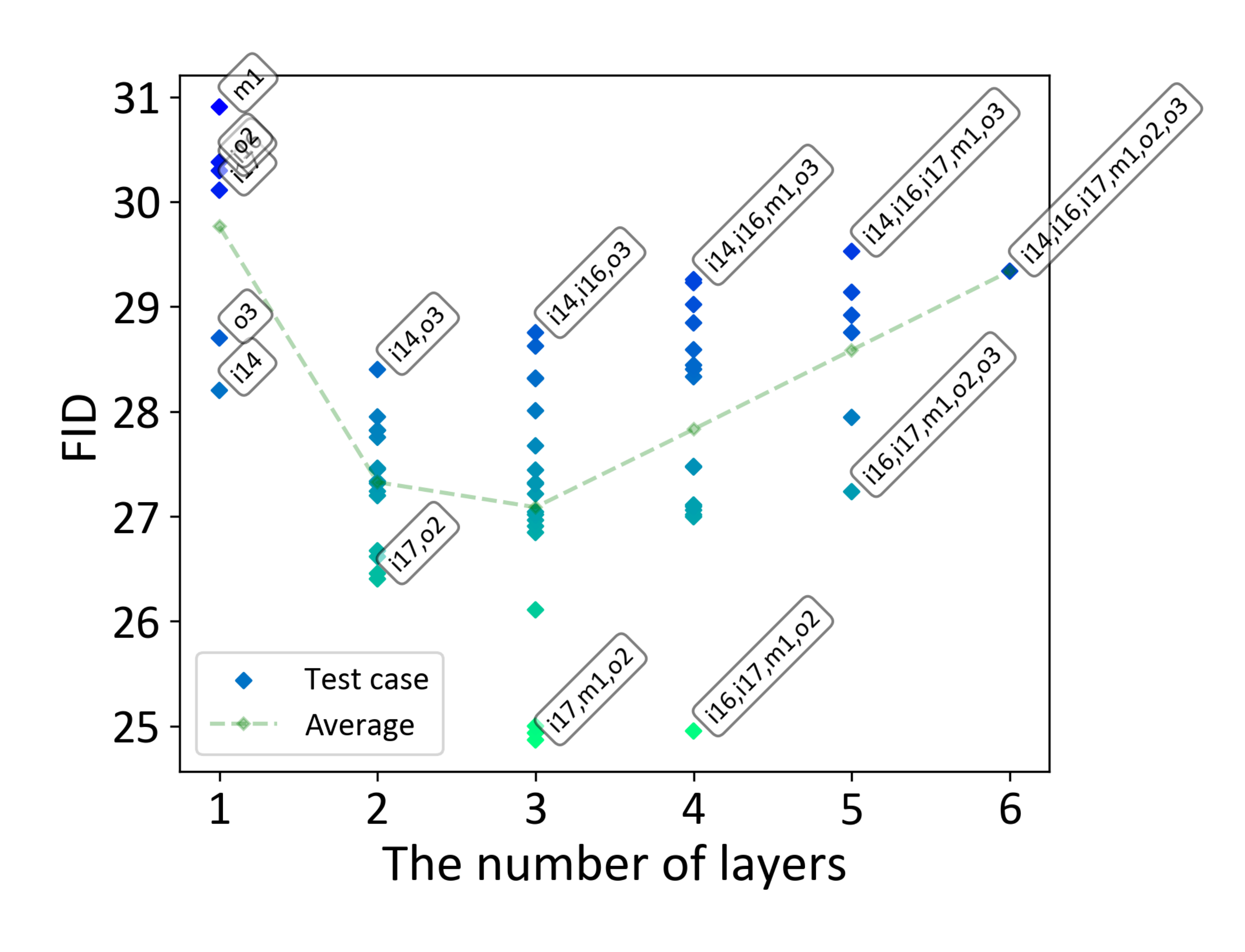}
        \caption{\textbf{Ablation study on layer combination for perturbed self-attention application in ADM~\cite{dhariwal2021diffusion}.} Each data point represents the FID obtained when perturbed self-attention (PSA) is applied to the corresponding combination of layers. The annotations of points represent the combined layers. The green dashed line denotes the average FID across all combinations for a given number of layers involved. This analysis reveals that applying PSA to multiple layers can enhance sample quality to a certain extent. However, this trend does not hold for Stable Diffusion~\cite{rombach2022high}, indicating that the effectiveness of layer-wise perturbation varies across different diffusion models.}
    \label{fig:sup:layer-combination-ablations}
    \vspace{-15pt}
\end{figure*}

We conduct an ablation study to determine the optimal layers for perturbing the self-attention map with the outcomes presented in Fig.~\ref{fig:sup:layer-ablations}. The experiments include both ADM~\cite{dhariwal2021diffusion} ImageNet 256×256 unconditional model and Stable Diffusion~\cite{rombach2022high}.
Observations indicate that perturbations applied to deeper layers generally yield relatively better outcomes compared to those applied to shallower layers of U-Net~\cite{ronneberger2015u}. 
We apply perturbations to all combinations of the top-6 layers (\texttt{input.14}, \texttt{input.16}, \texttt{input.17}, \texttt{middle.1}, \texttt{output.2}), as ranked by FID, and present the results in Fig.~\ref{fig:sup:layer-combination-ablations} and Table~\ref{table:supple_layer_abl}. Some combinations show improved results for the ADM unconditional model but do not yield the same improvements in the case of Stable Diffusion\cite{rombach2022high}. Additionally, although experiments involving the random selection of layers for each timestep were conducted, we discover that selecting fixed layers across timesteps yields better outcomes.

Fig.~\ref{fig:sup:layer-ablations} visualizes the FID scores obtained by perturbing each layer of the ADM~\cite{dhariwal2021diffusion} ImageNet~\cite{deng2009imagenet} 256$\times$256 unconditional models. A guidance scale of $s=1.0$ is employed. FID scores are calculated using 5K image samples. Note that outlier values (\texttt{o6}, \texttt{o7}, \texttt{o8}) are clipped. It can be seen that perturbations on deeper layers, particularly near the bottleneck layer of U-Net, tend to show relatively better performance than those on shallower layers. The ablation results through DDIM~\cite{song2020denoising} 25 step sampling are as follows, and in the case of sampling 5K images with DDPM~\cite{ho2020denoising} 250 step sampling, the layer we selected on Table.~\ref{table:adm-quan} shows the highest performance.

\begin{figure*}[!h]
    \centering
    \includegraphics[width=0.8\textwidth]{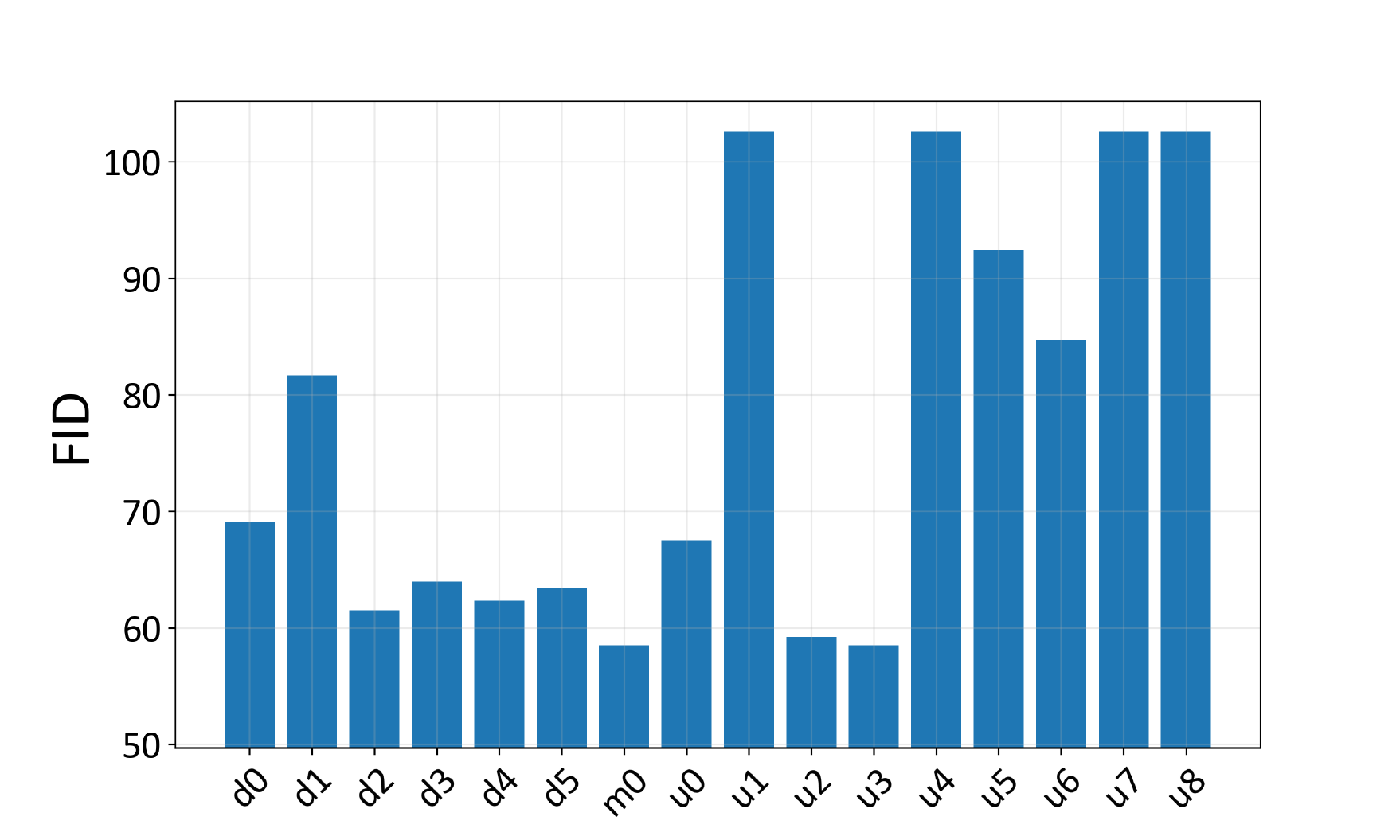}
    \caption{\textbf{Ablation study on which layer to apply perturbation with Stable Diffusion~\cite{rombach2022high}.}}
    \label{fig:sup:layer-ablations-sd}
\end{figure*}
Fig.~\ref{fig:sup:layer-ablations-sd} shows the FID results from generating 5k samples using Stable Diffusion with PAG guidance scale $s=$ 2.5 and DDIM 25 step sampling. We applied perturbation to different layers: ``d0'' represents the outermost encoder layer, ``u8'' is the outermost decoder layer, and ``m0'' is the mid-block. The best performance was achieved when perturbation was applied to the mid-block ``m0''.

\clearpage

\begin{table}[!t]
    \centering
    \caption{\textbf{Layer ablation on ADM.} We evaluate the FID~\cite{heusel2017gans} of 5K samples from ImageNet~\cite{deng2009imagenet} 256$\times$256 unconditional model using DDIM~\cite{song2020denoising} 25 step sampling.}
    \captionsetup{skip=5pt}
    \setlength{\tabcolsep}{25pt}
    \resizebox{0.6\textwidth}{!}{
    \begin{tabular}{ccc}
        \toprule
        \# layers & layers & FID $\downarrow$ \\
        \midrule
        \multirow{6}{*}{1} & i14 & 28.20 \\
        & i16 & 30.30 \\
        & i17 & 30.11 \\
        & m1 & 30.90 \\
        & o2 & 30.38 \\
        & o3 & 28.70 \\
        \midrule
        \multirow{15}{*}{2} & i14 i16 & 27.95 \\
        & i14 i17 & 27.82 \\
        & i14 m1 & 27.82 \\
        & i14 o2 & 26.62 \\
        & i14 o3 & 28.40 \\
        & i16 i17 & 27.20 \\
        & i16 m1 & 27.31 \\
        & i16 o2 & 26.45 \\
        & i16 o3 & 27.45 \\
        & i17 m1 & 27.33 \\
        & i17 o2 & 26.40 \\
        & i17 o3 & 27.24 \\
        & m1 o2 & 26.67 \\
        & m1 o3 & 27.46 \\
        & o2 o3 & 27.75 \\
        \midrule
        \multirow{20}{*}{3} & i14 i16 i17 & 28.33 \\
        & i14 i16 m1 & 28.01 \\
        & i14 i16 o2 & 27.02 \\
        & i14 i16 o3 & 28.75 \\
        & i14 i17 m1 & 27.67 \\
        & i14 i17 o2 & 26.85 \\
        & i14 i17 o3 & 28.62 \\
        & i14 m1 o2 & 26.91 \\
        & i14 m1 o3 & 28.31 \\
        & i14 o2 o3 & 28.32 \\
        & i16 i17 m1 & 26.11 \\
        & i16 i17 o2 & 25.00 \\
        & i16 i17 o3 & 27.04 \\
        & i16 m1 o2 & 24.93 \\
        & i16 m1 o3 & 27.21 \\
        & i16 o2 o3 & 27.31 \\
        & i17 m1 o2 & \textbf{24.87} \\
        & i17 m1 o3 & 26.96 \\
        & i17 o2 o3 & 27.44 \\
        & m1 o2 o3 & 27.32 \\
        \midrule
        \multirow{15}{*}{4} & i14 i16 i17 m1 & 28.44 \\
        & i14 i16 i17 o2 & 27.47 \\
        & i14 i16 i17 o3 & 29.23 \\
        & i14 i16 m1 o2 & 27.48 \\
        & i14 i16 m1 o3 & 29.26 \\
        & i14 i16 o2 o3 & 28.59 \\
        & i14 i17 m1 o2 & 27.09 \\
        & i14 i17 m1 o3 & 29.02 \\
        & i14 i17 o2 o3 & 28.84 \\
        & i14 m1 o2 o3 & 28.40 \\
        & i16 i17 m1 o2 & 24.95 \\
        & i16 i17 m1 o3 & 27.11 \\
        & i16 i17 o2 o3 & 27.01 \\
        & i16 m1 o2 o3 & 27.06 \\
        & i17 m1 o2 o3 & 27.00 \\
        \midrule
        \multirow{6}{*}{5} & i14 i16 i17 m1 o2 & 27.94 \\
        & i14 i16 i17 m1 o3 & 29.53 \\
        & i14 i16 i17 o2 o3 & 29.14 \\
        & i14 i16 m1 o2 o3 & 28.92 \\
        & i14 i17 m1 o2 o3 & 28.75 \\
        & i16 i17 m1 o2 o3 & 27.24 \\
        \midrule
        6 & i14 i16 i17 m1 o2 o3 & 29.34 \\
        \bottomrule
    \end{tabular}
    }
    \label{table:supple_layer_abl}
\end{table}

\clearpage

\begin{figure*}[!t]
    \centering
    \includegraphics[width=\textwidth]{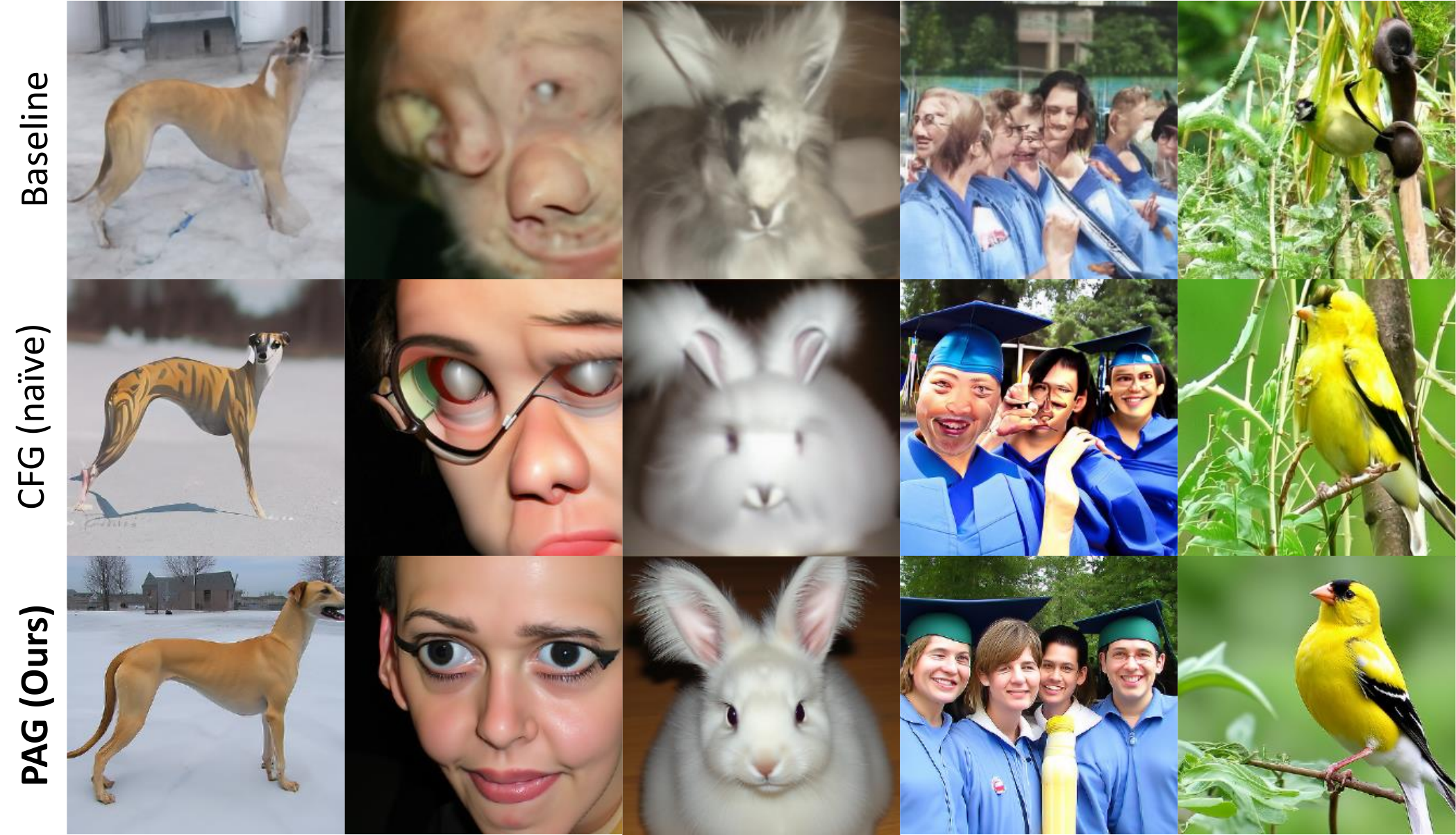}
    \caption{\textbf{Samples Using CFG with Separately Trained Models.} We implement CFG~\cite{ho2022classifier} by employing separately trained ADM~\cite{dhariwal2021diffusion} ImageNet 256×256 conditional and unconditional models. Compared to samples with PAG in row 3, samples with naïve CFG in row 2 show inferior image quality. This suggests that when the conditional prediction \(\epsilon_{\theta_1}(x_t,c)\) and \(\epsilon_{\theta_2}(x_t)\) do not align, the guiding signal becomes ineffective, resulting in low-quality samples, where \(\theta_1\) and \(\theta_2\) are parameters from the conditional and unconditional models, respectively. Here, a guidance scale of 3.0 is employed for both naïve CFG and PAG, using the same seed and latent.}
    \label{fig:sup:naive-cfg}
\end{figure*}

\section{Discussion}
\label{sec:sup:discussion}

\subsection{Theoretical Insights on Using Identity Matrix as Perturbation}
\label{sec:sup:theorticial-insights}

Several studies have sought to establish its theoretical foundation, with a promising approach being its interpretation through pattern storage and retrieval behavior within the Energy-Based Model (EBM) framework. Based on this, we will explain why replacing the identity matrix works. 
Hopfield networks~\cite{j.1982,j.1984,krotov2016,krotov2017,demircigil2017,pascal1994} are associative memories that retrieve the pattern most similar to the input. They model an energy landscape with basins of attraction around desired patterns. \cite{ramsauer2020hopfield} generalizes the energy function for continuous embeddings and demonstrates that the proposed update rule ensures global convergence: $\boldsymbol{\Xi}_{n+1} = \boldsymbol{X} \ \mathrm{softmax} \left( \boldsymbol{X}^T \boldsymbol{\Xi}_n \right).$

As shown in~\cite{ramsauer2020hopfield}, this implicit energy minimization equation is closely linked to the self-attention forward pass of transformers by mapping \(X\) to \(K\) and \(\Xi\) to \(Q\) via projection matrices and introducing \(W_v\) for the key contents: $
\mathbf{Q}^{\mathrm{new}} = \mathrm{softmax} \left( \frac{\mathbf{Q} \mathbf{K}^T}{\sqrt{d}} \right) \mathbf{V}.$ This connection provides an insightful theoretical foundation for the attention mechanism. It suggests that the transformer's attention mechanism operates as an inner-loop optimization step minimizing the energy function determined by queries, keys, and values.

\begin{figure}
    \centering
    \includegraphics[width=\textwidth]{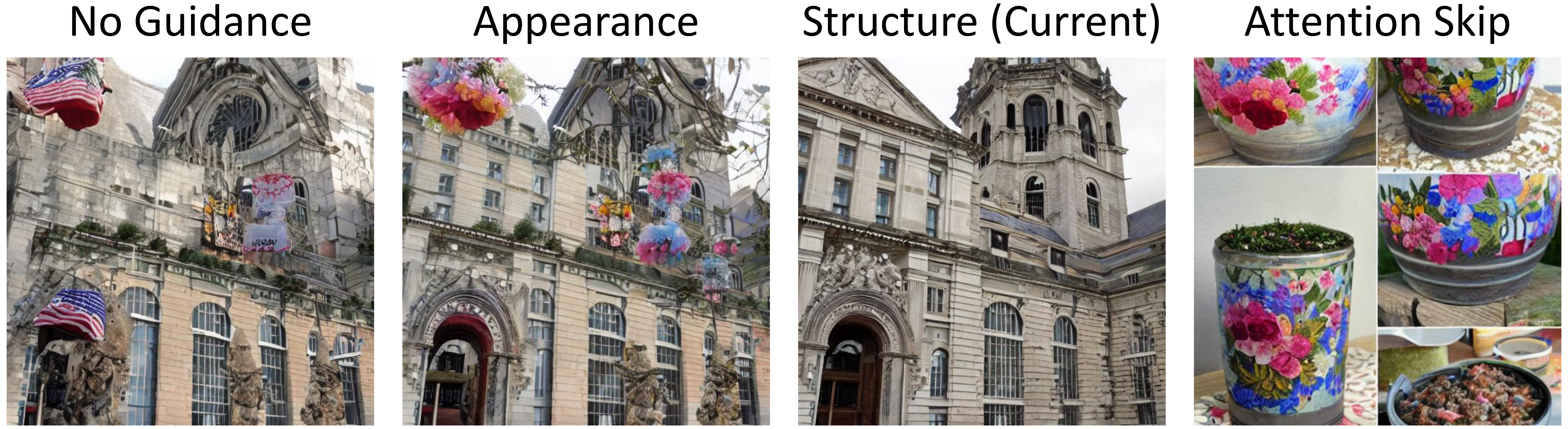}
    \caption{\textbf{Sampled images with various perturbations.} Instead of replacing \textit{structure} component $\mathrm{Softmax}$ $({Q_t}{K^T_t}/\sqrt{d})$ with an identity matrix (Structure (Current)), we test perturbation on \textit{appearance} component, replacing all tokens in $V_t$ with its spatial average (Appearance). We also completely skip self-attention by ignoring self-attention which is learned in a residual manner (Attention Skip).}
    \label{fig:perturb}
\end{figure}

Thus, the forward pass of self-attention corresponds to \textit{pattern retrieval} in the Hopfield network, and the backward pass updates the projection matrices ($W_q, W_k, W_v$) to reduce the final loss, implicitly learning to map the inputs to \textit{useful patterns}. Specifically, \(W_q\) and \(W_k\) learn to model the relationships between inputs, while \(W_v\) learns the content patterns to be aggregated. Through pattern matching, self-attention effectively captures contextual relationships within the input.

We suggest replacing $\mathrm{Softmax}({Q_t}{K^T_t}/\sqrt{d})$ with an identity matrix to remove these relationships, resulting in an \textit{undesirable} distribution in terms of structure. However, we keep the learned \textit{content patterns} by passing value features, preserving local texture to make distribution \textit{in-domain}. If we ignore both context and content patterns (value), it results in completely different images (Fig.~\ref{fig:perturb} ‘Attention Skip’). Thus, identity-attention is a method intended to \textbf{maximize the use of learned self-attention patterns} to create an \textit{in-domain} but \textit{undesirable} distribution.

In summary, we selectively use learned intermediate representations to model the undesirable in terms of structure but still in-domain distribution by keeping the appearance information. There might be better perturbations but we observed that identity matrix replacement is an effective method, theoretically and empirically. We leave it for future works to find better perturbations for different tasks and models.

\subsection{Further Analysis on CFG and PAG}
\label{sec:sup:cfg-further-analysis}
\subsubsection{CFG with separately trained models.}
As mentioned in the Sec.~\ref{subsubsec:cfg} in the main paper, the guidance term in \textbf{CFG}~\cite{ho2022classifier} originates from the gradient of the \textit{implicit classifier} derived from Bayes' rule. Therefore, in principle, CFG can be implemented by training the conditional and unconditional models separately. However, the authors implemented it using a single neural network by assigning a null token as the class label for the unconditional model. They mentioned, ``It would certainly be possible to train separate models instead of jointly training them together,'' suggesting it as an option during design. But in practice, this is not the case. We discover that as can be seen in Fig.~\ref{fig:sup:naive-cfg}, implementing CFG with separately trained conditional and unconditional models does not work properly (2nd row). This implies that CFG enhances image quality not merely by trading diversity but operates by some other key factor. The secret may be that as analyzed in Fig.~\ref{fig:cfg-ours-visualization} of main paper, predicting a sample missing salient features (such as eyes and nose) from the original conditional prediction and then adding the difference to reinforce those salient features. In other words, simply subtracting the unconditional generation made by a separate model does not suffice for its operation, highlighting the utility of our \textbf{PAG}. While CFG creates a \textit{perturbed} path missing salient features at the additional cost of training an unconditional model jointly, perturbed self-attention (PSA) in our \textbf{PAG} can produce predictions missing such salient features without any additional training or external model, simply by manipulating the self-attention map of U-Net. Especially when compared to SAG~\cite{hong2023improving} and other perturbations (perturbation ablations and Sec.~\ref{sec:sup:perturbation-ablations}), PSA can be considered an efficient and effective method.
\subsubsection{Connections to delta denoising score.}
According to prior works~\cite{hertz2023delta,katzir2023noise} that use diffusion models for score distillation sampling (SDS), the term \(\hat{\epsilon}_\theta\) in our guidance framework can be interpreted as the model's inherent \textit{bias}. Delta denoising score~\cite{hertz2023delta} suggests that when conducting SDS, the gradient term contains \textit{bias}, and by subtracting this from another gradient obtained with similar prompts, structures, and the same noise, one can eliminate the shared noisy components. From this perspective, CFG~\cite{ho2022classifier} and PAG can be interpreted as the removal of noisy components, which make locally aligned structures, in the diffusion model's epsilon prediction as class label dropping and attention perturbing, respectively. This perspective underscores the importance of carefully calibrating perturbations to avoid significant deviations from the original sample. SAG~\cite{hong2023improving} has shown a tendency to produce samples that diverge excessively from the original sample, 
due to aggressive perturbation applied directly to the model's input, leading to out-of-distribution (OOD) samples and high hyperparameter sensitivity. Scale-wise qualitative result illustrates that PAG exhibits lower sensitivity to scale adjustments, attributed to the strategic perturbed self-attention approach, which preserves appearance information of the original sample. For a comprehensive comparison, see Sec.~\ref{sec:comparision_with_sag}.

\subsubsection{Additional training for stability.}
Although our carefully designed perturbed self-attention (PSA, e.g., self-attention map replacement with identity) method effectively mitigates the out-of-distribution (OOD) issue without additional training, incorporating training can further improve its ability to address the OOD problem and enhance its robustness to hyperparameter settings. 

Similar to various self-supervised learning or augmentation techniques~\cite{he2022masked,srivastava2014dropout,wan2013regularization} that achieve comparable results with augmented inputs/models to those with original inputs/models, $\hat{\epsilon}_\theta$ can be trained with PSA to produce more stable samples that maintain appearance and lack structural information. This improvement can be achieved by introducing a switching input to control the on/off status of self-attention map usage and fine-tuning the model while providing this switching input. Compared to training a new unconditional model for CFG~\cite{ho2022classifier}, fine-tuning the model incurs lower computational costs.
Furthermore, unlike CFG, which entangles sample quality and diversity, PAG with trained 
$\hat{\epsilon}_\theta$ enhances sample quality without compromising diversity. We leave this exploration as future work.

\subsection{Complementarity between CFG and PAG}
\label{sec:sup:complementarity-between-pag-sag}
Recent research~\cite{balaji2022ediffi,park2024understanding} has shed light on the \textit{temporal dynamics} of text-to-image diffusion models during their sampling process. The analysis, focusing on the model's self-attention and cross-attention maps under different noise conditions, demonstrates a transition in the model's operational focus from text to the pixels being generated. Initially, at the beginning of the sampling process where the network's input is close to random noise, the model significantly relies on the text prompt for direction in the sampling. However, as the process continues, there's a noticeable shift towards leveraging visual features for image denoising, showing higher activation of self-attention map, with the model gradually paying less attention to the text prompt. This shift is logical; in the early stage, the model relies on the prompt for cues on what to denoise in the image. As the denoising process progresses and the images take shape, the model shifts focus to refine these emerging visual details.

\begin{figure}
    \centering
    \includegraphics[width=\textwidth]{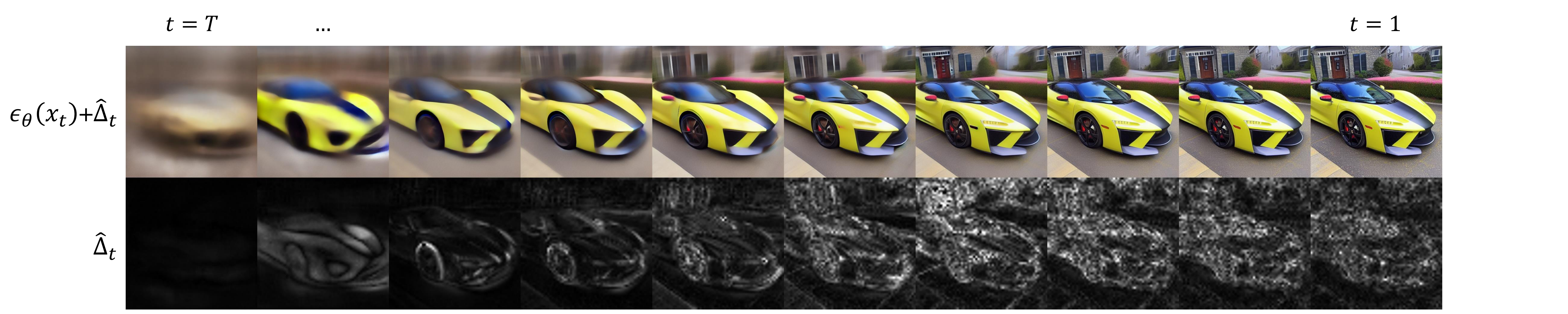}
    \caption{\textbf{Visualization of $\hat{\Delta}_t={\epsilon}_{\theta}(x_t) - {\hat{\epsilon}}_{\theta}(x_t)$ during reverse process with PAG.} text-to-image generation using PAG with a prompt \textit{``a fancy sports car''.}}
    \label{fig:sup:pag-delta_vis}
\end{figure}
This phenomenon can also be observed in PAG and CFG contexts. Fig.~\ref{fig:sup:pag-delta_vis} visualizes $\hat{\Delta}_t ={\epsilon}_{\theta}(x_t) - {\hat{\epsilon}}_{\theta}(x_t)$ during sampling with PAG. As mentioned earlier, since the self-attention map is not highly activated in the early stages of the diffusion sampling process, the difference $\hat{\Delta}_t$ between the predicted epsilon with self-attention map dropped and the original predicted epsilon appears weak initially. As the sampling process progresses and the image starts to take shape, the activation of the self-attention map gradually strengthens, leading to an increasing $\hat{\Delta}_t$ observable over time.

\begin{figure}
    \centering
    \includegraphics[width=\textwidth]{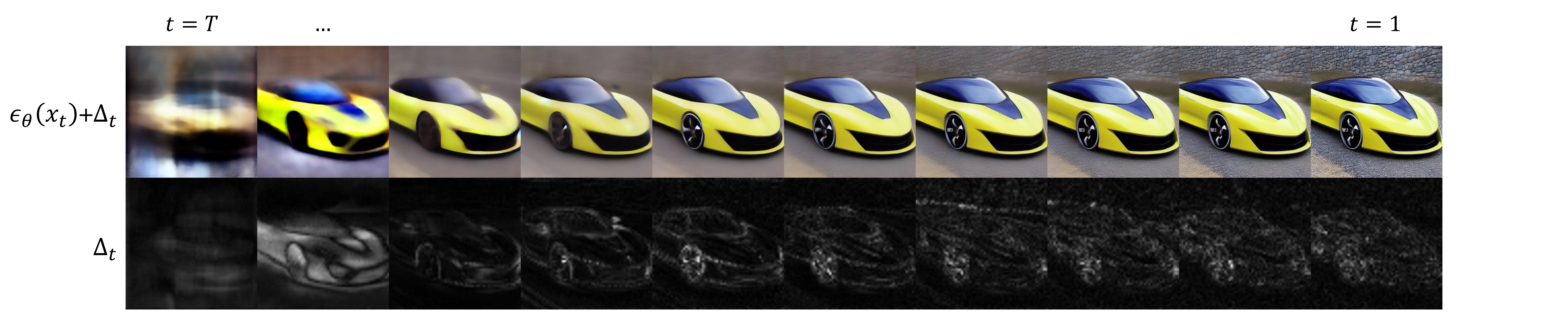}
    \caption{\textbf{Visualization of $\Delta_t= {\epsilon}_{\theta}(x_t,c) - {{\epsilon}}_{\theta}(x_t,\phi)$ during reverse process with CFG.} text-to-image generation using CFG with a prompt \textit{``a fancy sports car''.}}
    \label{fig:sup:cfg-delta_vis}
\end{figure}
In contrast, CFG exhibits a different behavior. Fig.~\ref{fig:sup:cfg-delta_vis} displays the timestep-wise predicted epsilon difference $\Delta_t = {\epsilon}_{\theta}(x_t,c) - {{\epsilon}}_{\theta}(x_t,\phi)$ during sampling with CFG. As previously discussed, in the initial stages of generation, the diffusion model predominantly relies on the prompt to create images, leading to high activation in the cross-attention map. CFG creates a perturbed path using a null prompt for the prompt, which can be understood as applying perturbation to the cross-attention. (Indeed, we observe that making cross-attention map zero yield effects somewhat similar to CFG, though these effects were suboptimal.) Therefore, a high $\Delta_t$ is observed in the early stages of the sampling process, where the model focuses on the prompt, and the difference diminishes later on.

\begin{figure}
    \centering
    \includegraphics[width=0.7\textwidth]{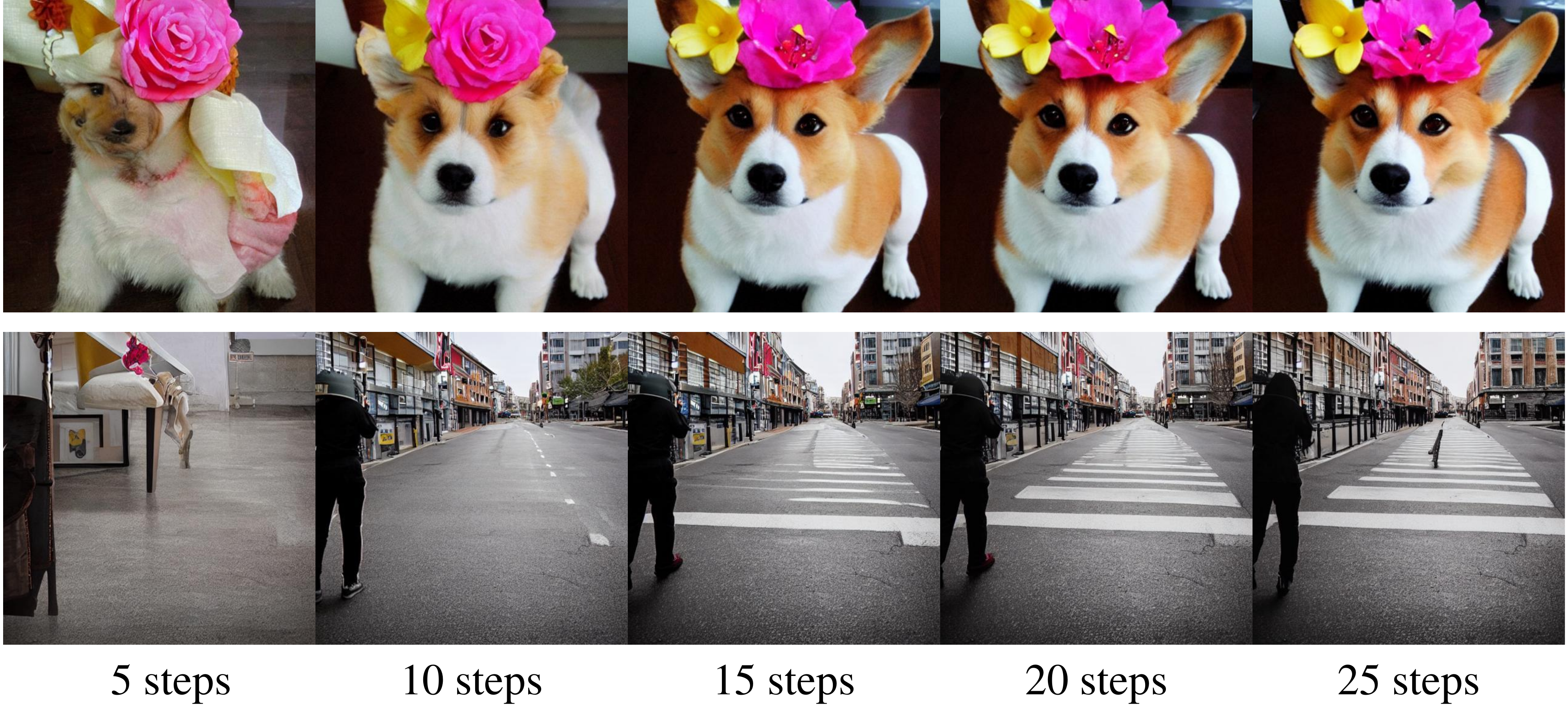}
    \caption{\textbf{Early stopping of CFG~\cite{ho2022classifier}.} The process involves a total of 25 steps. The prompts used are \textit{``A corgi with a flower crown''} (top) and \textit{``A person walking on the street''} (bottom).}
    \label{fig:sup:cfg-interval}
\end{figure}
The impact of CFG being primarily in the early stages of the generation process can be validated through another observation. Fig.~\ref{fig:sup:cfg-interval} shows the results when CFG is applied only in the first 5 steps, then in the first 10, 15, 20, and throughout all 25 steps of a 25 step generation process. It can be observed that applying CFG for the initial 60\% of the total steps (15 steps) yields results comparable to those achieved when CFG is utilized for the full 25 steps.

Compared with CFG, PAG continues to influence throughout the mid to late stages of the timestep, offering highly detailed guidance, especially in the latter half as can be seen in Fig.~\ref{fig:sup:pag-delta_vis}. This indicates that PAG continues to provide a positive signal even in the later stages.

Therefore, using CFG and PAG together can guide the image towards better quality across the entire sampling process. 
This approach effectively utilizes both the self-attention and cross-attention maps, resulting in effective guidance throughout the entire timestep.
Indeed, qualitative comparison between CFG and CFG $+$ PAG, and Stable Diffusion quantitative results demonstrate that combining CFG and PAG yields superior outcomes compared to employing CFG alone. We also present a human evaluation of samples utilizing CFG versus CFG $+$ PAG in Fig.~\ref{fig:human_eval}.

\subsection{Comparison with SAG}
\label{sec:sup:comparision-with-sag}
In this section, we summarize the differences between SAG~\cite{hong2023improving} and PAG, focusing on their formulation, stability, speed, and effectiveness. SAG emerged as an initial method for enhancing guidance in unconditional generation within diffusion models.

\subsubsection{Generalizability.}
Both SAG and PAG aim to generalize guidance, albeit through distinct formulations. SAG proposes an imaginary regressor $p_\mathrm{im}$ to predict ${h}_t$ given ${ x}_t$, where ${h}_t$ represents a generalized condition including external condition or internal information of ${x}_t$ or both, and $\bar{x}_t$ is a perturbed sample missing ${ h}_t$ from ${ x}_t$. 
For instance, blur guidance in SAG leverages $\bar{x}_t=\tilde{x_t}$ and ${h}_t = {x}_t - \tilde{x}_t$, where $\tilde{x}_t$ represents a sample with the high-frequency components of the original sample ${x}_t$ removed. Specifically, $\hat{x}_0$ is derived from $x_t$ by Eq.~\ref{eq:eps-to-x0} and subsequently blurred using a Gaussian filter $G_\sigma$ (expressed as $\tilde{x}_0 = \hat{x}_0 * G_\sigma$, with $*$ denoting a convolution operation), and then diffused back by incorporating the noise $\epsilon_\theta(x_t)$. The guided sampling can be formulated as $    \tilde{\epsilon}(\bar{x}_t, h_t) = \epsilon_\theta(\bar{x}_t, h_t)-s\sigma_t\nabla_{\bar{x}_t}\log p_{\textrm{im}}(h_t|\bar{x}_t)
$. However, this approach does not allow for perturbations on the model's internal representation, whereas SAG can be considered a specific instance within our broader framework, as $\hat{\epsilon}_\theta(\cdot)$ could represent any perturbation process, including the adversarial blurring used by SAG.

\begin{figure}
    \centering
    \includegraphics[width=\textwidth]{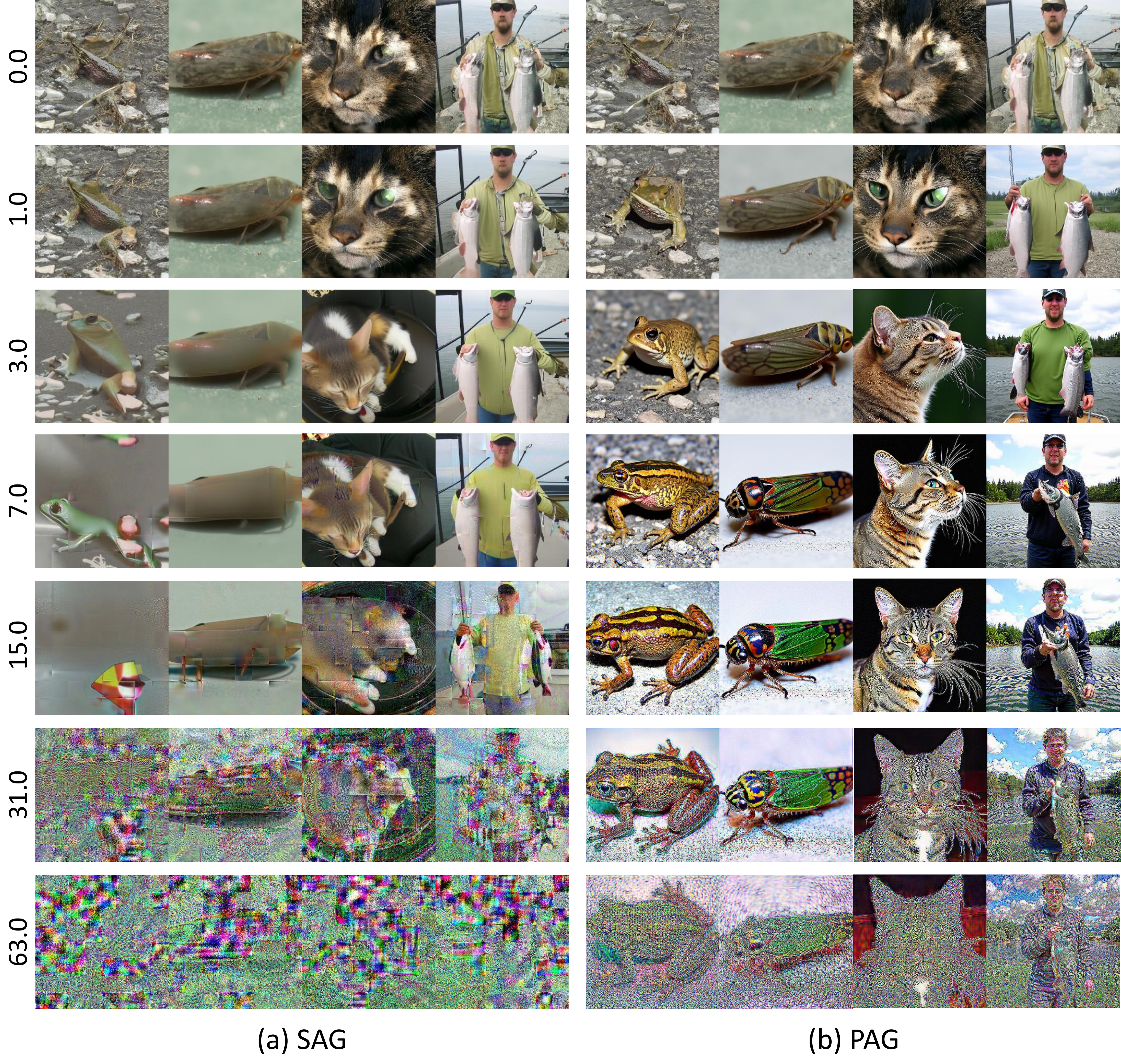}
    \caption{\textbf{Comparison of samples with SAG~\cite{hong2023improving} and PAG for different guidance scales.} Samples are generated by  ADM~\cite{dhariwal2021diffusion} conditional ImageNet 256\(\times\)256 model, showcasing the impact of incrementally increasing the guidance scale from 0.0 to 63.0, from the top to the bottom of the figure. \textbf{(a)}: Samples generated with a high guidance scale using SAG exhibit artifacts and over-smoothness due to excessive perturbation, specifically blurring on the input, with the outlines of the blur mask clearly visible. \textbf{(b)}: Compared to SAG, samples generated with higher scale PAG display high-quality results, characterized by well-structured shape and high detail. Within each group, from left to right, the classes are \textit{bell toad}, \textit{leafhopper}, \textit{tabby cat}, and \textit{silver salmon}.}
    \label{fig:sup:sag-ours-scale-comparision}
\end{figure}

\subsubsection{Hyperparameter count and sensitivity to guidance scale.}
Since SAG utilizes Gaussian blur, it requires the setting of multiple hyperparameters. Hyperparameters related to blur include the blur kernel size and the \(\sigma\) of the blur kernel. Additionally, determining the area for adversarial blurring necessitates selecting the layer from which to extract the self-attention map and specifying a threshold value. In contrast, PAG simply requires the selection of the layer to which perturbed self-attention will be applied. Additionally, as seen in Fig.~\ref{fig:sup:sag-ours-scale-comparision}, SAG is sensitive to the guidance scale. The figure shows that as the guidance scale increases, the boundaries of the adversarial mask area become visible, and high-frequency artifacts appear. Therefore, SAG cannot use a large guidance scale, which is a significant drawback considering that stronger guidance often results in greater improvements in image quality. Indeed, considering CFG employs a large scale of around 7.5, this limitation is significantly notable. In contrast, PAG maintains the plausibility of object shapes and enhances details even at relatively high guidance scales.

\subsubsection{Inference speed.} SAG requires the extraction of self-attention maps during its first forward pass and the application of blur to the model input for adversarial blurring. Our method can be implemented to handle PSA and regular self-attention within the same batch, allowing guidance to be applied with a single evaluation of the denoising neural network, similar to CFG~\cite{ho2022classifier}. Therefore, if the GPU can perform concurrent computations swiftly, PAG could theoretically be up to more than twice as fast as SAG. We discuss the results of comparing the speed of PAG, implemented in this manner, with CFG and SAG in Sec.~\ref{sec:sup:commputational_cost}.

\label{sec:comparision_with_sag}

\section{Limitation and Future Works}
Although PAG demonstrates effectiveness across various tasks, it shares certain limitations with CFG. Notably, at high guidance scales, results can exhibit over-saturation. This highlights the need for careful calibration of the guidance scale to balance quality improvement with potential visual artifacts. Additionally, PAG requires two forward paths for each generation step. Future research could explore techniques to reduce this computational overhead or develop alternative guidance mechanisms with lower resource requirements.
\label{sec:sup:limitations}

\section{Changelog}
\subsubsection{Jul 06, 2025.}  
Updated following the ECCV camera-ready version: added SDXL examples and theoretical motivation for using an identity matrix; additionally, named Eq.~\ref{eq:PAG-derivation} and included a discussion on subsequent works.

\end{document}